\documentclass[10pt,twocolumn,letterpaper]{article}

\usepackage{cvpr}              
\usepackage[most]{tcolorbox}
\usepackage{booktabs}
\usepackage{pifont}
\usepackage{multirow}
\usepackage{array}        
\usepackage{makecell}     
\usepackage{booktabs}    
\usepackage{dsfont}
\usepackage{amssymb}
\usepackage[table]{xcolor}
\usepackage{float}
\usepackage[export]{adjustbox}
\usepackage{tikz}
\usepackage{progressbar}
\usepackage{pgfplots}
\usepackage{bm}
\pgfplotsset{compat=1.18}

\usepackage[misc]{ifsym}

\definecolor{cvprblue}{rgb}{0.21,0.49,0.74}
\usepackage[pagebackref,breaklinks,colorlinks,allcolors=cvprblue]{hyperref}

\definecolor{myred}{RGB}{255,0,0}

\def\paperID{2535} 
\def\confName{CVPR}
\def\confYear{2026}

\title{
Do VLMs Perceive or Recall? Probing Visual Perception vs. Memory with Classic Visual Illusions
}
\author{
Xiaoxiao Sun$^{1}$$$\thanks{Equal contribution.} \quad
Mingyang Li$^{1*}$ \quad
Kun Yuan$^{2, 3}$ \quad
Min Woo Sun$^{1}$ \quad
Mark Endo$^{1}$ \quad
Shengguang Wu$^{1}$ \quad \\
Changlin Li$^{1}$ \quad
Yuhui Zhang$^{1}$ \quad
Zeyu Wang$^{1}$ \quad
Serena Yeung-Levy$^{1}$ \textsuperscript{\Letter}
\and 
$^{1}$Stanford University\quad
$^{2}$University of Strasbourg \quad
$^{3}$Technical University of Munich  \\
{\tt\small xxsun@stanford.edu, mli89@stanford.edu, syyeung@stanford.edu}
}

\newcommand{\cellbar}[2][normal]{%
    \begin{tikzpicture}[baseline=0.5ex, x=1cm, y=1cm]
        \pgfmathsetmacro{\barwidth}{#2/2.2}
        \shade[left color=blue!40, right color=blue!20, rounded corners=1pt] 
            (0,0) rectangle ({\barwidth},0.3);
        \pgfmathsetmacro{\centerpos}{0.5}
        \def\tempstyle{#1}%
        \def\boldstyle{bold}%
        \def\underlinestyle{underline}%
        \ifx\tempstyle\boldstyle
            \node at (\centerpos,0.15) {\textbf{\fontsize{7}{8.5}\selectfont #2}};
        \else
            \ifx\tempstyle\underlinestyle
                \node at (\centerpos,0.15) {\underline{\fontsize{7}{8.5}\selectfont #2}};   
            \else
                \node at (\centerpos,0.15) {\fontsize{7}{8.5}\selectfont #2};   
            \fi
        \fi
    \end{tikzpicture}%
}

\newcommand{\cellbarP}[2][normal]{%
    \def\maxlength{1}
    \begin{tikzpicture}[baseline=0.5ex, x=1cm, y=1cm]
        \pgfmathsetmacro{\barwidth}{#2*\maxlength/100}
        \shade[left color=gray!40, right color=gray!20, rounded corners=1pt] 
            (0,0) rectangle ({\barwidth},0.3);
        \pgfmathsetmacro{\centerpos}{\maxlength/2}
        \def\tempstyle{#1}%
        \def\boldstyle{bold}%
        \def\underlinestyle{underline}%
        \ifx\tempstyle\boldstyle
            \node at (\centerpos,0.15) {\textbf{\fontsize{7}{8.5}\selectfont #2}};
        \else
            \ifx\tempstyle\underlinestyle
                \node at (\centerpos,0.13) {\underline{\fontsize{7}{8.5}\selectfont #2}};
            \else
                \node at (\centerpos,0.15) {\fontsize{7}{8.5}\selectfont #2};
            \fi
        \fi
    \end{tikzpicture}%
}

\newcommand{\cellbarPB}[2][normal]{%
    \def\maxlength{1}
    \begin{tikzpicture}[baseline=0.5ex, x=1cm, y=1cm]
        \pgfmathsetmacro{\barwidth}{#2*\maxlength/100}
        \shade[left color=blue!40, right color=blue!20, rounded corners=1pt] 
            (0,0) rectangle ({\barwidth},0.3);
        \pgfmathsetmacro{\centerpos}{\maxlength/2}
        \def\tempstyle{#1}%
        \def\boldstyle{bold}%
        \def\underlinestyle{underline}%
        \ifx\tempstyle\boldstyle
            \node at (\centerpos,0.15) {\textbf{\fontsize{7}{8.5}\selectfont #2}};
        \else
            \ifx\tempstyle\underlinestyle
                \node at (\centerpos,0.13) {\underline{\fontsize{7}{8.5}\selectfont #2}};
            \else
                \node at (\centerpos,0.15) {\fontsize{7}{8.5}\selectfont #2};
            \fi            
        \fi
    \end{tikzpicture}%
}

\newcommand{\cellbarPos}[2][normal]{%
    \def\maxlength{1.2}
    \begin{tikzpicture}[baseline=0.5ex, x=1cm, y=1cm]
        \pgfmathsetmacro{\barwidth}{#2*\maxlength/100}
        \definecolor{myblue}{HTML}{4a5bc7}
        \shade[left color=myblue!80, right color=myblue!30, rounded corners=1pt]
            (0,0) rectangle ({\barwidth},0.3);
        \pgfmathsetmacro{\centerpos}{\maxlength/2}
        \def\tempstyle{#1}%
        \def\boldstyle{bold}%
        \def\underlinestyle{underline}%
        \ifx\tempstyle\boldstyle
            \node at (\centerpos,0.15) {\textbf{\fontsize{7}{8.5}\selectfont +#2}};
        \else
            \ifx\tempstyle\underlinestyle
                \node at (\centerpos,0.13) {\underline{\fontsize{7}{8.5}\selectfont +#2}};
            \else
                \node at (\centerpos,0.15) {\fontsize{7}{8.5}\selectfont +#2};
            \fi
        \fi
    \end{tikzpicture}%
}

\newcommand{\cellbarNeg}[2][normal]{%
    \def\maxlength{1.2}
    \begin{tikzpicture}[baseline=0.5ex, x=1cm, y=1cm]
        \pgfmathsetmacro{\barwidth}{abs(#2)*\maxlength/100}
        \definecolor{myred}{HTML}{bc5b4c}
        \shade[left color=myred!80, right color=myred!30, rounded corners=1pt]
            (0,0) rectangle ({\barwidth},0.3);
        \pgfmathsetmacro{\centerpos}{\maxlength/2}
        \def\tempstyle{#1}%
        \def\boldstyle{bold}%
        \def\underlinestyle{underline}%
        \ifx\tempstyle\boldstyle
            \node at (\centerpos,0.15) {\textbf{\fontsize{7}{8.5}\selectfont #2}};
        \else
            \ifx\tempstyle\underlinestyle
                \node at (\centerpos,0.13) {\underline{\fontsize{7}{8.5}\selectfont #2}};
            \else
                \node at (\centerpos,0.15) {\fontsize{7}{8.5}\selectfont #2};
            \fi
        \fi
    \end{tikzpicture}%
}

\begin{document}
\maketitle



\begin{abstract}
Large Vision-Language Models (VLMs) often answer classic visual illusions ``correctly'' on original images, yet persist with the same responses when illusion factors are inverted, even though the visual change is obvious to humans.
This raises a fundamental question: \emph{do VLMs perceive visual changes or merely recall memorized patterns?} While several studies have noted this phenomenon, the underlying causes remain unclear. 
To move from observations to systematic understanding, this paper introduces \textbf{VI-Probe}, a controllable visual-illusion framework with graded perturbations and matched visual controls (without illusion inducer) that disentangles visually grounded perception from language-driven recall. 
Unlike prior work that focuses on averaged accuracy, we measure stability and sensitivity using Polarity-Flip Consistency, Template Fixation Index, and an illusion multiplier normalized against matched controls.
Experiments across different families reveal that response persistence arises from heterogeneous causes rather than a single mechanism. For instance, GPT-5 exhibits memory override, Claude-Opus-4.1 shows perception-memory competition, while Qwen variants suggest visual-processing limits. Our findings challenge single-cause views and motivate probing-based evaluation that measures both knowledge and sensitivity to controlled visual change. Data and code are available at \href{https://sites.google.com/view/vi-probe/home}{VI-Probe Website}.
\end{abstract}

\section{Introduction}
\label{sec:intro}

\begin{figure}[ht]
\begin{center}
	\includegraphics[width=.98\linewidth]{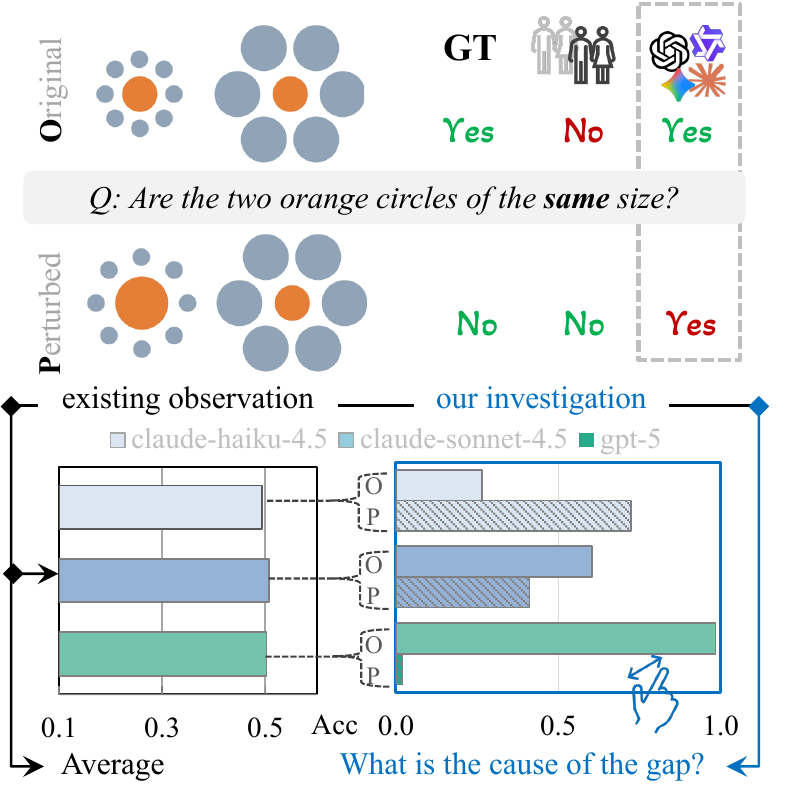}
\end{center}
\vspace{-6mm}
\caption{\textbf{Motivation for developing VI-Probe.} On classic visual illusions (\eg, Ebbinghaus~\cite{weintraub1979ebbinghaus}), large VLMs often score well on \textbf{O}\textcolor{gray}{riginal} images yet fail to \emph{flip} along with GT on \textbf{P}\textcolor{gray}{erturbed} images (factor inverted), even when the change is perceptually obvious to humans. Prior evaluations typically report \emph{average} accuracy and often do not expose the Original$\rightarrow$Perturbed gap or explain its cause. A few studies have noted this discrepancy, but without deeper analysis. We propose VI-Probe to \emph{decompose} model behavior and attribute the gap to potential contributing factors.
}
\label{fig:fig-motivation}
\vspace{-5pt}
\end{figure}

\begin{table*}[t]
\centering
\footnotesize
\setlength{\tabcolsep}{4pt}{
\begin{tabular}{l|ccc|cccc|cc|c}
\toprule
\multirow{2}{*}{\textbf{Dataset}} & \multirow{2}{*}{\textbf{\#Cases}} & \textbf{Original} & \textbf{Perturb} & \textbf{Visual} & \textbf{Visual} & \textbf{Language} & \textbf{Language} & \multicolumn{2}{c|}{\textbf{Metric}} & \multirow{2}{*}{\textbf{Purpose}} \\
 & & \textbf{\#Img} & \textbf{\#Img} & \textbf{Control} & \textbf{Hints} & \textbf{Control} & \textbf{Reverse} & \textbf{PFC} & \textbf{$R$} & \\
\midrule
HallusionBench~\cite{guan2024hallusionbench} & 15 & 72 & 72 & \ding{55} & \ding{55} & \ding{55} & \ding{55} & \ding{55} & \ding{55} & Illusion reasoning\\
$\text{IllusionBench+}^\dagger$~\cite{guan2024hallusionbench} & $\text{-}$ & 164 & 19 & \ding{55} & \ding{55} & \ding{55} & \ding{55} & \ding{55} & \ding{55} & - \\
IllusionVQA~\cite{shahgir2024illusionvqa} & 20 & 374 & 0 & \ding{55} & \ding{55} & \ding{55} & \ding{55} & \ding{55} & \ding{55} & Illusion recognition \\
VLMBiased~\cite{vo2025visionlanguagemodelsbiased} & 6 & 198 & 198 & \ding{55} & \ding{55} & \ding{55} & \ding{55} & \ding{55} & \ding{55} & VLMs bias evaluation \\
\midrule
\multirow{2}{*}{\textbf{Ours}} & \multirow{2}{*}{\textbf{27}} & \textbf{arbitrary} & \textbf{arbitrary} & \textbf{arbitrary} & \textbf{arbitrary } & \multirow{2}{*}{\ding{51}} & \multirow{2}{*}{\ding{51}} & \multirow{2}{*}{\ding{51}} & \multirow{2}{*}{\ding{51}} & \textbf{Controlled Fine-Grained} \\
&  & \textbf{(870)} & \textbf{(870$\times$10)} & \textbf{(870$\times$11)} & \textbf{(870$\times$11)} &  &  &  &  & \textbf{VLMs probing} \\
\bottomrule
\end{tabular}}
\vspace{-5pt}
\caption{\textbf{Dataset comparison}. Our benchmark provides comprehensive coverage with target control images, visual hints, and fine-grained illusion probing capabilities. ``arbitrary'' indicates that our benchmark allows for continuous generation of unlimited samples per category. Values under ``arbitrary''  are the parameters for the arbitrary used in our experiments. $^\dagger$ means dataset not fully publicly accessible.}
\label{tab:benchmark_comparison}
\vspace{-10pt}
\end{table*}

Human visual perception is not perfect~\cite{bach2006optical,binz2025foundation, carbon2014understanding, faisal2008noise}. Classic visual illusions such as the Ebbinghaus and M\"uller--Lyer~\cite{muller1889optische} effects show how easily human vision can be deceived, revealing fundamental limitations of biological perception~\cite{carbon2014understanding, bialek1987physical}. Interestingly, recent studies have found that large Vision–Language Models (VLMs) behave differently~\cite{guan2024hallusionbench, ullman2024illusion, vo2025visionlanguagemodelsbiased}. Models often provide physically correct answers that contradict human perception, creating the impression that they ``see through'' the illusion~\cite{zhang2023grounding,long2025understanding}. However, this observation is widely understood as a result of memorization rather than true perception~\cite{vo2025visionlanguagemodelsbiased,yang2025illusions}.
A closer examination reveals an asymmetry: as shown in Fig.~\ref{fig:fig-motivation}, when illusion factors are inverted, many models continue to give the same responses and ignore the visual change, even though humans readily perceive the difference. This contrast raises a fundamental question: \textit{do VLMs perceive visual changes, or do they merely recall memorized patterns?}

Several works~\cite{alavi2023large, vo2025visionlanguagemodelsbiased, guan2024hallusionbench, lee2025vlind, liu2025phd} have observed this phenomenon and concluded that VLMs rely more on language priors than visual evidence. While these studies describe \emph{what} happens, they offer limited insights into \emph{how} and \emph{why}. Moreover, existing evaluations rely on average accuracy, which obscures real differences~\cite{liu2025seeing, vo2025visionlanguagemodelsbiased}. Fig.~\ref{fig:fig-motivation} (bottom) shows mean results clustering near 50\% across models, yet paired comparisons reveal large gaps and rank reversals. Such averaging masks critical aspects of model behavior: stability under perturbations, sensitivity to visual factors, and conditions producing response persistence. The field still lacks a benchmark to probe why and how VLMs persist in giving the same answers despite clear visual changes.

To address this gap, we introduce \textbf{VI-Probe}, a controllable visual-illusion framework that disentangles visually grounded perception from language-driven priors through fine-grained manipulation of both visual and linguistic stimuli. We collect a diverse set of classic visual illusions involving size, color, and geometry, and introduce parametric controls that change, remove, or invert the illusory effect. This design enables precise measurement of model behavior under minimal perturbations. For each illusion case, we construct Original and Perturbed illusion images, paired with matched visual controls that remove illusion-inducing cues.
These controls allow us to isolate and quantify the impact of illusion patterns on model responses. 
On the language side, we design prompt variants to probe language priors and their interaction with visual evidence.

In addition, beyond standard accuracy, we introduce multiple metrics to capture stability and sensitivity from different dimensions: Polarity-Flip Consistency (PFC)~\cite{gardner2020evaluating} and Template Fixation Index (TFI) measure linguistic robustness, while an illusion multiplier ($R$) normalizes effects against matched controls to distinguish memory-driven from perception-driven behavior.

Through extensive experiments across multiple open- and closed-source VLM families (\eg GPT~\cite{achiam2023gpt}, Gemini~\cite{comanici2025gemini}, Claude~\cite{anthropic2025claude4}, Qwen~\cite{yang2025qwen3}), we find that the causes of response persistence vary across models, challenging the dominant explanation that attributes the phenomenon solely to language priors. For example, GPT-5 shows near-complete memory override, whereas Claude-Opus-4.1 exhibits perception-memory competition (dose-dependent illusion accuracy). In contrast, Qwen 2.5 variants reveal visual-processing limitations: representation entanglement and perception bottlenecks. Different model families also show distinct sensitivities to visual factors such as color, size, and spatial configuration, suggesting that response persistence arises from heterogeneous mechanisms in how models integrate or discount visual evidence.

In summary, our main contributions are:

\begin{itemize}
  \item We build \textbf{VI-Probe}, a controllable illusion-based framework that enables reproducible and fine-grained manipulation of visual and linguistic cues to probe VLMs.
  \item We introduce a perception–memory diagnostic paradigm that combines paired-prompt consistency (PFC, TFI) and illusion-normalized effect size ($R$), enabling factor-isolated evaluation beyond static accuracy.
  \item We reveal heterogeneous failure mechanisms across model families (memory override, perception-memory competition, visual-processing limits).
  \item We offer practical guidance for evaluation and model design by clarifying the balance between perception and memory in VLMs.
\end{itemize}


\section{Related Work}
\label{sec:related-work}

\begin{figure*}[htp]
\begin{center}
	\includegraphics[width=\linewidth]{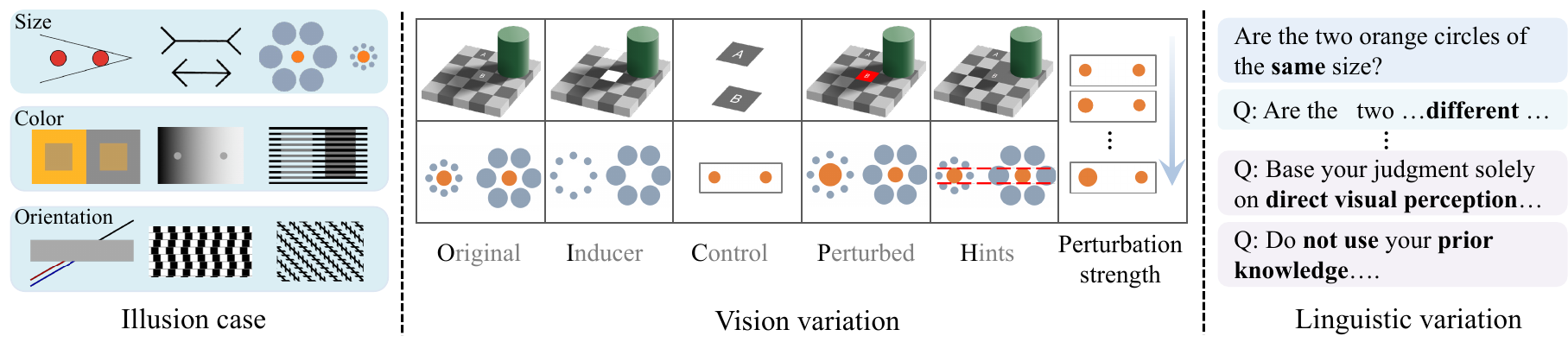}
\end{center}
\vspace{-15pt}
\caption{\textbf{Examples, vision and language variations in VI-Probe.} For each original illusion case, we have six versions of images (perturbed-control is not shown in the figure), in which perturbed, perturbed-control and perturbed-hints have a series of images of different perturbation strength. For language, we have three versions of prompting questions.}
\label{fig:fig-VI-Pro}
\vspace{-5pt}
\end{figure*}

\noindent\textbf{Visual Illusion Benchmarks for VLMs.}
Visual illusions have long served as diagnostic tools to reveal the limits and mechanisms of human perceptual processing~\cite{hartley1982roles}, and recent research has extended this approach to VLMs~\cite{guan2024hallusionbench, zhang2025illusionbench, shahgir2024illusionvqa, vo2025visionlanguagemodelsbiased, ullman2024illusion} (Table~\ref{tab:benchmark_comparison}), consistently showing that VLMs perform poorly compared to humans on illusion-based tasks. Despite this progress, existing benchmarks share three key limitations: (i) they use fixed rather than graded illusion strengths, preventing measurement of perceptual thresholds; (ii) evaluation metrics rely on binary accuracy, which masks model biases and confidence patterns; and (iii) they do not systematically disentangle visual perception from linguistic priors, leaving it unclear whether models are ``seeing'' or recalling. To address these gaps, we propose \textbf{VI-Probe}, a controllable benchmark built upon classic visual illusions for probing the perception–memory boundary in VLMs. As shown in Table~\ref{tab:benchmark_comparison}, VI-Probe provides comprehensive coverage with vision controls, visual hints, language variations, and fine-grained metrics (PFC and $R$) that enable factor-isolated evaluation beyond static accuracy.

\noindent\textbf{Bias in VLMs.}
Social~\cite{gallegos2024bias,blodgett2020language,nangia2020crows}, cultural~\cite{naous2024having,tao2024cultural,navigli2023biases}, and demographic biases have been extensively studied in LLMs~\cite{liu2023visual, radford2021learning, achiam2023gpt}. VLMs exhibit systematic biases~\cite{ghate2025biases,favero2024multi} arising from textual pretraining and visual-linguistic integration~\cite{janghorbani2023multimodal,deng2025words}. Models may over-rely on language priors, producing stable predictions despite visual changes~\cite{alavi2023large, vo2025visionlanguagemodelsbiased, leng2024mitigating}. This perceptual bias reveals a gap between perception and recall. Our work explores this through the lens of visual illusions, providing a controlled setting to probe how VLMs handle conflicting or ambiguous visual–linguistic cues.


\section{VI-Probe}

To systematically probe perception and memory in large VLMs, we develop \textbf{VI-Probe}, a controllable \textbf{V}isual \textbf{I}llusion evaluation framework that enables the creation of large-scale, reproducible illusion-language pairs with fine-grained control over both visual and linguistic dimensions. This design allows us to isolate the effect of specific perceptual cues and to test how models respond to controlled perturbations that alter or remove illusion-inducing elements. We design multiple question variants to explore how language influences model perception. Fig.~\ref{fig:fig-VI-Pro} shows examples in VI-Probe, with variations in vision and linguistic aspects.


\subsection{Controllable Illusion Generation Pipeline}
\label{sec:dataset}

\noindent\textbf{Illusion Cases.}
We curate 27 classical optical illusions from perception studies, covering size distortion (\textit{e.g.,} Ebbinghaus, M\"uller--Lyer), geometric misalignment (\textit{e.g.,} Poggendorff, Z\"ollner), and brightness or contrast effects (\textit{e.g.,} Checker-shadow, Mach bands).
We group them by their primary controlling factors into \emph{size and length}, \emph{brightness and color}, and \emph{orientation (parallelism)}. The list of names for the illusion cases is in the supplementary materials. Each illusion case can be manipulated along two independent axes: \textit{visual} and \textit{linguistic}. 

\begin{figure}[tp]
\begin{center}
	\includegraphics[width=0.9\linewidth]{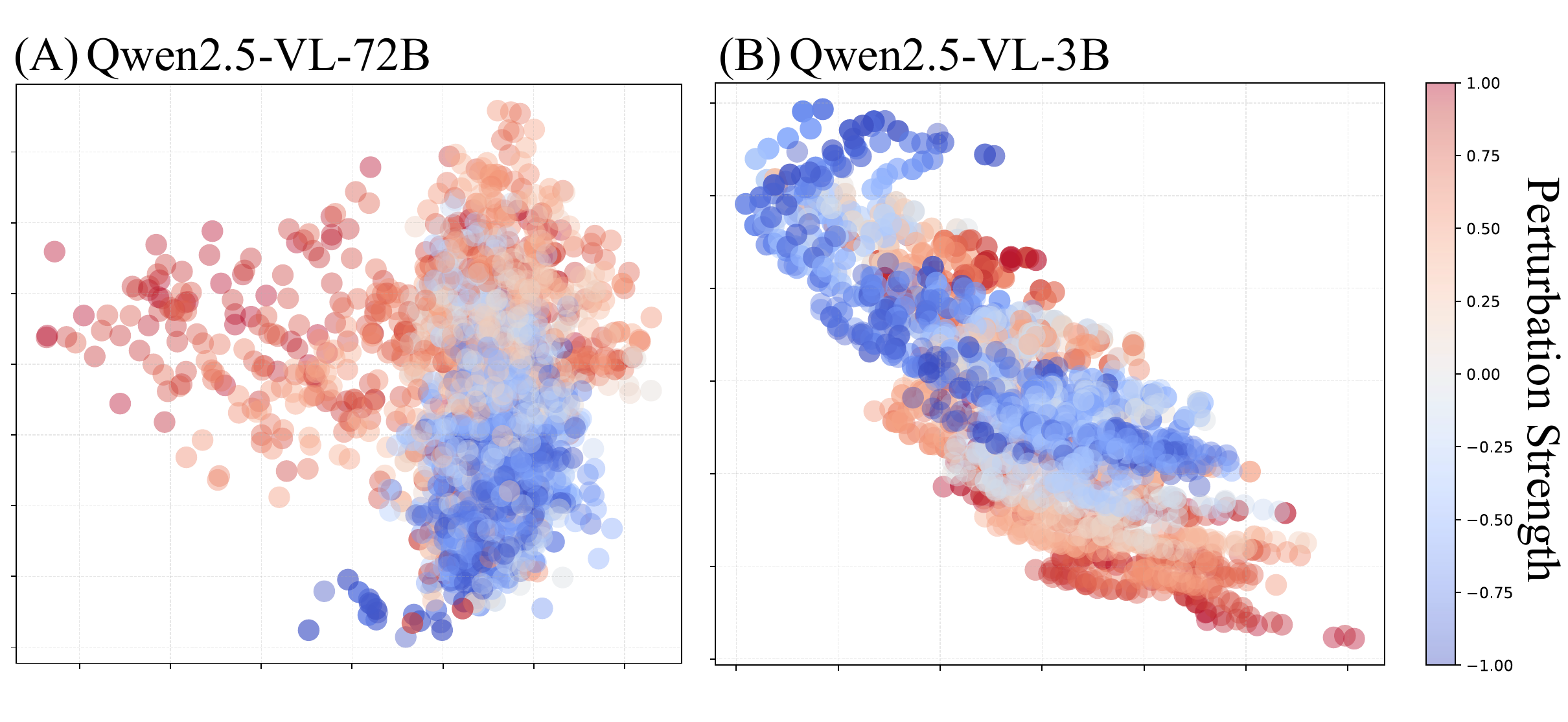}
\end{center}
\vspace{-15pt}
\caption{\textbf{Distribution of perturbation images at different strengths.} 
2D PCA visualization of embeddings from (A) Qwen2.5-VL-72B and (B) -3B, with color coding for perturbation strength. Left shows clear separation between different perturbation levels, indicating the validity of the data generation pipeline. 
}
\label{fig:fig-v-feature}
\vspace{-5pt}
\end{figure}

\noindent\textbf{Vision Variation.}
For each illusion case, we use the classic image as a seed and manipulate unrelated factors (\textit{e.g.,} color in size illusions) to create additional original images that incorporate the illusion. For each \textbf{original} image, we generate (1) a set of \textbf{perturbed} images where control factors (such as size ratio, line length, local contrast, or orientation) are inverted at graded levels $(\alpha)$, (2) matched \textbf{visual controls} where the \textbf{inducers} are removed, and (3) \textbf{hinted} versions with overlay cues (Fig.~\ref{fig:fig-VI-Pro}). These variants enable well-matched, controlled comparisons that isolate the illusion effect and measure model sensitivity to minimal but meaningful visual changes. Fig.~\ref{fig:fig-v-feature} shows that the generated samples form structured distributions in the embedding space of Qwen2.5-VL-72B and 3B, confirming that perturbations provide continuous control over illusion intensity.

\noindent\textbf{Linguistic Variation.} 
We introduce linguistic perturbations to test how language changes affect model responses. Two types of textual edits are implemented: (1) \textbf{Reversed questions}, which flip the logical polarity (\textit{e.g.,} ``Are the lines the same length?" vs. ``Are the lines different in length?"), and (2) \textbf{Instructional variants} that add explicit system instructions to disregard prior knowledge and focus on visual judgment. Selected instructions are shown below with full instructions in the Supplementary Materials.

\begin{tcolorbox}[
    colback=gray!3,           
    colframe=gray!40,          
    rounded corners,           
    width=\linewidth,      
    arc=2mm,                   
    boxrule=0.5pt,            
    left=3pt, right=3pt,      
    top=2pt, bottom=2pt,       
]
{\noindent\footnotesize\texttt{\noindent Visual Comparison Instructions:\\
Base your judgment exclusively on direct visual perception of the image. Compare the two targets systematically using only what is visible in the image itself...}
}
\end{tcolorbox}

\subsection{Dataset schema and notation}


\noindent\textbf{Vision variants.}
For each illusion case we collect a base collection of images $\bm{x}$ that consists of an \emph{original} illusion image $x^{\mathrm{O}}$ or a \emph{perturbed} illusion image $x^{\mathrm{P}}$.
We also provide a \emph{visual control} $x^{\mathrm{OC}}$/$x^{\mathrm{PC}}$ that removes the illusion pattern while keeping global semantics, and an optional \emph{hinted} version $x^{\mathrm{OH}}$/$x^{\mathrm{PH}}$ that overlays minimal visual hints.

\noindent\textbf{Linguistic variants.}
Each image is paired with a \emph{forward} question $q^{\mathrm{f}}$ (\textit{e.g.,} ``Are the two targets the same?''), its \emph{reversed} polarity $q^{\mathrm{r}}$ (\textit{e.g.,} ``Are the two targets different?''), and an \emph{instructional} variant $q^{\mathrm{I}}$ that adds explicit comparison instructions while preserving polarity.
Let $\mathrm{pol}(q)\in\{+1,-1\}$ denote the question polarity, where $\mathrm{pol}(q^{\mathrm{r}})=-\mathrm{pol}(q^{\mathrm{f}})$ and $\mathrm{pol}(q^{\mathrm{I}})=\mathrm{pol}(q^{\mathrm{f}})$.
For any image $x\in\{x^{\mathrm{O}},x^{\mathrm{P}},x^{\mathrm{OC}},x^{\mathrm{OH}},x^{\mathrm{PC}},x^{\mathrm{PH}}\}$ and question $q\in\{q^{\mathrm{f}},q^{\mathrm{r}},q^{\mathrm{I}}\}$ we define binary labels:
\[
y^{\mathrm{f}}(x)\in\{0,1\}, \quad
y^{\mathrm{r}}(x)=1-y^{\mathrm{f}}(x), \quad
y^{\mathrm{I}}(x)=y^{\mathrm{f}}(x).
\]
When needed we index perturbation strength by $\alpha\in\mathcal{A}$ and write $x^{\mathrm{P}}_{\alpha}$, $x^{\mathrm{PC}}_{\alpha}$ and $x^{\mathrm{PH}}_{\alpha}$.

\subsection{Metrics Beyond Accuracy}
\label{sec: Metrics}

\noindent\textbf{Paraphrase–pair consistency (same vs.\ different).}
For each image we ask a complementary pair of questions, $q^{\mathrm{f}}$ (\textit{e.g.,} ``same?") and $q^{\mathrm{r}}$ (\textit{e.g.,} ``different?"), with labels $y^{\mathrm{r}} = 1 - y^{\mathrm{f}}$.
Let $a^{\mathrm{f}}, a^{\mathrm{r}} \in \{0,1\}$ be the answers of the model. We compute three metrics:

1) Polarity-Flip Consistency (PFC) measures whether the two answers are complements, regardless of correctness: $\mathrm{PFC} = \mathbb{E}[\mathds{1}(a^{\mathrm{r}} = 1 - a^{\mathrm{f}})]$.
2) Polarity-Flip Accuracy (PFA) further requires both answers to be correct: $\mathrm{PFA} = \mathbb{E}[\mathds{1}(a^{\mathrm{f}} = y^{\mathrm{f}} \land a^{\mathrm{r}} = y^{\mathrm{r}})]$.
3) Template Fixation Index (TFI) is the fraction of pairs where the model repeats the same polarity: $\mathrm{TFI} = \mathbb{E}[\mathds{1}(a^{\mathrm{r}} = a^{\mathrm{f}})]$.
%
High PFC with low accuracy indicates linguistically coherent but visually wrong behavior, whereas high TFI signals polarity-insensitive repetition. We define \emph{Coherent but Wrong (CbW)} as $\mathrm{CbW} = \mathrm{PFC} - \mathrm{PFA}$, which quantifies cases where the model provides complementary answers (linguistically coherent) but both are incorrect (visually wrong).
These paired metrics disentangle linguistic polarity handling from visual judgment and complement image-level accuracy.

\noindent\textbf{Illusion multiplier.}
We define illusion multiplier (IM) as the ratio of the illusion effect size to the control effect size, normalizing the impact of illusion against baseline perturbation effects:
\begin{equation}
R= \frac{\left|\mathbb{E}_{(x^\text{O}\text{,}x^\text{P})\sim \bm{x}}\bigl[\text{Acc}(x^{\mathrm{O}}) - \text{Acc}(x^{\mathrm{P}})\bigr]\right|}{\left|\mathbb{E}_{(x^{\text{OC}}\text{,}x^{\text{PC}})\sim \bm{x}}\bigl[\text{Acc}(x^{\mathrm{OC}}) - \text{Acc}(x^{\mathrm{PC}})\bigr]\right|+\epsilon},
\label{eq:Ra}
\end{equation}
where $\epsilon{=}0.001$ prevents division by zero.
$R{>}1$ indicates models are more susceptible to illusion patterns than general perturbations; $R{<}1$ indicates that perturbations have less impact when illusion context is present, typically when baseline visual processing is weak; $R{\approx}1$ means illusion and control perturbations have similar effects, suggesting visual and memory signals compete rather than one fully dominating. By normalizing against matched controls, $R$ isolates the specific contribution of illusion patterns and helps distinguish perceptual limitations from memory-driven biases.

\section{Experiments}
\label{sec:evaluation}

\subsection{Experimental Setup}

\noindent\textbf{Model Families.}
We evaluate 15 vision-language models spanning four families, covering both the latest closed-source and open-source models. We report results for four main families in the paper, with additional analyses in the supplementary materials:

\begin{itemize}
  \item OpenAI~\cite{singh2025openai} \includegraphics[width=0.3cm,valign=c]{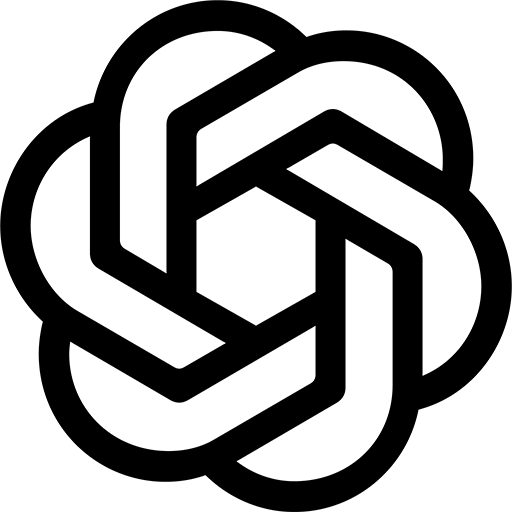}: GPT-5, GPT-5-Mini, and GPT-5-Nano.
  \item Anthropic~\cite{anthropic2025claude4} \includegraphics[width=0.3cm,valign=c]{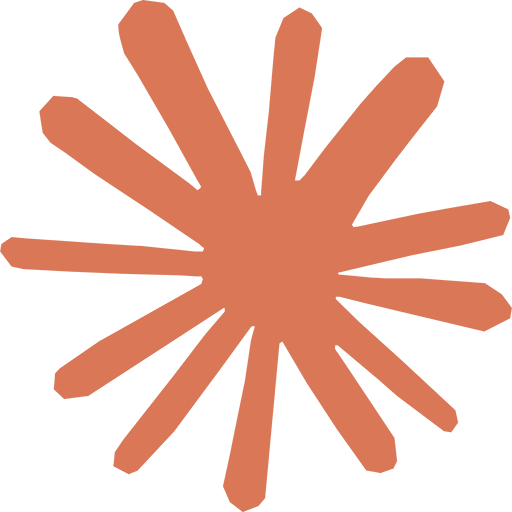}:   Claude-Opus-4.1, Claude-Sonnet-4.5, and Claude-Haiku-4.5 
  \item  Google~\cite{comanici2025gemini} \includegraphics[width=0.35cm,valign=c]{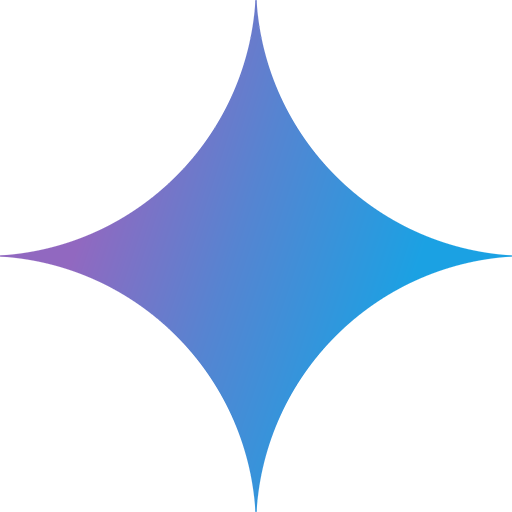}:    Gemini-2.5-Flash and -2.5-Flash-Lite.
  \item Qwen3-VL and Qwen2.5-VL series~\cite{yang2025qwen3} \includegraphics[width=0.3cm,valign=c]{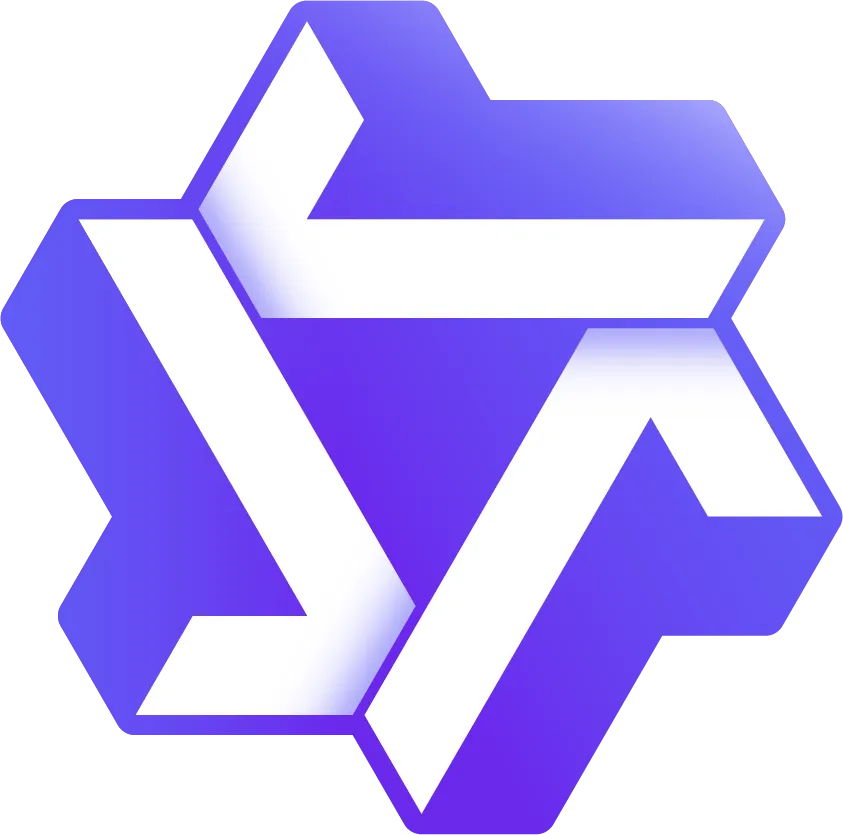}: Qwen3-VL-235B-A22B, Qwen3-VL-32B, and Qwen3-VL-8B; Qwen2.5-VL-72B, Qwen2.5-VL-32B, Qwen2.5-VL-7B, and Qwen2.5-VL-3B.
\end{itemize}

All models are accessed via APIs\footnote{We use OpenRouter (\url{https://openrouter.ai/}) for evaluation. For models without temperature control, we use the default settings.} to ensure uniform inference conditions. Each model is prompted in a unified zero-shot setting. 
We will release images, code, and generation scripts, together with prompts and scoring tools, to ensure full reproducibility and facilitate follow-up research.

\subsection{Analysis of Visual Perception \textit{vs.} Memory}
Our goal is to probe two key aspects: (1) whether VLMs \emph{flip} their predictions when visual evidence contradicts prior knowledge (flip sensitivity), and (2) whether their responses remain unchanged across visual and prompt variations.

\subsubsection{Polarity-Flip Consistency for Opposite Questions}
\noindent\textbf{Setup:} As described in Sec.~\ref{sec: Metrics}, for each image we ask a complementary pair of questions\footnote{Complementary questions ask the same fact with opposite polarity.
} (Fig.~\ref{fig:fig-VI-Pro}, Linguistic Variation), and then disentangle linguistic polarity handling from visual judgment based on paired responses. For this experiment, the image types include the original, the perturbed, their controls, and the perturbed-with-hint.

\begin{figure}[t]
\centering
\includegraphics[width=\linewidth]{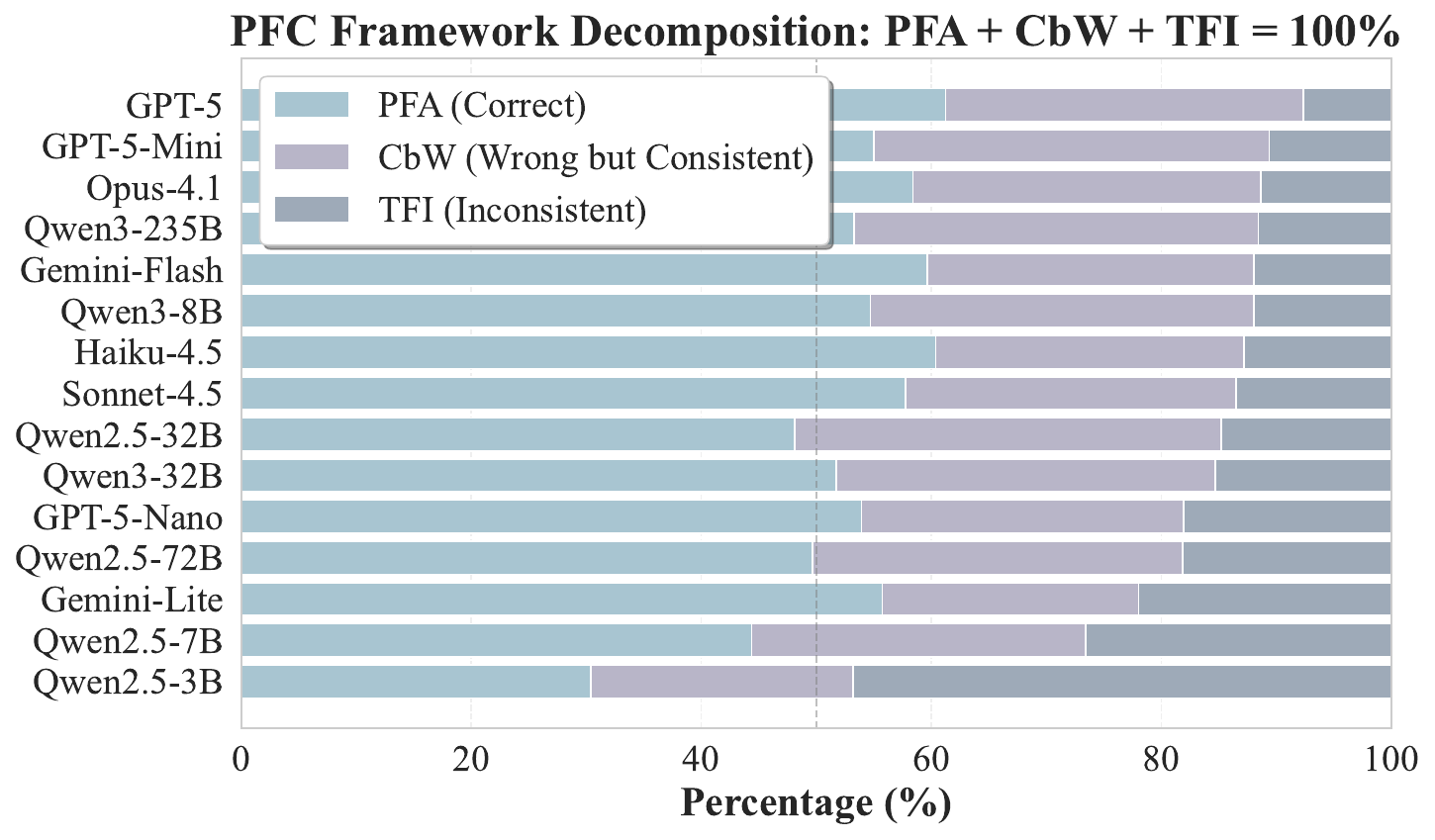}
\caption{\textbf{Polarity–Flip Consistency (PFC) decomposition across VLMs.}
Each stacked bar partitions paired responses into:
\textbf{PFA} = correct \emph{and} complementary (both answers correct),
\textbf{CbW} = complementary but not fully correct,
and \textbf{TFI} = non–complementary.
Models are ordered by overall consistency, $\mathrm{PFC}=\mathrm{PFA}+\mathrm{CbW}$; the dashed line at $50\%$ marks random.
}
\label{fig:pfc_decomposition}
\vspace{-1.5em}
\end{figure}

\noindent\textbf{Result:} We find that high consistency does not guarantee high accuracy. Models can keep complementary answers across reversed prompts yet be wrong. Fig.~\ref{fig:pfc_decomposition} decomposes responses into PFA (both correct), CbW (complementary but wrong) and TFI (same polarity on two prompts).

\textbf{First, high PFC with high CbW reveals systematic visual errors masked by linguistic coherence.} 
OpenAI models achieve among the highest PFC (GPT-5: 92.32\%, GPT-5-Mini: 89.38\%) yet also show large CbW (GPT-5-Mini: 34.37\%, GPT-5: 31.08\%, GPT-5-Nano: 28.02\%). This means these models often flip answers consistently with polarity changes, but both answers can still be wrong. For example, GPT-5 achieves 92.32\% PFC but only 61.24\% PFA, yielding 31.08\% CbW. In contrast, Claude models are relatively more balanced (PFC: 86.49–88.61\%, CbW: 26.78–30.21\%), while Google models show mixed behavior (Gemini-2.5-Flash: PFC 88.06\%, CbW 28.40\%; Gemini-2.5-Flash-Lite: PFC 78.00\%, CbW 22.28\%).

\textbf{Second, model scale does not monotonically reduce linguistic fixation or visual bias.}
Within Qwen, scaling is non-monotonic: Qwen3-VL-8B reaches 88.01\% PFC, above Qwen3-VL-32B (84.70\%) and close to Qwen3-VL-235B (88.43\%). Qwen2.5-VL-32B (85.20\%) also exceeds the larger Qwen2.5-VL-72B (81.84\%). Yet Qwen2.5-VL-32B still has high CbW (37.10\%), and Qwen2.5-VL-3B shows strong template fixation (TFI: 46.82\%). Overall, parameter count alone is not decisive.

\textbf{Third, small models fail to maintain linguistic invariance, limiting reliable visual reasoning.}
Qwen2.5-VL-3B has TFI = 46.82\%, meaning nearly half of paired responses keep the same polarity for opposite questions (e.g., answering ``No'' to both ``Are they the same?'' and ``Are they different?''). This polarity collapse indicates weak semantic handling before visual judgment, making downstream accuracy less reliable.

\noindent
\begin{tcolorbox}[
    colback=gray!10,           
    colframe=gray!40,          
    rounded corners,           
    width=\linewidth,      
    arc=2mm,                   
    boxrule=0.5pt,            
    left=3pt, right=3pt,      
    top=2pt, bottom=2pt       
]
\textbf{Takeaways}
\begin{itemize}
    \item Linguistic robustness is a prerequisite. Models with high TFI (\eg, $>$45\%) cannot reliably perform visual reasoning regardless of raw accuracy.
    \item PFC serves as a quality threshold: small models \textit{e.g.,} Qwen2.5-VL-3B-Instruct show high TFI ($>$ 45\%), indicating insufficient linguistic robustness for further reliable visual analysis.
\end{itemize}
\end{tcolorbox}

\begin{table*}[ht]
\scriptsize
\centering
\setlength{\tabcolsep}{4pt}{
\begin{tabular}{ll|c|ccc|ccc|ccc |c}
\toprule
\multirow{2}{*}{Group} & \multirow{2}{*}{Model} & \multirow{2}{*}{PFC} & \multicolumn{3}{c|}{Illusion Pattern (\%)} & \multicolumn{3}{c|}{Control Pattern (\%)} & \multicolumn{3}{c|}{Illusion Effect (\%)} & \multirow{2}{*}{$R$}   \\
      &       &       & Original & Perturbed & Ave. & Original  & Perturbed  & Ave. & $\Delta$Original & $\Delta$Perturbed & $\Delta$Ave. &  \\
\midrule
\multirow{3}{*}{\includegraphics[width=0.4cm,valign=c]{fig/logo/openai.png}} 
& GPT-5 & \cellbarP[underline]{82.51} & \textbf{91.72} & 4.45 & \cellbarPB{48.09} & \textbf{96.55} & 52.24 & \cellbarP[bold]{74.40} & \textbf{-4.83} & -47.79 & \cellbarP[bold]{26.31} & \hspace{0.1cm}\cellbar{1.97}\\
& GPT-5-Mini & \cellbarP[bold]{84.86} & \underline{87.24} & 8.97 & \cellbarPB[underline]{48.10} & 93.45 & 30.38 & \cellbarP{61.91} & \underline{-6.21} & -21.41 & \cellbarP{13.81} & \cellbar{1.24}\\
& GPT-5-Nano & \cellbarP{65.64} & 46.21 & \underline{47.41} & \cellbarPB{46.81} & 55.86 & \textbf{66.14} & \cellbarP{61.00} & -9.66 & -18.72 & \cellbarP{14.19} & \cellbar{0.12}\\
\cline{1-13}
\multirow{3}{*}{\includegraphics[width=0.4cm,valign=c]{fig/logo/anthropic.png}} 
& Claude-Opus-4.1 & \cellbarP{72.68} & 67.93 & 27.55 & \cellbarPB{47.74} & 88.97 & 49.17 & \cellbarP{69.07} & -21.03 & -21.62 & \cellbarP{21.33} & \cellbar{1.01} \\
& Claude-Sonnet-4.5 & \cellbarP{77.65} & 70.69 & 26.17 & \cellbarPB{48.43} & 88.97 & 45.10 & \cellbarP{67.03} & -18.28 & -18.93 & \cellbarP{18.60} & \cellbar{1.01}\\
& Claude-Haiku-4.5 & \cellbarP{68.59} & 45.52 & \textbf{50.66} & \cellbarPB{48.09} & 83.79 & \underline{61.55} & \cellbarP[underline]{72.67} & -38.28 & -10.90 & \cellbarP[underline]{24.59} & \cellbar{0.23}\\
\cline{1-13}
\multirow{2}{*}{\includegraphics[width=0.4cm,valign=c]{fig/logo/gemini.png}} 
& Gemini-2.5-Flash & \cellbarP{77.66} & 75.52 & 20.90 & \cellbarPB[bold]{48.21} & 92.07 & 50.17 & \cellbarP{71.12} & -16.55 & -29.28 & \cellbarP{22.91} & \cellbar{1.30}\\
& Gemini-2.5-Flash-Lite & \cellbarP{63.25} & 52.07 & 42.00 & \cellbarPB{47.03} & 82.07 & 46.72 & \cellbarP{64.40} & -30.00 & -4.72 & \cellbarP{17.36} & \cellbar{0.28}\\
\cline{1-13}
\multirow{7}{*}{\includegraphics[width=0.4cm,valign=c]{fig/logo/qwen.png}} 
& Qwen3-VL-235B-A22B & \cellbarP{75.31} & 70.69 & 23.14 & \cellbarPB{46.91} & 88.97 & 30.34 & \cellbarP{59.66} & -18.28 & -7.21 & \cellbarP{12.74} & \cellbar{0.81}\\
& Qwen3-VL-32B & \cellbarP{79.15} & 77.93 & 15.55 & \cellbarPB{46.74} & 89.31 & 24.17 & \cellbarP{56.74} & -11.38 & -8.62 & \cellbarP{10.00} & \cellbar{0.96}\\
& Qwen3-VL-8B & \cellbarP{68.16} & 63.45 & 32.72 & \cellbarPB{48.09} & \underline{93.79} & 28.79 & \cellbarP{61.29} & -30.34 & \underline{3.93} & \cellbarP{13.21} & \cellbar{0.47}\\
& Qwen2.5-VL-72B & \cellbarP{81.46} & 65.17 & 17.83 & \cellbarPB{41.50} & 91.72 & 23.83 & \cellbarP{57.78} & -26.55 & -6.00 & \cellbarP{16.28} & \cellbar{0.70}\\
& Qwen2.5-VL-32B & \cellbarP{74.15} & 61.03 & 23.93 & \cellbarPB{42.48} & 88.28 & 19.14 & \cellbarP{53.71} & -27.24 & \textbf{4.79} & \cellbarP{11.22} & \cellbar{0.54}\\
& Qwen2.5-VL-7B & \cellbarP{65.99} & 55.17 & 17.83 & \cellbarPB{36.50} & 83.79 & 20.69 & \cellbarP{52.24} & -28.62 & -2.86 & \cellbarP{15.74} & \cellbar{0.59}\\
& Qwen2.5-VL-3B & \cellbarP{56.15} & 22.41 & 14.07 & \cellbarPB{18.24} & 74.48 & 10.72 & \cellbarP{42.60} & -52.07 & 3.34 & \cellbarP{24.36} & \cellbar{0.13}\\
\bottomrule
\end{tabular}}
\vspace{-2mm}
\caption{\textbf{Isolating illusion effects with matched controls.} For each model, we report (i) pairwise consistency (PFC), (ii) accuracy on \emph{Illusion} images (Original \textit{vs.}\ Perturbed) and their matched \emph{Control} images, (iii) the corresponding illusion effects (drops relative to controls), and (iv) the \emph{illusion multiplier} $R$.
Image-level averages can hover near $50\%$ and mask large Original$\rightarrow$Perturbed cliffs; $R$ reveals family–specific mechanisms and distinguishes memory–driven degradation from pure visual difficulty.}
\label{tab:model_results_aggregated}
\vspace{-2em}
\end{table*}


\subsubsection{Isolating Memory Effects from Visual Perception}
\label{sec:main-results}

\noindent\textbf{Setup:} 
We evaluate four image types per case: Illusion–Original, Illusion–Perturbed, Control–Original, and Control–Perturbed. Matched controls remove illusion patterns while preserving global layout, isolating the net effect of illusion-specific cues. We use $R$ to normalize the Original$\rightarrow$Perturbed drop on illusions by the corresponding drop on controls, revealing whether degradation is driven by memory/recall or visual complexity. Accuracies reported here and in the following sections are all based on PFA.

\noindent\textbf{Result:} We find that image-level averaging masks critical sensitivity patterns, and that $R$ reveals family-specific mechanisms behind response persistence. Table~\ref{tab:model_results_aggregated} shows the evaluation results. PFC in Table~\ref{tab:model_results_aggregated} is computed on the four image types used in this section only.

\textbf{First, image-level averaging near 50\% conceals extreme Original$\rightarrow$Perturbed cliffs.} Illusion averages cluster at 46--48\%, appearing near-random. However, disaggregating by condition exposes sharp asymmetries. GPT-5 achieves 91.72\% on original illusions but collapses to 4.45\% when cues invert (drop: 87.27\%); GPT-5-Mini shows similar patterns (87.24\%$\rightarrow$8.97\%, drop: 78.27\%). Matched controls show smaller drops: GPT-5 falls from 96.55\% to 52.24\% (44.31\%), less than half the illusion drop. This divergence reveals models answer classic illusions via memorized patterns rather than visual perception, failing to update when visual evidence changes.

\textbf{Second, $R$ quantifies the relative contribution of memory versus visual processing.} $R>1$ indicates memory override: illusion drops exceed control drops. GPT-5 ($R{=}1.97$) and Gemini-2.5-Flash ($R{=}1.30$) show illusion drops 1.97$\times$ and 1.30$\times$ larger than controls, revealing prior knowledge interferes with visual input. $R<1$ indicates visual processing bottlenecks. Qwen models uniformly show $R<1$; Qwen2.5-3B ($R{=}0.13$) exhibits nearly identical degradation on illusions and controls, suggesting weak foundational vision rather than illusion-specific recall. Claude-Opus-4.1 and Sonnet-4.5 ($R{\approx}1.01$) show perception–memory competition where neither fully dominates.

\textbf{Third, some small models outperform larger counterparts on visual perception despite lower illusion accuracy.} Claude-Haiku-4.5 reaches 72.67\% on controls versus Claude-Opus-4.1 (69.07\%) and Claude-Sonnet-4.5 (67.03\%), and 61.55\% on perturbed controls versus Opus's 49.17\%. Uniquely, Haiku-4.5's perturbed-illusion accuracy (50.66\%) exceeds original-illusion (45.52\%), showing reliance on current input over memory. This inverted pattern ($R{=}0.23$) confirms perception-first processing.

\textbf{Fourth, flagship models show the strongest memory reliance, contradicting the assumption that scale improves visual grounding.} Among frontier models, GPT-5 ($R{=}1.97$), Gemini-2.5-Flash ($R{=}1.30$), and GPT-5-Mini ($R{=}1.24$) all exhibit $R{>}1.2$, indicating that increased model capability amplifies prior-driven behavior rather than mitigating it. In contrast, GPT-5-Nano ($R{=}0.12$) behaves like a perception-limited model, suggesting that the perception-memory trade-off is driven by architectural or training differences rather than parameter count alone.

\vspace{1mm}
\noindent
\begin{tcolorbox}[
    colback=gray!10,           
    colframe=gray!40,          
    rounded corners,           
    width=\linewidth,      
    arc=2mm,                   
    boxrule=0.5pt,            
    left=3pt, right=3pt,      
    top=2pt, bottom=2pt       
]
\textbf{Takeaways}
	\vspace{-0mm}
	\begin{itemize}
		\item Models from different families exhibit different mechanisms under illusions, even when average accuracy is similar (\textit{e.g.}, OpenAI \textit{vs.} Anthropic). 
		\vspace{-0mm}
		\item Memory reliance scales with capability. Advanced models exhibit stronger prior-driven behavior, requiring counterfactual-aware training to balance perception and recall.
        \item  Some smaller models outperform their larger counterparts (Claude-Haiku-4.5 \textit{vs.} Claude-Opus-4.1).
		\vspace{-0mm} 
  \vspace{-0mm}
\end{itemize}
\end{tcolorbox}

\begin{figure*}[t]
\centering
\includegraphics[width=\linewidth]{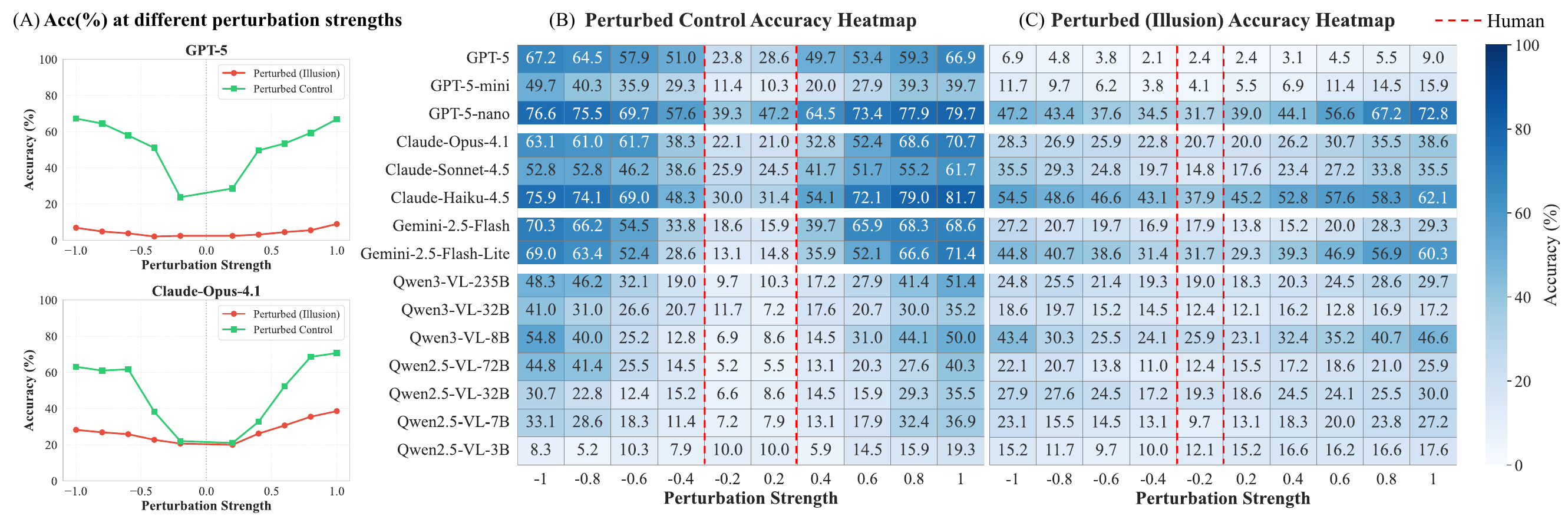}
\vspace{-10pt}
\caption{\textbf{Model performance under controlled perturbations.}
(A) Accuracy \textit{vs.}\ perturbation strength for two representative models (GPT-5 and Claude-Opus-4.1) under
\emph{perturbed–control} (vision-only) and \emph{perturbed–illusion} (factor-inverted) settings; the gap between the two curves reflects memory \textit{vs.}\ vision. 
(B) Heatmap for perturbed–control; (C) heatmap for perturbed–illusion.
Columns denote perturbation strength (larger value is stronger).
Accuracy typically decreases with strength, yet rankings diverge across (B) and (C). \textcolor{red}{Red} vertical lines indicate the threshold at which human observers reliably detect visual changes.
}
\label{fig:p_strenth}
\vspace{-8pt}
\end{figure*}

\subsubsection{Perturbation Strength \textit{vs.} Anti-Illusion}
\label{sec:Perturbation Strength}

\noindent\textbf{Setup:} We vary perturbation strength at 10 levels on two matched conditions: \emph{Perturbed-Control} (inducers removed) and \emph{Perturbed-Illusion} (inducers present), isolating visual complexity from illusion-specific interference.  We also collect human judgments on a subset of stimuli to establish perceptual baselines for comparison.

\noindent\textbf{Result:} Fig.~\ref{fig:p_strenth} shows \textbf{(A)} dose-response curves for GPT-5 and Opus-4.1, \textbf{(B--C)} heatmaps for all 15 models under control and illusion conditions at varying $\alpha$.

\textbf{First, GPT-5 exhibits complete memory override, while Opus-4.1 shows perception–memory competition.} Panel (A) exposes two failure modes. The control accuracy of GPT-5 declines 70\%→25\% as $\alpha$ increases, but illusion accuracy stays frozen at 0--5\%, indicating complete memory override. The illusion curve of Claude-Opus-4.1 retains dose-dependence (22\%→40\%) though dampened versus controls (70\%→20\%). Gaps quantify interference: GPT-5 constant $\sim$62$\%$, Opus-4.1 variable 30--50$\%$.

\textbf{Second, rankings reverse completely between control and illusion conditions.} Panels (B) and (C) at $\alpha{=}+1.0$ show systematic reordering. Top control performers (GPT-5: 2nd, GPT-5-Mini: 3rd, Opus-4.1: 4th) collapse under illusions (15th, 14th, 11th), while mid-tier models (Haiku-4.5: 9th, GPT-5-Nano: 8th) rise to top-3. This confirms orthogonal competencies: semantic integration aids noise handling but triggers template retrieval under illusions; low-level processing lacks strong priors to override.

\textbf{Third, families exhibit distinct threshold behaviors.} Panel (C) shows at $\alpha{=}\pm0.5$ low within-family variance (clustering at 40--60\%), but sharp divergence at $\alpha{=}\pm1.0$. OpenAI flagships collapse (45--55\% → 2--8\%), while GPT-5-Nano degrades gradually (55\%→38\%). Anthropic models spread widely (Haiku: 52\%, Sonnet: 28\%, Opus: 18\%). Qwen families split by generation: Qwen2.5 collapses sharply, Qwen3 degrades smoothly, suggesting training corpus updates outweigh architecture scaling. 

\textbf{Fourth, human perceptual thresholds reveal a fundamental model–human dissociation.} Red vertical lines in panels (B--C) mark the perturbation strength at which human observers reliably detect visual changes ($\sim$95\% detection rate). Most VLMs fail catastrophically \emph{before} this human threshold under illusion conditions (panel C), yet maintain reasonable accuracy under controls (panel B) at identical perturbation magnitudes. This dissociation confirms failure stems from illusion-triggered template retrieval rather than insufficient perceptual capacity. The visual signal suffices for both humans and models under controls, yet models should flip predictions well before the human threshold but many persist far beyond it.

\vspace{1mm}
\noindent
\begin{tcolorbox}[
    colback=gray!10,           
    colframe=gray!40,          
    rounded corners,           
    width=\linewidth,      
    arc=2mm,                   
    boxrule=0.5pt,            
    left=3pt, right=3pt,      
    top=2pt, bottom=2pt       
]
\textbf{Takeaways}
	\vspace{-0mm}
	\begin{itemize}
\item \textbf{Dose–response curves diagnose mechanism.} Flat illusion curves (GPT-5: constant 0--5\%) signal complete memory override; variable gaps (Opus: 30--50\%)) indicate perception–memory competition.
		\vspace{-0mm}
  \item \textbf{Rankings reverse completely.} Models rank 2nd--4th on control yet 11th--15th on illusions, confirming anti-noise and anti-illusion are orthogonal skills requiring separate evaluation.
  \item \textbf{Model–human dissociation.} VLMs fail before human perceptual thresholds under illusions but some models can match or exceed humans under controls, confirming failure stems from template retrieval rather than perceptual limits.
		\vspace{-0mm} 
  \vspace{-0mm}
\end{itemize}
\end{tcolorbox}

\subsubsection{Illusion Categories and Model Interventions}
\label{sec:Interventions}

\begin{figure}[t]
\centering
\includegraphics[width=\linewidth]{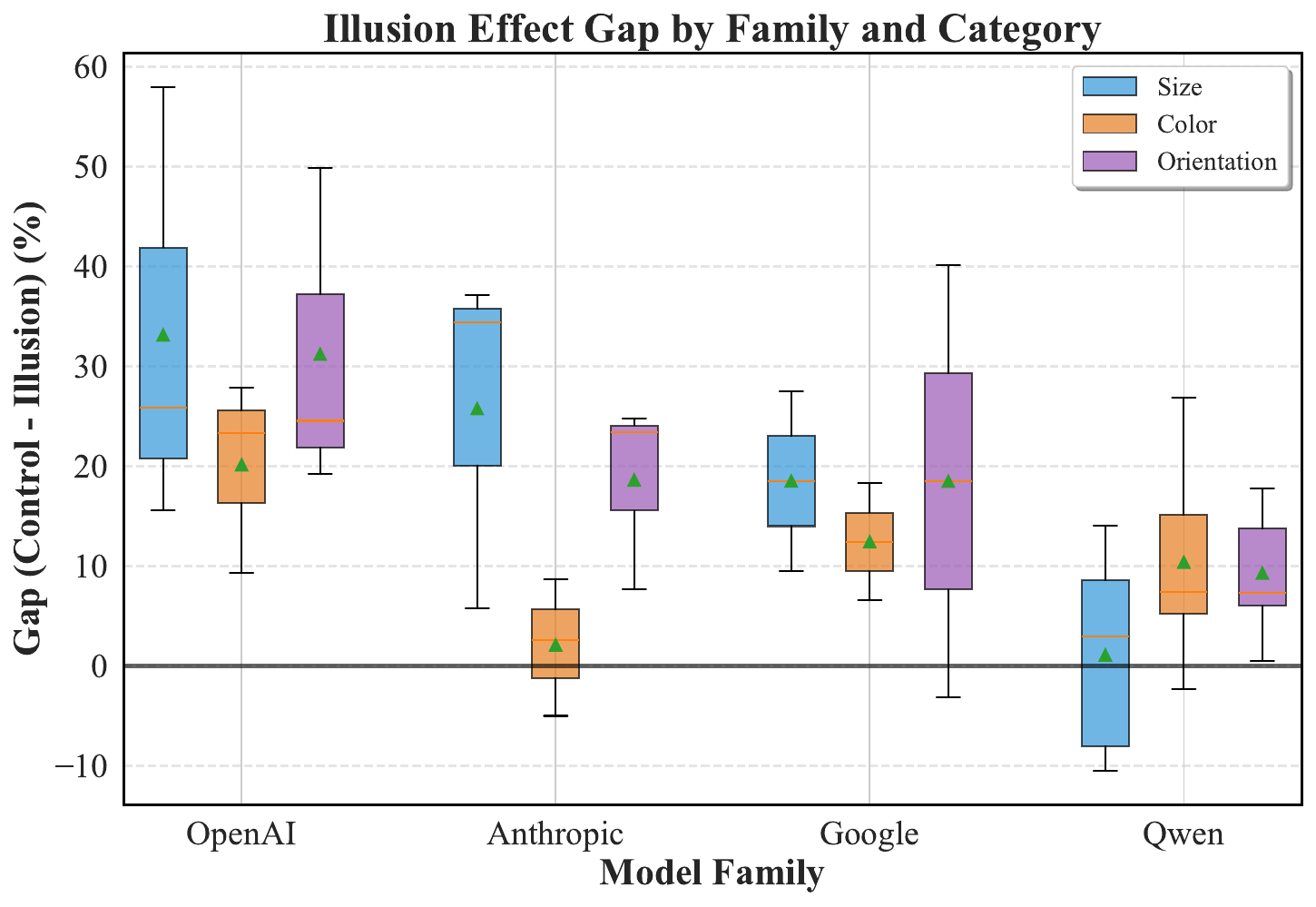}
\vspace{-15pt}
\caption{
\textbf{Susceptibility gaps by illusion category.} Boxes: Interquartile Range (IQR); triangles $\triangle$: means; black line: zero reference. Gap is computed as \texttt{perturbed\_control} minus \texttt{perturbed} paired accuracy.
}
\label{fig:family_gap}
\vspace{-5pt}
\end{figure}

\textbf{Model Family-Specific Illusion Susceptibility Patterns.}
Fig.~\ref{fig:family_gap} presents family-level susceptibility gaps (control - illusion accuracy) across Length, Color, and Orientation on controlled perturbations. OpenAI shows the largest overall gap (mean: 28.2\%), with particularly high gaps in Length (33.1\%) and Orientation (31.2\%). Anthropic and Google exhibit moderate overall gaps (15.5\% and 16.5\%), but differ in category profile: Anthropic is high on Length (25.8\%) and lower on Color (2.1\%), while Google is more uniformly positive across categories (Color: 12.4\%, Length: 18.5\%, Orientation: 18.5\%). Qwen has the smallest overall gap (6.9\%), with near-zero Length gap (1.1\%), moderate Orientation gap (9.3\%), and a positive Color gap (10.4\%). Overall, the figure indicates substantial family-level heterogeneity in illusion susceptibility patterns.

\noindent\textbf{Visual hints improve template matching but harm visual updating.} Table~\ref{tab:interventions_effect} (columns 2--3) shows that overlaid visual guidance (alignment marks, grids) consistently improves the Original accuracy (13/15 models, mean: +6.2pp; top gains: Qwen2.5-VL-7B +22.31\%, Haiku-4.5 +15.70\%). However, the same hints \emph{degrade} Perturbed accuracy for 12/15 models (mean: $-6.9$pp; largest drops: Sonnet-4.5 $-17.85$pp, Gemini-Flash $-12.48$pp). This asymmetry reveals hints function as pattern-completion cues that reinforce template retrieval rather than enabling flexible reasoning. When illusion factors are inverted, hints become misleading anchors pulling predictions toward memorized configurations. Only two Qwen models (Qwen3-VL-32B: +3.31\%, Qwen2.5-VL-72B: +0.33\%) show slight Perturbed gains, confirming weak template stores allow hints to guide perception rather than trigger recall.

\noindent\textbf{System prompts disengage memory at severe cost.} Prompts instructing models to ``ignore prior knowledge and compare carefully'' (Table~\ref{tab:interventions_effect}, columns 4 - 5) expose binary mode switching. Memory-driven models ($R{>}1.2$) show catastrophic trade-offs: Perturbed surges while Original collapses. GPT-5, with a Perturbed score of +63.97\%, shows visual capability when templates are suppressed, yet Original plunges $-84.30\%$. Similar patterns appear across GPT-5-Mini, Gemini-Flash, Claude-Sonnet, indicating prompts force all-or-nothing switching between template retrieval and visual analysis. In contrast, three smaller Qwen models (2.5-VL-3B/7B/32B; \textcolor{blue}{blue} in table) exhibit gains on both conditions (Original: +2.48\% to 26.45\%; Perturbed: +1.57\% to 13.30\%), suggesting weak template stores allow uniform improvement. Frontier VLMs lack mechanisms to adaptively balance memory and perception.

\section{Discussion}
\label{sec:discussion}

\noindent\textbf{Q1: Are current VLMs actually good at vision illusion cases?}
No. Many models retain the same response even when visual evidence changes (Table~\ref{tab:benchmark_comparison}). While static accuracy on fixed datasets may appear high, this often reflects language priors or memorized templates rather than the true perception of altered visual cues. For completeness, inducer-only results are in the Supplementary Materials.

\begin{table}[t]\footnotesize
\centering
\setlength{\tabcolsep}{4pt}
\begin{tabular}{l|cc|cc}
\toprule
\multirow{2}{*}{Model} & \multicolumn{2}{c|}{+ Visual Hints} & \multicolumn{2}{c}{+ System Prompt} \\
& original & perturbed & original & perturbed \\
\midrule
GPT-5 & \cellbarPos{1.65} & \cellbarNeg{-1.82} & \cellbarNeg[bold]{-84.30} & \cellbarPos[bold]{63.97} \\
GPT-5-Mini & \cellbarPos{2.48} & \cellbarNeg{-3.97} & \cellbarNeg{-28.93} & \cellbarPos{34.62} \\
GPT-5-Nano & \cellbarPos{14.87} & \cellbarNeg{-2.31} & \cellbarNeg{-4.13} & \cellbarPos{11.49} \\
Claude-Opus-4.1 & \cellbarPos{3.31} & \cellbarNeg{-6.28} & \cellbarNeg{-18.18} & \cellbarPos{21.16} \\
Claude-Sonnet-4.5 & \cellbarPos{4.13} & \cellbarNeg[bold]{-17.85} & \cellbarNeg{-21.49} & \cellbarPos{22.90} \\
Claude-Haiku-4.5 & \cellbarPos{15.70} & \cellbarNeg{-7.68} & \cellbarNeg{-4.96} & \cellbarPos{4.71} \\
Gemini-2.5-Flash & \cellbarPos{4.13} & \cellbarNeg{-12.48} & \cellbarNeg{-19.01} & \cellbarPos{28.26} \\
Gemini-2.5-Flash-Lite & \cellbarPos{2.48} & \cellbarNeg{-7.69} & \cellbarNeg{-11.57} & \cellbarPos{18.59} \\
\hline
Qwen3-VL-235B & \cellbarPos{14.87} & \cellbarNeg{-9.09} & \cellbarNeg{-13.23} & \cellbarPos{13.14} \\
Qwen3-VL-32B & \cellbarNeg[bold]{-10.74} & \cellbarPos[bold]{3.31} & \cellbarNeg{-28.92} & \cellbarPos{20.66} \\
Qwen3-VL-8B & \cellbarPos{8.26} & \cellbarNeg{-4.30} & \cellbarNeg{-5.79} & \cellbarPos{2.39} \\
Qwen2.5-VL-72B & \cellbarNeg{-5.78} & \cellbarPos{0.33} & \cellbarNeg{-5.78} & \cellbarPos{20.66} \\
Qwen2.5-VL-32B & \cellbarPos{4.96} & \cellbarNeg{-3.30} & \textcolor{blue}{\cellbarPos{2.48}} & \textcolor{blue}{\cellbarPos[bold]{1.57}} \\
Qwen2.5-VL-7B & \cellbarPos[bold]{22.31} & \cellbarNeg{-5.79} & \textcolor{blue}{\cellbarPos{5.78}} & \textcolor{blue}{\cellbarPos{6.36}} \\
Qwen2.5-VL-3B & \cellbarPos{9.09} & \cellbarNeg{-2.65} & \textcolor{blue}{\cellbarPos[bold]{26.45}} & \textcolor{blue}{\cellbarPos{13.30}} \\
\bottomrule
\end{tabular}
\vspace{-3pt}
\caption{\textbf{Effect of visual hints and system prompts on model performance.} Measured on Size illusion cases across Original and Perturbed conditions. Visual hints: overlaid alignment marks or measurement grids. System prompt: explicit instruction to ``ignore prior knowledge and compare carefully." Effect = accuracy\_with\_intervention $-$ accuracy\_without\_intervention. \textcolor{blue!60!black}{Blue bars} indicate improvement; \textcolor{red!60!black}{red bars} indicate degradation. 
}
\label{tab:interventions_effect}
\vspace{-5pt}
\end{table}

\noindent\textbf{Q2: What is the underlying technical problem?}
A reliable VLM should flip its answer when visual changes reverse the predicate, and keep it when edits preserve it. Current models often fail due to: (1) entangled visual representations; (2) language-prior dominance~\cite{yang2025illusions}; (3) interface bottlenecks such as low resolution~\cite{guo2024llava} or weak cross-attention~\cite{feng2025vision}; (4) objective mismatch from missing counterfactual consistency during training~\cite{vo2025visionlanguagemodelsbiased}; and (5) inference inertia, where decoding defaults to high-prior continuations. This paper focuses on analyzing key factors and offering insights into remaining considerations.

\noindent\textbf{Q3: How can we design better VLMs to address these issues?}
Addressing these failures requires rethinking how VLMs integrate perception and memory. Based on the analysis of this paper, we think potential key directions include: (1) \textit{perception-first architectures} that explicitly compare visual evidence before concluding; (2) \textit{counterfactual consistency objectives} that train models to reverse responses when visual factors invert (such as perturbed data in Fig.~\ref{fig:fig-VI-Pro}); (3) \textit{graded datasets} with control pairs to distinguish perception-driven from memory-driven responses (e.g., controlled perturbations discussed in Sec.~\ref{sec:Perturbation Strength}); and (4) \textit{inference-time prompting} requiring explicit visual verification steps (Sec.~\ref{sec:Interventions} provides some results and insights).

\section{Conclusion}
\label{sec: conclusion}
This work introduced VI-Probe, a controllable visual-illusion framework that disentangles visual perception from language-driven memory in VLMs. Through systematic manipulation of visual and linguistic cues, VI-Probe reveals that response persistence arises from heterogeneous causes: memory override, perception-memory competition, and visual-processing limits. Our analyses expose several critical gaps in current VLMs, \textit{e.g.}, (1) consistent yet visually ungrounded answers, revealing fundamental decoupling between perception and memory, and (2) robustness that does not scale monotonically with model size.
Beyond illusions, these findings highlight a broader challenge in integrating visual evidence with prior knowledge. We hope that counterfactual-aware objectives, perception-first architectures, and probing-based evaluation will be essential towards models that genuinely \emph{see} rather than merely \emph{recall} and extend flip-consistency evaluation from illusions to real-world scenes (e.g., charts, UIs, medical images).


{
    \small
    \bibliographystyle{ieeenat_fullname}
    \bibliography{main}
}
\appendix
%

In this appendix, we provide additional materials that complement the main paper and offer a more complete view of our dataset construction, experimental setup, and empirical findings. Specifically, we first present further details about the data in VI-Probe, including the illusion categories and the exact prompts used in our evaluations. We then summarize the models evaluated in the main paper and include additional experimental results that were omitted from the main submission due to space constraints. We also provide more visual examples from VI-Probe to illustrate the diversity and structure of the benchmark. Finally, we include further discussion to help contextualize the main findings and support a more detailed understanding of how visual illusions can be used to probe perception-versus-memory behavior in vision-language models.

\section{VI-Probe Details $\&$ Examples}
\label{sec:data-VI-Probe}

This section provides a more detailed overview of VI-Probe, with a particular focus on the benchmark composition, prompting format, and representative examples. Our goal is to make the dataset construction and evaluation protocol fully transparent so that readers can better understand what each test case measures and how the benchmark isolates perception-driven behavior from memory-driven responses. In particular, we clarify the set of illusion categories covered by VI-Probe, the exact natural-language questions posed to the models, and the instruction templates used throughout our experiments.

Table~\ref{tab:all_illusions} lists the 27 illusion cases currently included in VI-Probe, spanning size, color, and orientation phenomena. The corresponding case-specific questions are provided in Table~\ref{tab:illusion_questions_only}. Together, these materials give a concrete view of how we convert classic visual illusions into standardized VLM evaluation instances while keeping the query format simple and consistent across cases. We will release the images and code used to generate the different data variants, and we also plan to maintain an online resource so that the benchmark can be expanded with additional illusion families and new diagnostic settings over time.

\begin{table}[h]
\centering
\small
\setlength{\tabcolsep}{7pt}{
\begin{tabular}{c|ll}
\toprule
\textbf{\#} & \textbf{Illusion Name} & \textbf{Category} \\
\midrule
1 & Müller Lyer Illusion & size \\
2 & Circle Müller Lyer Illusion & size \\
3 & Ponzo Illusion & size \\
4 & Ponzo Trapezoid Illusion & size \\
5 & Ebbinghaus Illusion & size \\
6 & Ebbinghaus Illusion Rectangular & size \\
7 & Delboeuf Illusion & size \\
8 & Oppel Kundt Illusion & size \\
9 & Irradiation Illusion & size \\
10 & Irradiation Pentagon Illusion & size \\
11 & Circle Ponzo Illusion & size \\
\midrule
12 & Cornsweet Illusion & color \\
13 & Simultaneous Contrast Illusion & color \\
14 & Munker White Illusion & color \\
15 & Mach Band Illusion & color \\
16 & Mach Band Illusion Case2 & color \\
17 & Chubb Illusion & color \\
18 & Cornsweet Illusion Case1 & color \\
\midrule
19 & Hering Illusion & orientation \\
20 & Hering Illusion Vertical & orientation \\
21 & Zöllner Illusion & orientation \\
22 & Zöllner Illusion Vertical & orientation \\
23 & Twisted Cord Illusion & orientation \\
24 & Twisted Cord Illusion Light & orientation \\
25 & Poggendorff Illusion & orientation \\
26 & Poggendorff Horizontal Illusion & orientation \\
27 & Ehrenstein Illusion & orientation \\
\hline
\end{tabular}}
\caption{List of all Illusions cases, categorized into three groups.}
\label{tab:all_illusions}
\end{table}

\begin{table}[h]\scriptsize
\centering
\begin{tabular}{c|p{5.5cm}|c}
\toprule
\textbf{\#} & \textbf{Question} & \textbf{Ori-Answer} \\
\midrule
1  & Are the two black lines of equal length? & Yes \\
2  & Are the two black lines of equal length? & Yes \\
3  & Are the two horizontal black lines of equal length? & Yes \\
4  & Are the two horizontal black lines of equal length? & Yes \\
5  & Are the two orange circles the same size? & Yes \\
6  & Are the two orange circles the same size? & Yes \\
7  & Are the two solid circles the same size? & Yes \\
8  & Are the distances between the vertical markers labeled A–B and B–C equal? & Yes \\
9  & Are the left white square and the right black square equal in size? & Yes \\
10 & Are the left white pentagon and the right black pentagon equal in size? & Yes \\
11 & Are the two circles the same size? & Yes \\
\midrule
12 & Are the two circles of the same color? & Yes \\
13 & Are the two small squares of the same color? & Yes \\
14 & Are the two rectangle the same color? & Yes \\
15 & Is there a boundary in between every adjecent regions? & No \\
16 & Is there a boundary in between every adjecent regions? & No \\
17 & Are the two circles of the same color? & Yes \\
18 & Are the two vertical bands of the same color? & Yes \\
\midrule
19 & Are the two horizontal lines straight? & Yes \\
20 & Are the two vertical lines straight? & Yes \\
21 & Are the those red lines straight? & Yes \\
22 & Are the those red lines straight? & Yes \\
23 & Are those vertical columns parallel? & Yes \\
24 & Are those vertical columns parallel? & Yes \\
25 & Are the red and black solid diagonal lines aligned? & Yes \\
26 & Are the red and black solid diagonal lines aligned? & Yes \\
27 & Do the squares on the left and right have straight edges? & Yes \\
\hline
\end{tabular}
\caption{Question prompts for all illusion cases, together with the ground-truth answer for the original illusion image.}
\label{tab:illusion_questions_only}
\end{table}

\noindent\textbf{Original Full Prompts} used in the experiments are listed below to clarify the exact instructions given to the models and to facilitate reproducibility. We include the prompt template in full because small wording differences can meaningfully affect VLM behavior, especially in tasks that require careful visual comparison under potentially conflicting semantic priors.

\begin{tcolorbox}[
    colback=gray!3,           
    colframe=gray!40,         
    rounded corners,          
    width=\linewidth,         
    arc=2mm,                  
    boxrule=0.5pt,            
    left=3pt, right=3pt,      
    top=2pt, bottom=2pt       
]
{\noindent\footnotesize\texttt{\noindent
Are the two black lines of equal length?\\
Answer Instructions:\\
1. Write your reasoning inside <reasons>...</reasons>.\\
\ \ - Use natural language explanation.\\
2. Give the final numeric answer inside <answer>...</answer>.\\
\ \ - Use "1" if yes.\\
\ \ - Use "0" if no.\\
\ \ - Do not write anything else inside <answer>.}}
\end{tcolorbox}


The first line is adapted to each illusion case according to the questions in Table~\ref{tab:illusion_questions_only}, while the answer format and reasoning instructions remain fixed. This design lets us vary the visual task itself without changing the response protocol, making the benchmark more controlled and allowing differences in model performance to be attributed more directly to the visual content rather than prompt variation.

\noindent\textbf{Visual Comparison Instruction Prompt}
The following prompt is added to the system prompt for linguistic variation and for isolating the model from prior semantic knowledge. It is intended to explicitly encourage image-grounded reasoning, reduce reliance on memorized illusion templates, and test whether models can shift from prior-driven responses to direct visual comparison when such behavior is instructed.

\begin{tcolorbox}[
    colback=gray!3,           
    colframe=gray!40,          
    rounded corners,           
    width=\linewidth,      
    arc=2mm,                   
    boxrule=0.5pt,            
    left=3pt, right=3pt,      
    top=2pt, bottom=2pt,       
]
{\noindent\footnotesize\texttt{\noindent Visual Comparison Instructions:\\
Base your judgment exclusively on direct visual perception of the image. Compare the two targets systematically using only what is visible in the image itself.\\
Critical constraints:\\
	•	Disregard language priors and linguistic biases\\
	•	Ignore implications from question phrasing\\
	•	Set aside world knowledge and assumptions\\
	•	Do not rely on typical patterns or expectations\\
Required approach:\\
	1.	Evaluate the "equal” hypothesis against visual evidence\\
	2.	Evaluate the "not equal” hypothesis against visual evidence\\
	3.	Compare which hypothesis better matches the observable data\\
	4.	Provide a binary answer based solely on this visual analysis\\
Your response must be grounded entirely in what you can directly perceive in the image.}
}
\end{tcolorbox}


Table~\ref{tab: Notation} provides a quick reference for the notation used in Section 3.2 of the paper.  

\begin{table}[h]
\centering
\footnotesize
\begin{tabular}{ll}
\toprule
Symbol & Meaning \\
\midrule
$x^O$ & Original illusion image \\
$x^P$ & Perturbed image (control factor inverted) \\
$x^{OC}\text{, }x^{PC}$ & Control (inducers removed) \\
$x^{OH}\text{, }x^{PH}$ & Hinted (visual cues overlaid) \\
$q^f \text{, } q^r \text{, } q^I$ & Forward / Reversed / Instructional question \\
$\alpha$ & Perturbation strength \\
\bottomrule
\end{tabular}
\caption{Notation quick reference of VI-Probe} \label{tab: Notation}
\end{table}
$u$ represents any image of $x^O$, $x^P$, $x^{OC}$, $x^{PC}$, or $x^{OH}$, $x^{PH}$. The metrics for Paraphrase-pair consistency (same vs. different) are calculated as follows:

\noindent\textbf{Polarity-Flip Consistency (PFC).}
\[
\mathrm{PFC}
=\frac{1}{|\mathcal{U}|}\sum_{u\in\mathcal{U}}
\mathbf{1}\!\big[a(u,q^{\mathrm{r}})=1-a(u,q^{\mathrm{f}})\big].
\]

\noindent\textbf{Polarity-Flip Accuracy (PFA).}
\[
\mathrm{PFA}
=\frac{1}{|\mathcal{U}|}\sum_{u\in\mathcal{U}}
\mathbf{1}\!\big[a(u,q^{\mathrm{f}})=y^{\mathrm{f}}(u)\big]\,
\mathbf{1}\!\big[a(u,q^{\mathrm{r}})=y^{\mathrm{r}}(u)\big].
\]

\noindent\textbf{Template Fixation Index (TFI).}
\[
\mathrm{TFI}
=\frac{1}{|\mathcal{U}|}\sum_{u\in\mathcal{U}}
\mathbf{1}\!\big[a(u,q^{\mathrm{r}})=a(u,q^{\mathrm{f}})\big]
=1-\mathrm{PFC}.
\]

\section{Experiment}

\subsection{Experimental setting}
\label{sec:App-et}

\textbf{VLMs evaluated using VI-Probe.} The main paper evaluates the 15 most recent models from four model families (OpenAI~\cite{singh2025openai} \includegraphics[width=0.3cm,valign=c]{fig/logo/openai.png}, Anthropic~\cite{anthropic2025claude4} \includegraphics[width=0.3cm,valign=c]{fig/logo/anthropic.png}, Google~\cite{comanici2025gemini} \includegraphics[width=0.35cm,valign=c]{fig/logo/gemini.png}, Qwen3-VL and Qwen2.5-VL series~\cite{yang2025qwen3} \includegraphics[width=0.3cm,valign=c]{fig/logo/qwen.png}). Earlier versions have also been assessed in the supplementary materials. Model names are provided in Table~\ref{tab:eval_models}.


\begin{table}[h]\scriptsize
\setlength{\tabcolsep}{1pt}{
\begin{tabular}{@{}l|lll@{}}
\toprule
   \multicolumn{4}{c}{Model Series 1}                                   \\ \midrule
ChatGPT \cite{achiam2023gpt} & GPT-5              & GPT-5-mini        & GPT-5-nano            \\
Claude \cite{anthropic2025claude4}  & Claude Opus 4.1    & Claude Sonnet 4.5 & Claude Haiku 4.5      \\
Gemini \cite{comanici2025gemini}  & Gemini 2.5 Pro      & Gemini 2.5 Flash  & Gemini 2.5 Flash-Lite \\
\multirow{2}{*}{Qwen \cite{yang2025qwen3}}     & Qwen3-VL-235B-A22B & Qwen3-VL-30B      & Qwen3-VL-8B           \\
& Qwen2.5-VL-72B     & Qwen2.5-VL-32B    & Qwen2.5-VL-7B         \\
\midrule
         \multicolumn{4}{c}{Model Series 2}                                   \\
\midrule
ChatGPT \cite{achiam2023gpt} & GPT-4o             & GPT-4o-mini       &                       \\
Claude \cite{anthropic2025claude4}  &  Claude 3.5 Sonnet & Claude 3.5 Haiku    &   \\
\bottomrule
    \end{tabular}}
\caption{List of evaluated models. Current versions and previous versions are tested for consistency.}
\label{tab:eval_models}
\end{table}

\noindent\textbf{Accuracy reported in the paper.} With the exception of the Polarity-Flip Consistency framework (Sec 4.2.1.), accuracy on all other data types is computed by requiring correct answers to both paired questions, \textit{i.e.} a case is marked correct only when the model answers both the forward and reverse prompts correctly.

\subsection{Additional results and analysis}

\subsubsection{Cross-Generation Consistency: Previous Models}

To assess consistency of illusion susceptibility patterns across model generations, we evaluated four predecessor models (Claude-3.5-Sonnet, Claude-3.5-Haiku, GPT-4o, GPT-4o-mini) using the same VI-Probe pipeline. Table~\ref{tab:paired_by_type_newmodels} reports aggregate accuracy across Size illusion categories, following the same metrics as the main paper.

\begin{table}[t]\scriptsize
\centering
\setlength{\tabcolsep}{4pt}{
\begin{tabular}{l|cc|cc|c}
\toprule
Model & Original & Perturbed & Original  & Perturbed  & $R$ \\
      & & & Control       & Control        &     \\
\midrule
claude-3.5-sonnet  & 86.78 & 12.40 & 90.91 & 32.89 & 1.28 \\
claude-3.5-haiku   & 60.33 & 35.04 & 72.73 & 55.29 & 1.45 \\
\hline
gpt-4o             & 70.25 & 20.50 & 96.69 & 49.01 & 1.04 \\
gpt-4o-mini        & 19.83 & 66.94 & 94.21 & 53.31 & 1.15 \\
\bottomrule
\end{tabular}}
\caption{Accuracy by image type for previous-generation models. $R$ = (Original $-$ Perturbed) / (OC $-$ PC) quantifies illusion-specific memory interference relative to baseline visual difficulty.}
\label{tab:paired_by_type_newmodels}
\end{table}

\vspace{1mm}
\noindent\textbf{Pattern replication across model generations.} Previous-generation models show qualitatively similar failure modes to their successors:

\begin{itemize}[leftmargin=*, itemsep=1pt, topsep=2pt]
    \item \textbf{Claude-3.5-Sonnet} replicates the memory-driven profile of Claude-Sonnet-4.5. It has resulting in a significant 74.38\% Original$\rightarrow$Perturbed drop (86.78$\rightarrow$12.40) coupled with high $R=1.28$. This 6$\times$ accuracy ratio (Original/Perturbed = 7.0) confirms a strong template reliance persists across Claude generations.

    \item \textbf{Claude-3.5-Haiku} shows more balanced performance (Original 60.33, Perturbed 35.04), similar to Claude-Haiku-4.5's relatively lower susceptibility. However, its $R=1.45$ is notably higher than Haiku-4.5's, suggesting older versions may have stronger illusion-specific biases despite lower overall capacity.

    \item \textbf{GPT-4o} shows moderate memory-driven behavior (70.25$\rightarrow$20.50, 49.75\% drop) with $R=1.04$, indicating illusion effects barely exceed baseline difficulty. This aligns with the GPT-5 profile, but it demonstrates weaker overall performance, which is consistent with the architectural improvements in GPT-5.

    \item \textbf{GPT-4o-mini} exhibits inverted susceptibility. Original (19.83) $<$ Perturbed (66.94). It mirrors GPT-5-Nano's anomalous pattern. This +47.11\% advantage on Perturbed images with strong control performance (OC: 94.21\%) confirms the inversion stems from weak illusion-specific templates rather than poor visual processing. When models lack memorized patterns, they use visual analysis and often perform better on perturbed images.
\end{itemize}

\vspace{1mm}
\noindent\textbf{Accuracy on control data as an indicator of model visual capacity.} Original Control accuracy serves as a proxy for baseline visual capability: GPT-4o (96.69\%) and GPT-4o-mini (94.21\%) both exceed Claude-3.5 models (72.73--90.91\%), suggesting OpenAI models possess stronger low-level visual processing. Yet this advantage disappears under illusions—Claude-3.5-Sonnet's Original accuracy exceeds GPT-4o, showing that template retrieval can override visual signals even when perceptual capacity is adequate.



\vspace{1mm}
\noindent\textbf{$R$ Values Confirm Architecture-Specific Biases.} The rank-ordering of $R$ values shows illusion susceptibility is not simply a function of model size or recency (Claude-3.5-Haiku (1.45) $>$ Claude-3.5-Sonnet (1.28) $>$ GPT-4o-mini (1.15) $>$ GPT-4o (1.04)). Instead, it reflects architectural and training choices:

\begin{itemize}[leftmargin=*, itemsep=1pt, topsep=2pt]
    \item \textbf{Claude models} consistently show $R>1.2$. This suggests that Anthropic's training pipeline emphasizes pattern matching and alignment with world knowledge, potentially increasing susceptibility to illusions.
    \item \textbf{Previous OpenAI models} show $R \approx 1.0$--$1.15$, indicating their architectures or training data distributions produce weaker illusion-specific biases.
    \item The inverted scaling within model families (Haiku $>$ Sonnet, Mini $>$ Base) replicates across generations, confirming that increased capacity systematically amplifies template reliance in some model families.
\end{itemize}

\vspace{1mm}
\noindent
\begin{tcolorbox}[
    colback=gray!10,
    colframe=gray!40,
    rounded corners,
    width=\linewidth,
    arc=2mm,
    boxrule=0.5pt,
    left=3pt, right=3pt,
    top=2pt, bottom=2pt
]
\textbf{Cross-Generation Takeaways}
\begin{itemize}[leftmargin=*, itemsep=1pt, topsep=2pt]
    \item \textbf{Failure modes persist across versions}: Claude-3.5-Sonnet's 7$\times$ accuracy drop mirrors Claude-4.5-Sonnet, confirming architectural biases dominate over incremental improvements.
    \item \textbf{Inverted susceptibility replicates}: GPT-4o-mini (Original 19.83, Perturbed 66.94) mirrors GPT-5-Nano, validating weak-template explanations.
    \item \textbf{Control dissociations are universal}: All models show larger Original$\rightarrow$Perturbed drops on illusions than controls, confirming memory interference transcends model generations.
\end{itemize}
\end{tcolorbox}

These cross-generation results validate that VI-Probe captures stable, architecture-level phenomena rather than transient artifacts of specific model versions. The consistency of failure patterns across iterations suggests fundamental limitations in how current VLM training paradigms balance perception and memory.

\subsubsection{Detailed Analysis of Intervention Effects on Size Illusions}
We now present full results for the intervention experiments in Sec.~4.2.4, focusing on Size illusions. We test two strategies: (1)~visual hints (alignment marks, measurement grids) and (2)~system prompts instructing models to ignore prior knowledge and rely on visual evidence. Tables~\ref{tab:guide_effect_length}, \ref{tab: sp_effect_length}, and \ref{tab: sp_effect_length_control} show accuracy breakdowns across all 15 models.

\noindent\textbf{Visual Hints: Reinforcing Templates Rather Than Enabling Visual Reasoning.} Table~\ref{tab:guide_effect_length} shows an asymmetry in how visual hints affect model performance. On Original illusion images, hints improve accuracy for 13 of 15 models (mean gain: +6.2\%). The largest improvements occur in perception-limited models: Qwen2.5-VL-7B (+22.31\%), Claude-Haiku-4.5 (+15.70\%), and Qwen3-VL-235B (+14.87\%). For models without strong illusion-specific templates, hints provide useful visual guidance.

However, on Perturbed images (where illusion factors are inverted), visual hints \emph{degrade} performance for 12 of 15 models (mean drop: $-6.9\%$). The largest degradations occur in memory-driven models: Claude-Sonnet-4.5 ($-17.85\%$), Gemini-2.5-Flash ($-12.48\%$), and Qwen3-VL-235B ($-9.09\%$). Visual hints thus act as \emph{pattern-completion cues} that reinforce memorized configurations. When visual evidence contradicts the expected template, hints become misleading anchors that pull predictions toward incorrect memorized patterns.
\begin{table}[t]\footnotesize
\centering
\setlength{\tabcolsep}{1pt}{
\begin{tabular}{l|ccc|ccc}
\toprule
\multirow{2}{*}{Model} & \multicolumn{3}{c|}{Original} & \multicolumn{3}{c}{Perturbed} \\
& w/o \tiny{Hints} & w/ \tiny{Hints} & Effect & w/o \tiny{Hints} & w/ \tiny{Hints} & Effect \\
& (\%) & (\%) & (\%) & (\%) & (\%) & (\%) \\
\midrule
GPT-5 & 98.35 & 100.00 & +1.65 & 1.98 & 0.17 & -1.82 \\
GPT-5-mini & 89.26 & 91.74 & +2.48 & 6.78 & 2.81 & -3.97 \\
GPT-5-nano & 19.01 & 33.88 & +14.87 & 66.03 & 63.72 & -2.31 \\
\hline
Claude-Opus-4.1 & 71.90 & 75.21 & +3.31 & 30.41 & 24.13 & -6.28 \\
Claude-Sonnet-4.5 & 60.33 & 64.46 & +4.13 & 41.07 & 23.22 & -17.85 \\
Claude-Haiku-4.5 & 26.45 & 42.15 & +15.70 & 72.31 & 64.63 & -7.68 \\
\hline
Gemini-2.5-Flash & 77.69 & 81.82 & +4.13 & 22.98 & 10.50 & -12.48 \\
Gemini-2.5-Flash-Lite & 42.15 & 44.63 & +2.48 & 46.86 & 39.17 & -7.69 \\
\hline
Qwen3-VL-235B & 80.17 & 95.04 & +14.87 & 12.48 & 3.39 & -9.09 \\
Qwen3-VL-32B & 94.21 & 83.47 & -10.74 & 2.23 & 5.54 & +3.31 \\
Qwen3-VL-8B & 63.64 & 71.90 & +8.26 & 31.49 & 27.19 & -4.30 \\
Qwen2.5-VL-72B & 71.07 & 65.29 & -5.78 & 6.03 & 6.37 & +0.33 \\
Qwen2.5-VL-32B & 56.20 & 61.16 & +4.96 & 20.58 & 17.27 & -3.30 \\
Qwen2.5-VL-7B & 53.72 & 76.03 & +22.31 & 19.26 & 13.47 & -5.79 \\
Qwen2.5-VL-3B & 11.57 & 20.66 & +9.09 & 17.36 & 14.71 & -2.65 \\
\bottomrule
\end{tabular}}
\caption{Guide Effect on Model Performance (Size Illusion). w/o Hints: without visual hints, w/ Hints: with visual hints overlaid. Effect = w/ Hints - w/o Hints. Positive values indicate improvement with visual guidance.}
\label{tab:guide_effect_length}
\end{table}

Only two models show slight improvements on Perturbed images with hints: Qwen3-VL-32B (+3.31\%) and Qwen2.5-VL-72B (+0.33\%). Both have weak illusion-specific templates (low $R$ in Table 2), so hints can guide visual perception without triggering strong template retrieval.

\noindent\textbf{System Prompts: Exposing Binary Mode Switching.} Tables~\ref{tab: sp_effect_length} and \ref{tab: sp_effect_length_control} show that instructing models to ``ignore prior knowledge and compare carefully'' triggers \emph{all-or-nothing mode switching} in memory-driven models. Table~\ref{tab: sp_effect_length} demonstrates severe trade-offs for frontier models with $R>1.2$:

\begin{itemize}[leftmargin=*, itemsep=1pt, topsep=2pt]
    \item \textbf{GPT-5}: Original drops $84.30\%$ (98.35\%$\rightarrow$14.05\%) while Perturbed surges $63.97\%$ (1.98\%$\rightarrow$65.95\%)
    \item \textbf{GPT-5-mini}: Original drops $28.93\%$ while Perturbed gains $34.62\%$
    \item \textbf{Gemini-2.5-Flash}: Original drops $19.01\%$ while Perturbed gains $28.26\%$
    \item \textbf{Claude-Sonnet-4.5}: Original drops $21.49\%$ while Perturbed gains $22.90\%$
\end{itemize}

These extreme swings demonstrate that system prompts force binary switching between two mutually exclusive modes: (1)~template retrieval (high Original, low Perturbed) and (2)~visual analysis (low Original, high Perturbed). For example, GPT-5 achieved an accuracy of 65.95\% on Perturbed images with system prompts, compared to only 1.98\% without them. This demonstrates that the model has sufficient visual capability when templates are suppressed. The failure stems not from perceptual limits but from the inability to adaptively balance memory and perception.

Conversely, three smaller Qwen models (2.5-VL-3B/7B/32B, shown in \textcolor{blue}{blue} in Table 3 of the main paper) exhibit \emph{simultaneous gains} on both Original and Perturbed conditions:

\begin{itemize}[leftmargin=*, itemsep=1pt, topsep=2pt]
    \item \textbf{Qwen2.5-VL-3B}: Original +26.45\%, Perturbed +13.30\%
    \item \textbf{Qwen2.5-VL-7B}: Original +5.78\%, Perturbed +6.36\%
    \item \textbf{Qwen2.5-VL-32B}: Original +2.48\%, Perturbed +1.57\%
\end{itemize}

This uniform improvement confirms that weak template stores allow system prompts to enhance visual reasoning without triggering mode collapse. These models benefit from explicit instructions because they lack strong priors to override.

\begin{table}[t]\footnotesize
\centering
\setlength{\tabcolsep}{1pt}{
\begin{tabular}{l|ccc|ccc}
\toprule
\multirow{2}{*}{Model} 
 & \multicolumn{3}{c|}{Original} 
 & \multicolumn{3}{c}{Perturbed} \\
\cline{2-7}
 & w/o SP & w/ SP  & Diff.
 & w/o SP  & w/ SP  & Diff. \\
\hline
GPT-5            & 98.35 & 14.05 & $-84.30$ &  1.98 & 65.95 & $+63.97$ \\
GPT-5-mini       & 89.26 & 60.33 & $-28.93$ &  6.78 & 41.40 & $+34.62$ \\
GPT-5-nano       & 19.01 & 14.88 & $ -4.13$ & 66.03 & 77.52 & $+11.49$ \\
Claude-Opus-4.1  & 71.90 & 53.72 & $-18.18$ & 30.41 & 51.57 & $+21.16$ \\
Claude-Sonnet-4.5& 60.33 & 38.84 & $-21.49$ & 41.07 & 63.97 & $+22.90$ \\
Claude-Haiku-4.5 & 26.45 & 21.49 & $ -4.96$ & 72.31 & 77.02 & $ +4.71$ \\
Gemini-2.5-Flash & 77.69 & 58.68 & $-19.01$ & 22.98 & 51.24 & $+28.26$ \\
Gemini-2.5-Flash-Lite 
                 & 42.15 & 30.58 & $-11.57$ & 46.86 & 65.45 & $+18.59$ \\
Qwen3-VL-235B    & 80.17 & 66.94 & $-13.23$ & 12.48 & 25.62 & $+13.14$ \\
Qwen3-VL-32B     & 94.21 & 65.29 & $-28.92$ &  2.23 & 22.89 & $+20.66$ \\
Qwen3-VL-8B      & 63.64 & 57.85 & $ -5.79$ & 31.49 & 33.88 & $ +2.39$ \\
Qwen2.5-VL-72B   & 71.07 & 65.29 & $ -5.78$ &  6.03 & 26.69 & $+20.66$ \\
Qwen2.5-VL-32B   & 56.20 & 58.68 & $ +2.48$ & 20.58 & 22.15 & $ +1.57$ \\
Qwen2.5-VL-7B    & 53.72 & 59.50 & $ +5.78$ & 19.26 & 25.62 & $ +6.36$ \\
Qwen2.5-VL-3B    & 11.57 & 38.02 & $+26.45$ & 17.36 & 30.66 & $+13.30$ \\
\bottomrule
\end{tabular}}
\caption{Comparison of Both Correct Accuracy with and without system prompting (SP) on \textbf{Original and Perturbed images}.}
\label{tab: sp_effect_length}
\end{table}

\begin{table}[t]\footnotesize
\centering
\setlength{\tabcolsep}{1pt}{
\begin{tabular}{l|ccc|ccc}
\toprule
\multirow{2}{*}{Model} 
 & \multicolumn{3}{c|}{Original} 
 & \multicolumn{3}{c}{Perturbed} \\
\cline{2-7}
 & w/o SP & w/ SP  & Diff.
 & w/o SP  & w/ SP  & Diff. \\
\hline
GPT-5            & 100.00 & 100.00 & $+0.00$  & 59.92 & 65.37 & $+5.45$  \\
GPT-5-mini       &  99.17 &  95.87 & $-3.30$  & 32.64 & 58.10 & $+25.46$ \\
GPT-5-nano       &  25.62 &  30.58 & $+4.96$  & 81.65 & 72.81 & $-8.84$  \\
Claude-Opus-4.1  &  96.69 &  95.87 & $-0.82$  & 67.52 & 66.36 & $-1.16$  \\
Claude-Sonnet-4.5&  92.56 &  89.26 & $-3.30$  & 75.45 & 73.55 & $-1.90$  \\
Claude-Haiku-4.5 &  90.91 &  85.12 & $-5.79$  & 78.10 & 80.00 & $+1.90$  \\
Gemini-2.5-Flash & 100.00 &  95.04 & $-4.96$  & 50.50 & 55.95 & $+5.45$  \\
Gemini-2.5-Flash-Lite
                 &  97.52 &  64.46 & $-33.06$ & 56.36 & 77.60 & $+21.24$ \\
Qwen3-VL-235B    & 100.00 & 100.00 & $+0.00$  & 16.28 & 26.28 & $+10.00$ \\
Qwen3-VL-32B     & 100.00 & 100.00 & $+0.00$  & 15.62 & 55.70 & $+40.08$ \\
Qwen3-VL-8B      &  99.17 &  99.17 & $+0.00$  & 23.31 & 29.50 & $+6.19$  \\
Qwen2.5-VL-72B   & 100.00 &  97.52 & $-2.48$  & 20.08 & 36.36 & $+16.28$ \\
Qwen2.5-VL-32B   &  93.39 &  98.35 & $+4.96$  & 12.73 & 17.85 & $+5.12$  \\
Qwen2.5-VL-7B    &  85.12 & 100.00 & $+14.88$ & 22.23 & 22.56 & $+0.33$  \\
Qwen2.5-VL-3B    &  84.30 &  99.17 & $+14.87$ &  6.86 & 11.57 & $+4.71$  \\
\hline
\end{tabular}}
\caption{Comparison of Correct Accuracy with and without system prompting (SP) on \textbf{Original Control and Perturbed Control images}. Difference = w/ SP $-$ w/o SP.}
\label{tab: sp_effect_length_control}
\end{table}

\noindent\textbf{Control Condition Reveals True Visual Capability.} Table~\ref{tab: sp_effect_length_control} shows system prompt effects on control images (illusion patterns removed). On Original Control images, most models show minimal changes or slight degradation (mean:$-2.1$\%), with the notable exception of smaller Qwen models showing improvements (Qwen2.5-VL-7B: +14.88\%, Qwen2.5-VL-3B: +14.87\%). This means flagship models already operate near ceiling on simple comparisons, leaving little room for prompt-driven gains

On Perturbed Control images, system prompts yield broader improvements: 12 out of 15 models gain accuracy (mean: +11.8\%). The largest gains occur in models with moderate baseline performance: Qwen3-VL-32B (+40.08\%), GPT-5-mini (+25.46\%), and Gemini-2.5-Flash-Lite (+21.24\%). This indicates that explicit visual comparison instructions help when visual signals are degraded but no interfering templates exist.

Comparing Tables~\ref{tab: sp_effect_length} and \ref{tab: sp_effect_length_control} exposes a critical dissociation: flagship models improve substantially on Perturbed Control (+5.45\% to +25.46\%) yet collapse on Original illusions ($-84.30$\% to $-11.57$\%). This confirms that template retrieval drives failure under illusions, rather than perceptual limits. \textbf{When illusion semantics are absent (controls), the same prompts that suppress templates enable visual processing.}

\vspace{1mm}
\noindent
\begin{tcolorbox}[
    colback=gray!10,           
    colframe=gray!40,          
    rounded corners,           
    width=\linewidth,      
    arc=2mm,                   
    boxrule=0.5pt,            
    left=3pt, right=3pt,      
    top=2pt, bottom=2pt       
]
\textbf{Takeaways}
	\vspace{-0mm}
	\begin{itemize}
 \item Visual hints act as pattern-completion cues, improving template matching (+6.2\% on Original) but harming visual updating ($-6.9$\% on Perturbed).
    \item System prompts expose binary mode switching in flagship models: suppressing templates boosts Perturbed accuracy (+64\% for GPT-5) but devastates Original accuracy ($-84$\%).
    \item Smaller models with weak templates (Qwen2.5-VL-3B/7B/32B) show uniform gains, confirming interventions help when strong priors are absent.
    \item The control vs. illusion dissociation proves failures stem from memory interference, not perceptual capacity: prompts that improve Perturbed Control by $+25\%$ can collapse Original illusions by $-84\%$ in the same model.
  \vspace{-0mm}
\end{itemize}
\end{tcolorbox}

These results demonstrate that current VLMs lack metacognitive mechanisms to apply instructions selectively. They cannot modulate template reliance based on task demands, instead switching uniformly between retrieval-dominant and perception-dominant modes across all inputs.

\begin{table}[ht]\footnotesize
\footnotesize
\centering
\setlength{\tabcolsep}{0.5pt}{
\begin{tabular}{ll|ccccc}
\toprule
Group & Model & Original & OC & Perturbed & PC  & Inducer \\
\midrule
\multirow{3}{*}{OpenAI} 
 & GPT-5 & 1.00 & 1.00 & 0.01 & 0.55 & 1.00 \\
 & GPT-5-mini & 1.00 & 1.00 & 0.00 & 0.27 & 1.00 \\
 & GPT-5-nano & 0.45 & 0.18 & 0.65 & 0.95 & 1.00 \\
\cline{1-7}
\multirow{3}{*}{Anthropic} 
 & Claude-Opus-4.1 & 1.00 & 1.00 & 0.03 & 0.88 & 0.64 \\
 & Claude-Sonnet-4.5 & 1.00 & 1.00 & 0.01 & 0.94 & 0.91 \\
 & Claude-Haiku-4.5 & 0.55 & 1.00 & 0.61 & 0.99 & 1.00 \\
\cline{1-7}
\multirow{2}{*}{Google} 
 & Gemini-2.5-Flash & 1.00 & 1.00 & 0.01 & 0.54 & 1.00 \\
 & Gemini-2.5-Flash-Lite & 0.73 & 1.00 & 0.16 & 0.63 & 1.00 \\
\cline{1-7}
\multirow{7}{*}{Qwen} 
 & Qwen3-VL-235B-A22B & 1.00 & 1.00 & 0.99 & 0.32 & 0.73 \\
 & Qwen3-VL-32B & 1.00 & 1.00 & 0.72 & 0.49 & 1.00 \\
 & Qwen3-VL-8B & 1.00 & 1.00 & 0.25 & 0.20 & 1.00 \\
 & Qwen2.5-VL-72B & 1.00 & 1.00 & 0.12 & 0.21 & 0.00 \\
 & Qwen2.5-VL-32B & 1.00 & 1.00 & 0.00 & 0.06 & 1.00 \\
 & Qwen2.5-VL-7B & 0.36 & 1.00 & 0.06 & 0.26 & 0.91 \\
 & Qwen2.5-VL-3B & 0.00 & 1.00 & 0.01 & 0.04 & 0.00 \\
\bottomrule
\end{tabular}}
\caption{Model performance on the Müller-Lyer Illusion.}
\label{tab:muller_lyer}
\end{table}

\begin{table}[ht]
\footnotesize
\centering
\setlength{\tabcolsep}{0.5pt}{
\begin{tabular}{ll|ccccc}
\toprule
Group & Model & Original & OC & Perturbed & PC & Inducer \\
\midrule
\multirow{3}{*}{OpenAI} 
 & GPT-5 & 1.00 & 1.00 & 0.00 & 0.83 & 1.00 \\
 & GPT-5-mini & 1.00 & 1.00 & 0.02 & 0.56 & 1.00 \\
 & GPT-5-nano & 0.45 & 0.00 & 0.62 & 0.93 & 0.00 \\
\cline{1-7}
\multirow{3}{*}{Anthropic} 
 & Claude-Opus-4.1 & 1.00 & 1.00 & 0.00 & 0.80 & 0.00 \\
 & Claude-Sonnet-4.5 & 0.82 & 0.82 & 0.29 & 0.93 & 0.00 \\
 & Claude-Haiku-4.5 & 0.00 & 0.64 & 0.96 & 0.95 & 0.00 \\
\cline{1-7}
\multirow{2}{*}{Google} 
 & Gemini-2.5-Flash & 1.00 & 1.00 & 0.00 & 0.71 & 0.18 \\
 & Gemini-2.5-Flash-Lite & 1.00 & 1.00 & 0.01 & 0.72 & 0.18 \\
\cline{1-7}
\multirow{7}{*}{Qwen} 
 & Qwen3-VL-235B-A22B & 1.00 & 1.00 & 0.00 & 0.45 & 0.00 \\
 & Qwen3-VL-32B & 1.00 & 1.00 & 0.00 & 0.38 & 0.00 \\
 & Qwen3-VL-8B & 1.00 & 1.00 & 0.06 & 0.62 & 0.00 \\
 & Qwen2.5-VL-72B & 1.00 & 1.00 & 0.06 & 0.16 & 0.00 \\
 & Qwen2.5-VL-32B & 0.73 & 1.00 & 0.11 & 0.03 & 0.00 \\
 & Qwen2.5-VL-7B & 0.91 & 0.64 & 0.22 & 0.38 & 0.00 \\
 & Qwen2.5-VL-3B & 0.09 & 1.00 & 0.91 & 0.28 & 0.00 \\
\bottomrule
\end{tabular}}
\caption{Model performance on the Ebbinghaus Illusion.}
\label{tab:ebbinghaus}
\end{table}

\subsubsection{Case-Specific Analysis: Inducer-Only Tests Expose Pure Template Bias}

Tables~\ref{tab:muller_lyer} and \ref{tab:ebbinghaus} present detailed results for Müller-Lyer and Ebbinghaus illusions across Original, Perturbed, and Control variants. Critically, the \textbf{Inducer} column tests pure template bias: models are shown \emph{only} the illusion-inducing elements (e.g., arrow heads without line segments, surrounding circles without center circles) and asked the same question (``Are the two lines equal length?''). High Inducer accuracy indicates the model reproduces Original answers despite \emph{no visual evidence}, showing pure hallucination driven by memorized patterns.

\vspace{1mm}
\noindent\textbf{Inducer-Only Reveals Extreme Template Bias.} All OpenAI and Google models achieve perfect Inducer scores (1.00) on Müller-Lyer, meaning they answer identically to Original images even when the comparison targets are completely absent. GPT-5 demonstrates the most extreme bias: Original (1.00), Inducer (1.00), yet Perturbed (0.01). It generates a false perception in the absence of lines and fails when visual evidence contradicts prior knowledge. Conversely, all Qwen models and Claude models show 0.00 Inducer accuracy on Ebbinghaus, indicating they do not reproduce memorized answers when visual evidence is removed, confirming weaker illusion-specific templates for this case.

\vspace{1mm}
\noindent\textbf{Illusion-Specific Patterns.} Müller-Lyer triggers stronger memory-driven failures than Ebbinghaus. Flagship models (GPT-5, Gemini-2.5-Flash, Claude-Sonnet-4.5) show 99\% Original$\rightarrow$Perturbed drops on Müller-Lyer, far exceeding their 45--60\%  drops on controls, quantifying pure illusion interference. Ebbinghaus exhibits non-monotonic scaling: Claude-Haiku-4.5 achieves 0.96 Perturbed accuracy (best among all models) versus 0.00--0.29 for larger Claude siblings, replicating the main paper's finding that increased capacity amplifies template reliance for certain illusion types.

\vspace{1mm}
\noindent\textbf{Exceptional Cases.} Two models show unique processing: (1)~Qwen3-VL-235B achieves near-perfect accuracy on both Original (1.00) and Perturbed (0.99) Müller-Lyer, with Perturbed accuracy \emph{exceeding} Perturbed Control by +67pp—suggesting reliance on geometric structure rather than semantic templates. (2)~GPT-5-Nano exhibits inverted susceptibility (Original $<$ Perturbed) on both illusions, confirming weak semantic priors make canonical configurations harder than perturbed variants.

\vspace{1mm}
\noindent
\begin{tcolorbox}[
    colback=gray!10,
    colframe=gray!40,
    rounded corners,
    width=\linewidth,
    arc=2mm,
    boxrule=0.5pt,
    left=3pt, right=3pt,
    top=2pt, bottom=2pt
]
\textbf{Key Takeaways}
\begin{itemize}[leftmargin=*, itemsep=1pt, topsep=2pt]
    \item \textbf{Inducer-only = pure hallucination test}: GPT-5/Gemini achieve Inducer=1.00 on Müller-Lyer (answering correctly despite no lines), proving extreme template bias.
    \item \textbf{99\% illusion-specific drops}: Müller-Lyer Original$\rightarrow$Perturbed drops (GPT-5, Sonnet-4.5) far exceed control drops, isolating memory interference.
    \item \textbf{Non-monotonic scaling}: Claude-Haiku-4.5 outperforms larger siblings (0.96 vs. 0.00--0.29 on Ebbinghaus), confirming capacity amplifies template reliance.
    \item \textbf{Illusion heterogeneity}: No model excels universally; Qwen3-VL-235B dominates Müller-Lyer (0.99) but fails Ebbinghaus (0.00).
\end{itemize}
\end{tcolorbox}

\section{Examples of VI-Probe}
This section provides visual examples from VI-Probe, illustrating how illusion-inducing elements (inducers) are manipulated across different data types, complete case-level variations for representative illusions, and visual embedding analyses that validate our perturbation pipeline.

\subsection{Illustration of inducer}
Figure examples below demonstrate how VI-Probe systematically manipulates illusion-inducing elements across Original, Perturbed, and Control conditions to isolate the specific contribution of each visual factor.

\onecolumn

\begin{figure}[H]
\centering
\includegraphics[width=\linewidth]{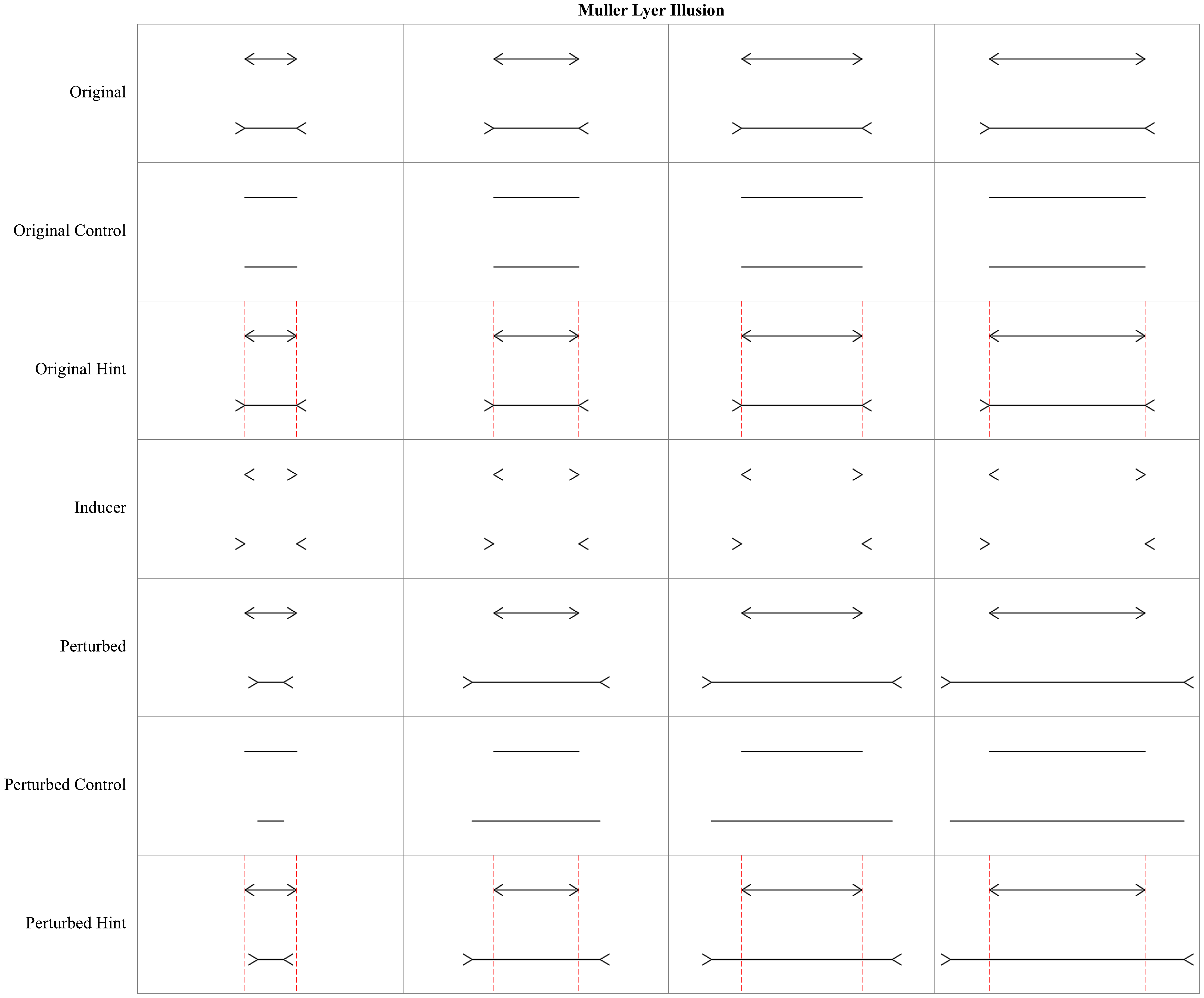}
\label{fig:p_strenth-c}
\end{figure}

\begin{figure}[H]
\centering
\includegraphics[width=\linewidth]{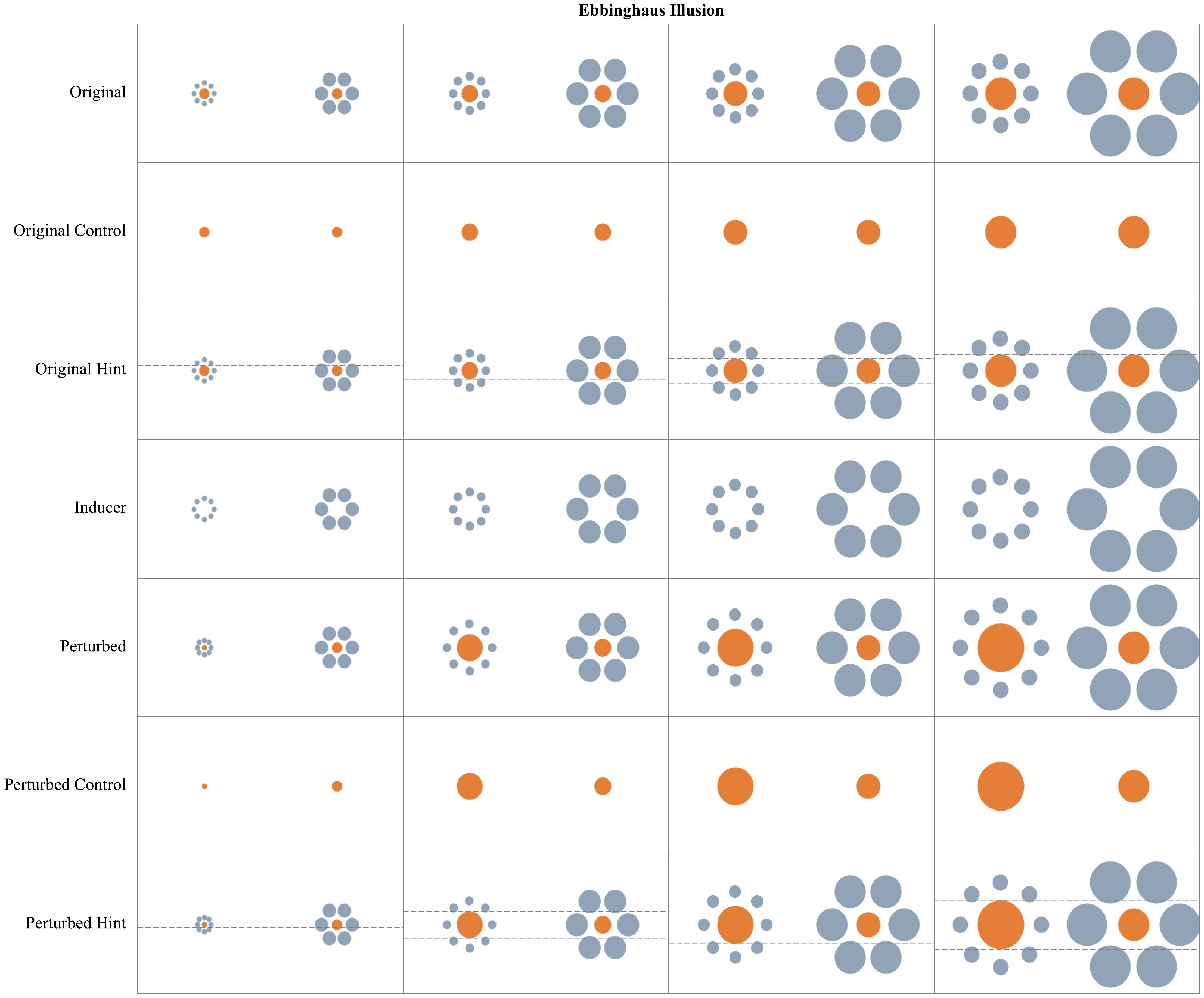}
\label{fig:p_strenth-c}
\end{figure}

\subsection{Visual examples of VI-Probe}
\begin{figure}[H]
\centering
\includegraphics[width=0.81\linewidth]{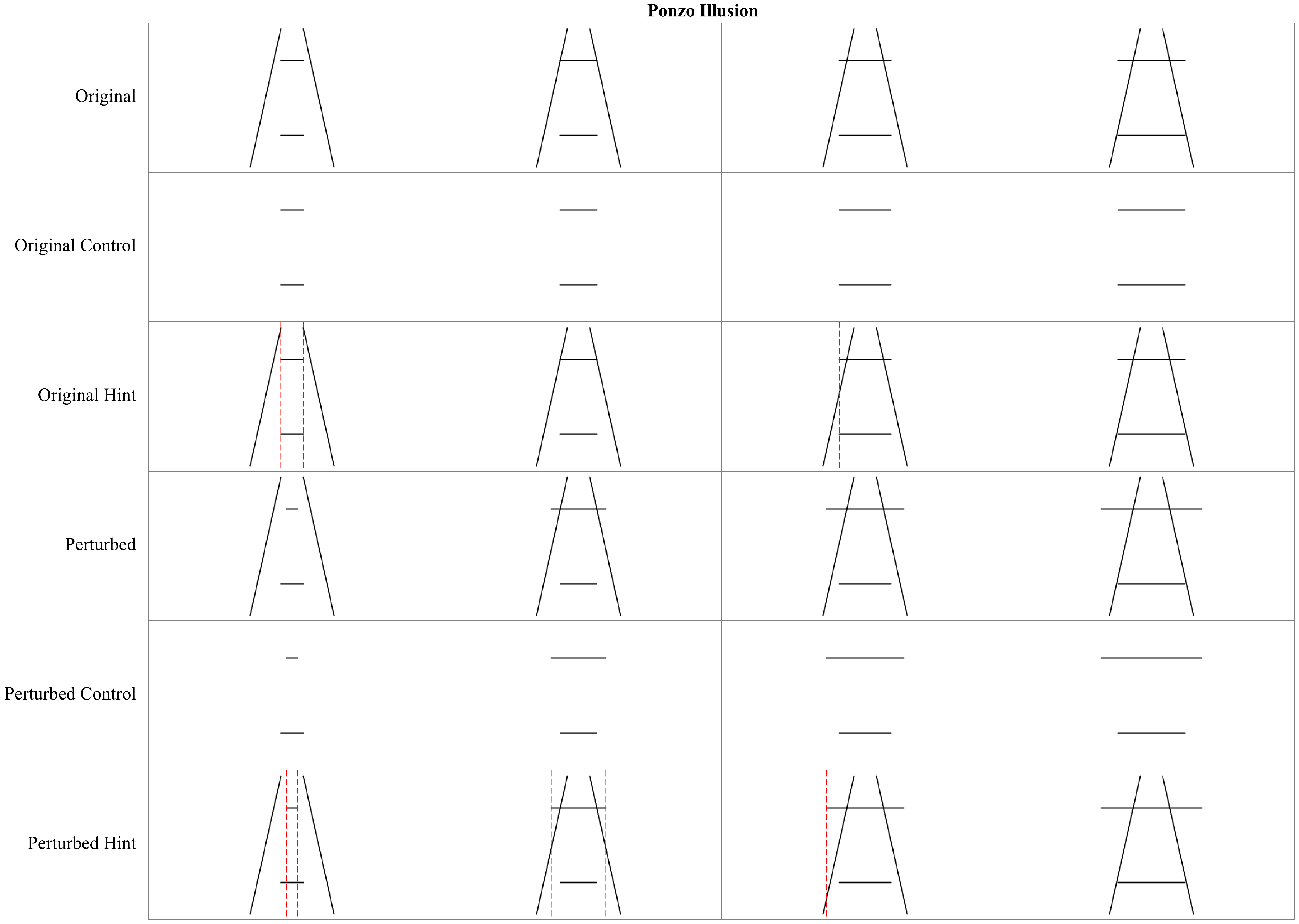}
\label{fig:p_strenth-c}
\end{figure}

\begin{figure}[H]
\centering
\includegraphics[width=0.81\linewidth]{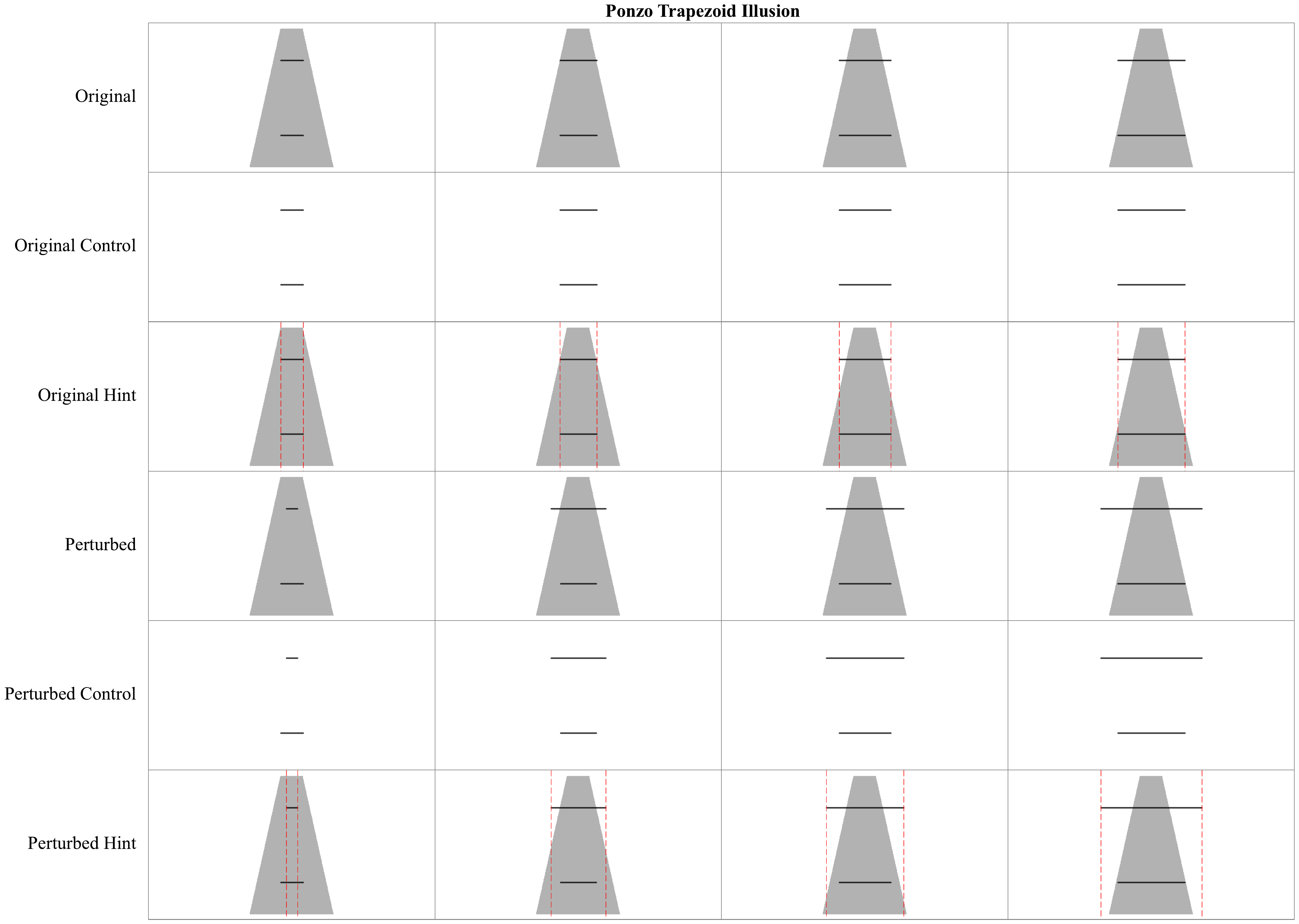}
\label{fig:p_strenth-c}
\end{figure}

\begin{figure}[H]
\centering
\includegraphics[width=0.81\linewidth]{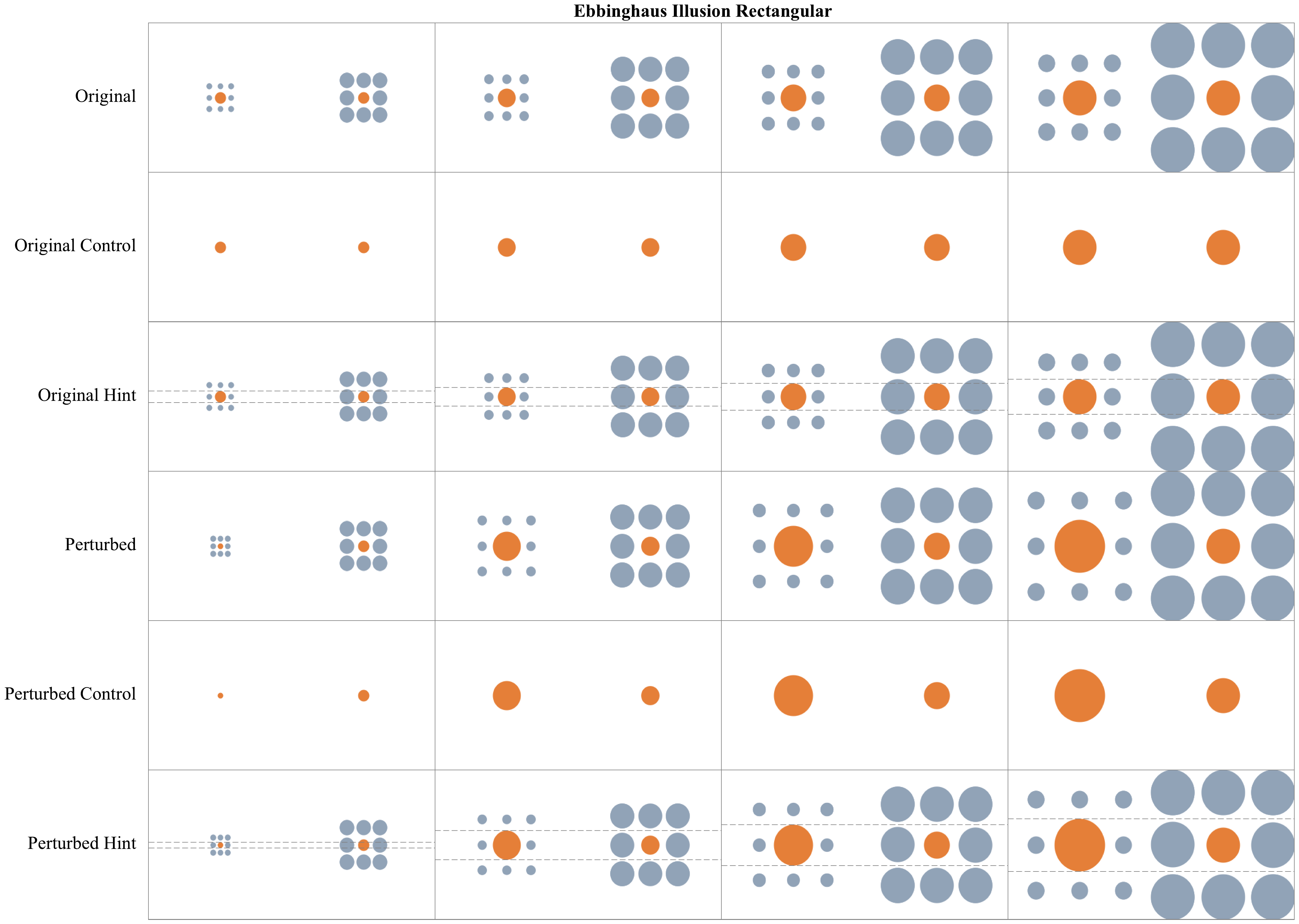}
\label{fig:p_strenth-c}
\end{figure}

\begin{figure}[H]
\centering
\includegraphics[width=0.81\linewidth]{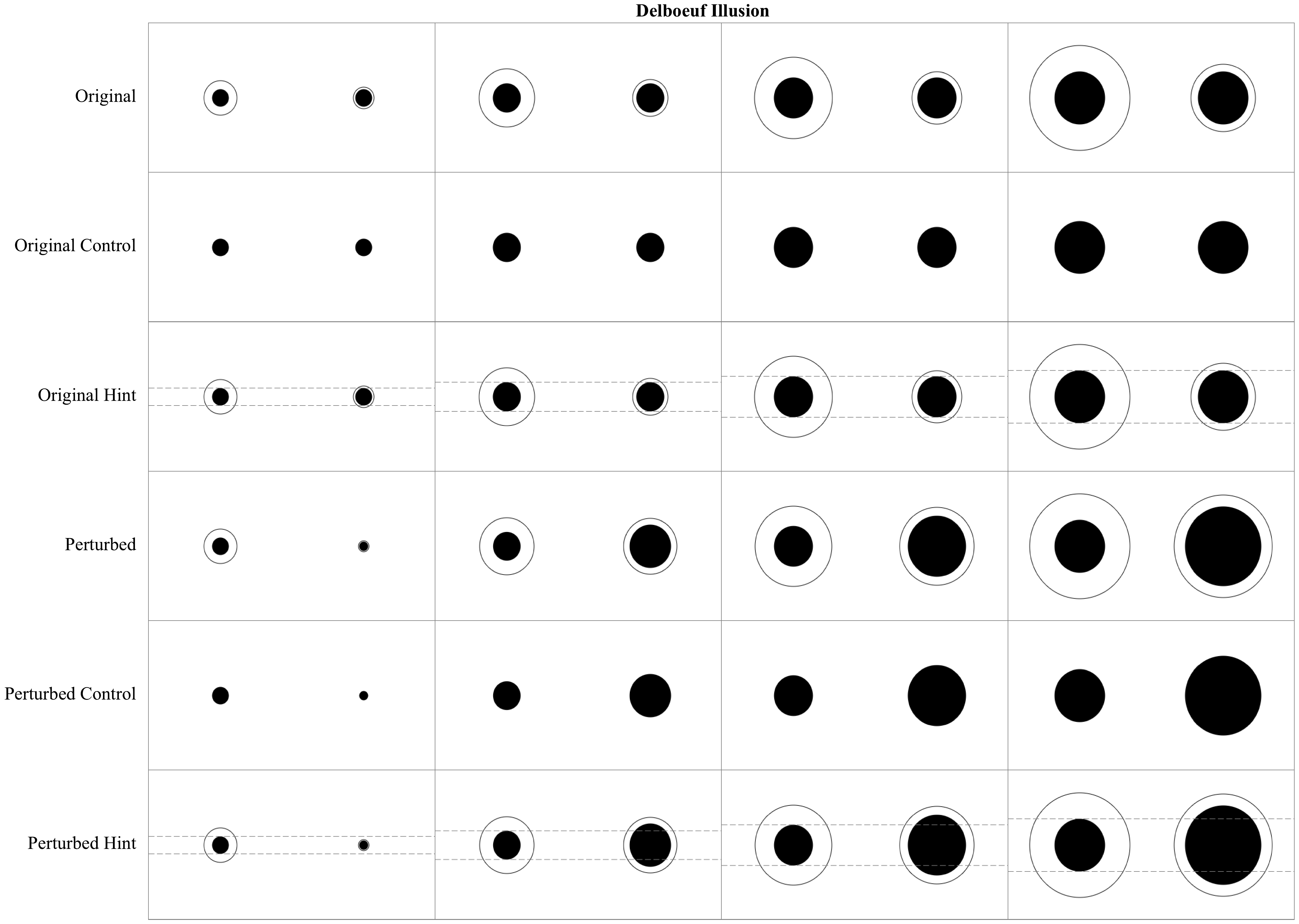}
\label{fig:p_strenth-c}
\end{figure}

\begin{figure}[H]
\centering
\includegraphics[width=0.81\linewidth]{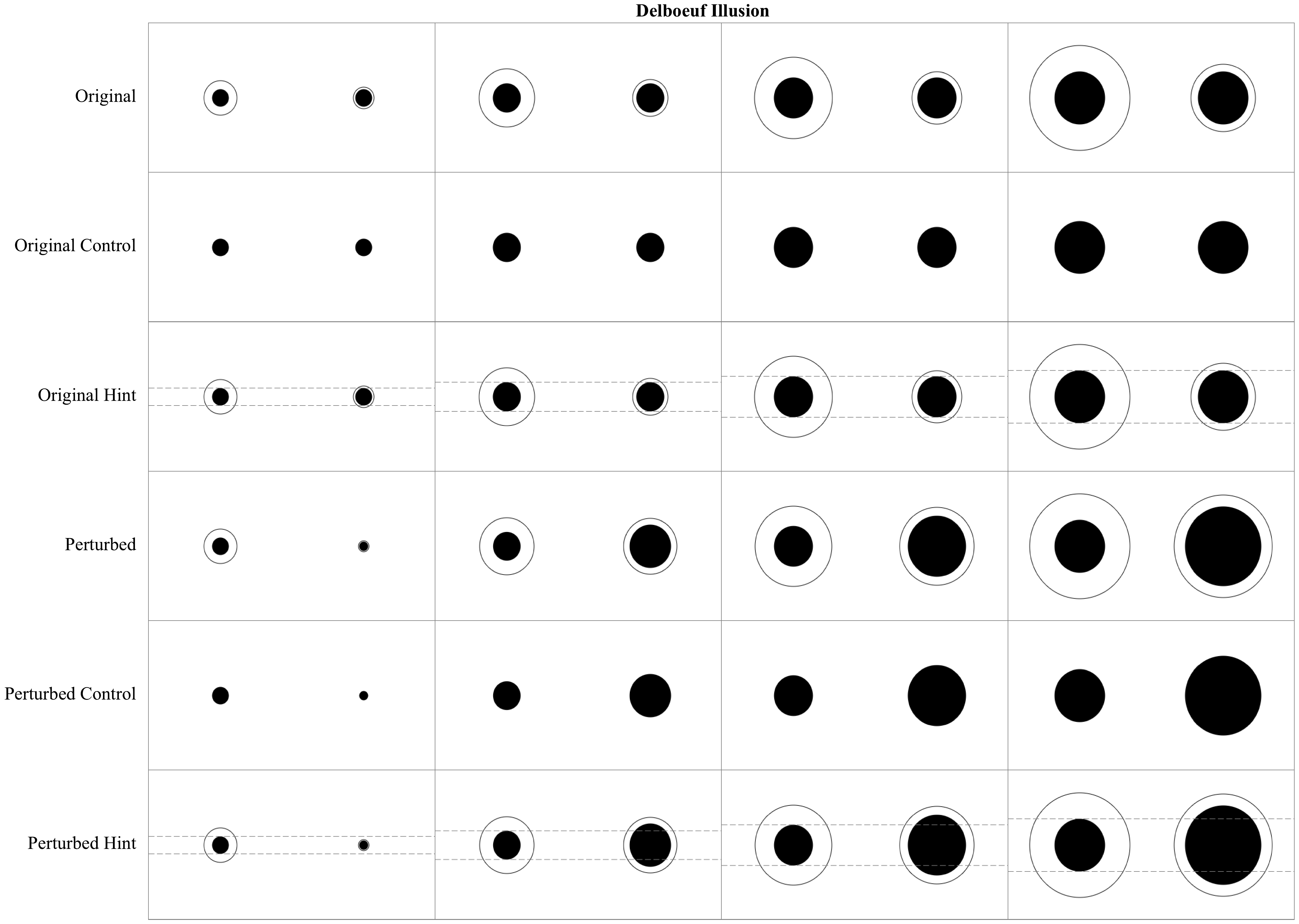}
\label{fig:p_strenth-c}
\end{figure}

\begin{figure}[H]
\centering
\includegraphics[width=0.81\linewidth]{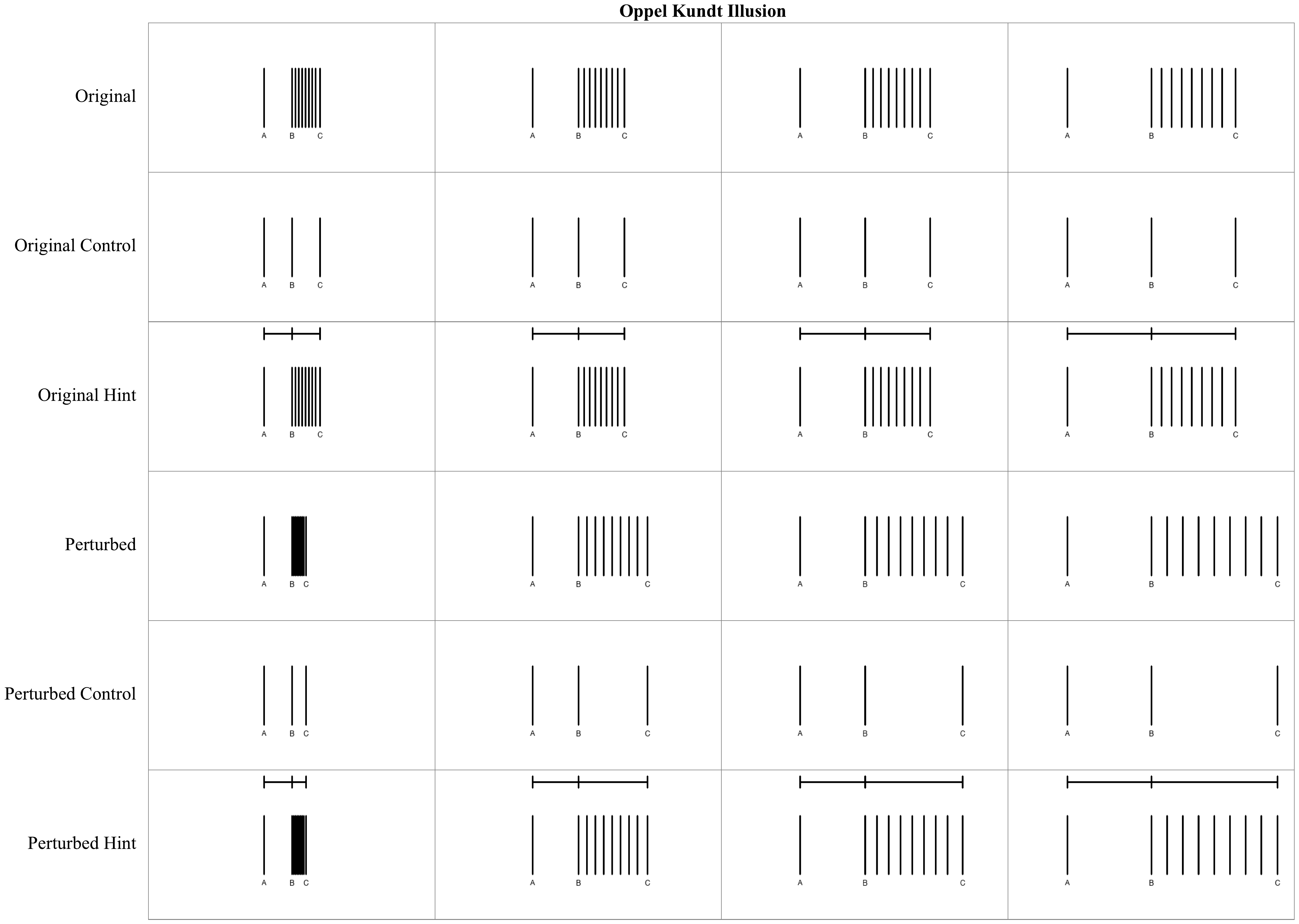}
\label{fig:p_strenth-c}
\end{figure}

\begin{figure}[H]
\centering
\includegraphics[width=0.81\linewidth]{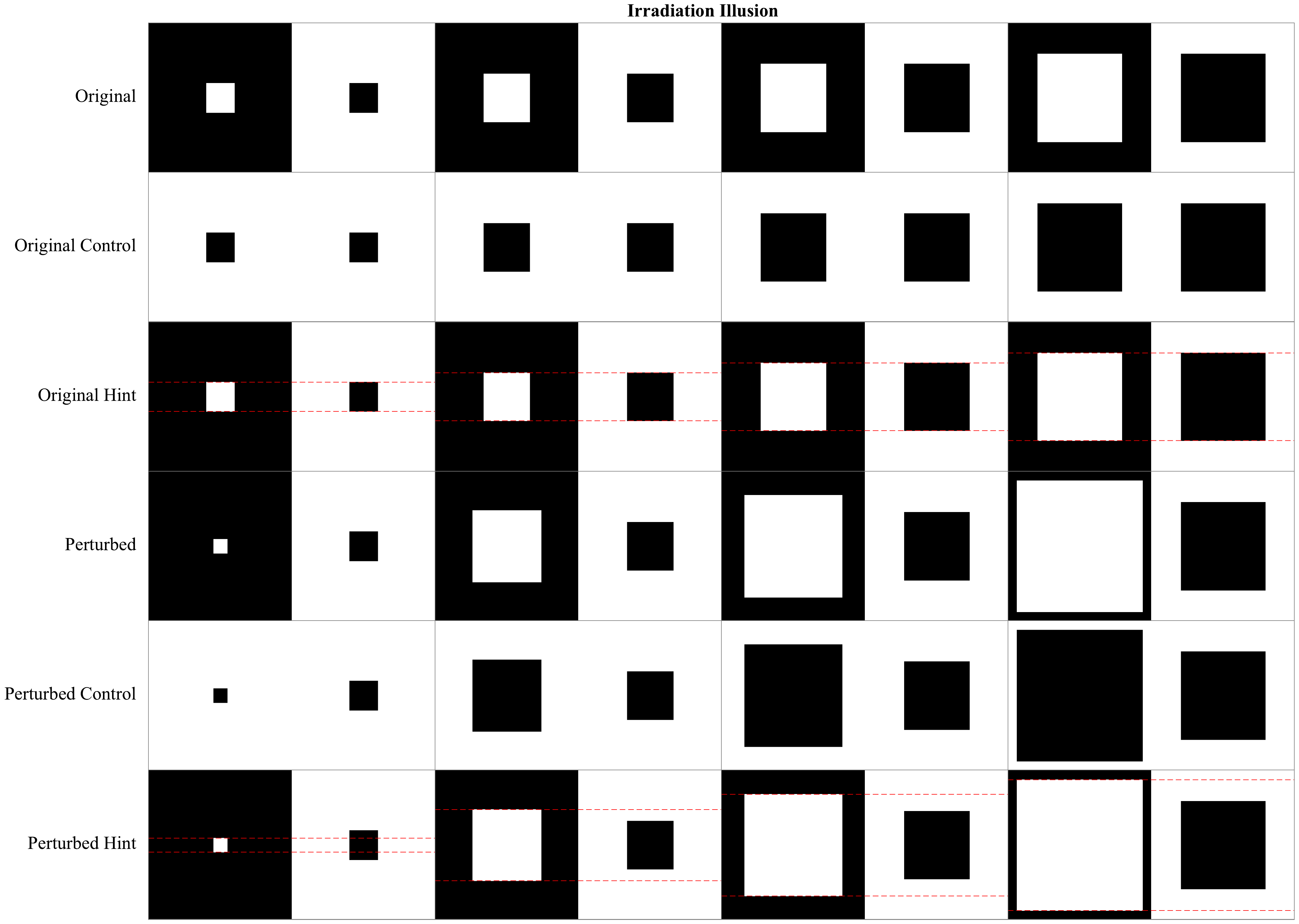}
\label{fig:p_strenth-c}
\end{figure}

\begin{figure}[H]
\centering
\includegraphics[width=0.81\linewidth]{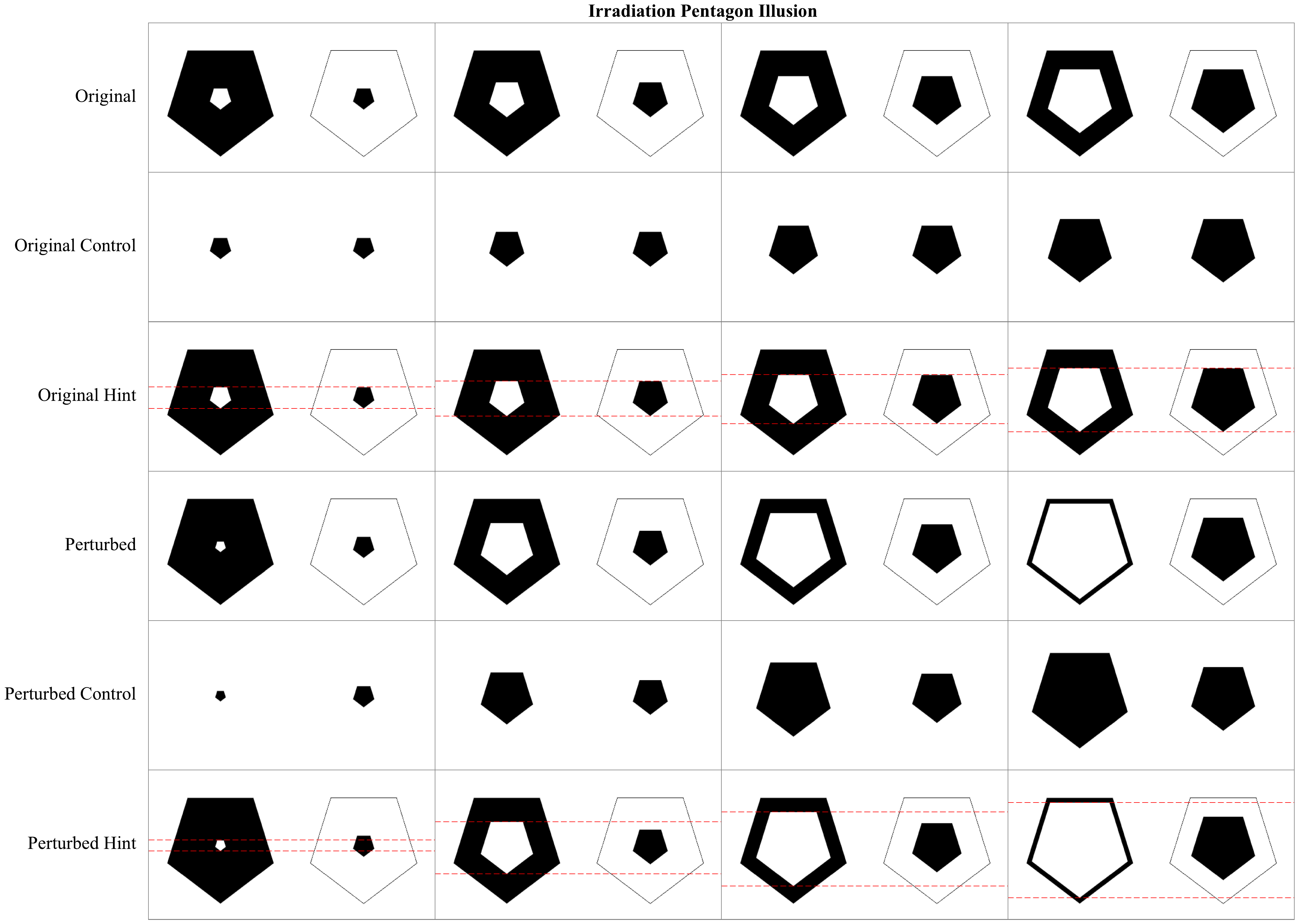}
\label{fig:p_strenth-c}
\end{figure}

\begin{figure}[H]
\centering
\includegraphics[width=0.81\linewidth]{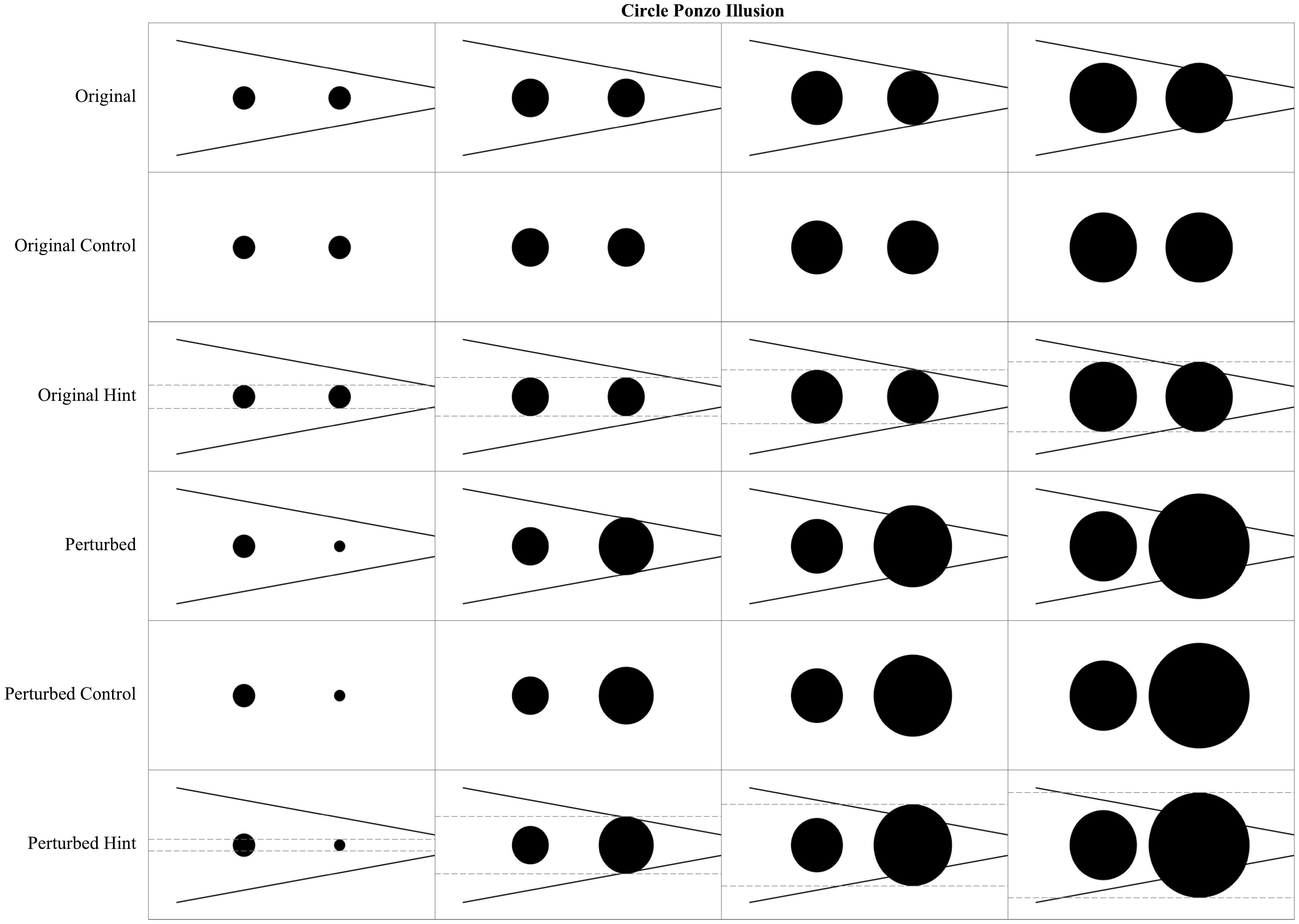}
\label{fig:p_strenth-c}
\end{figure}

\begin{figure}[H]
\centering
\includegraphics[width=0.81\linewidth]{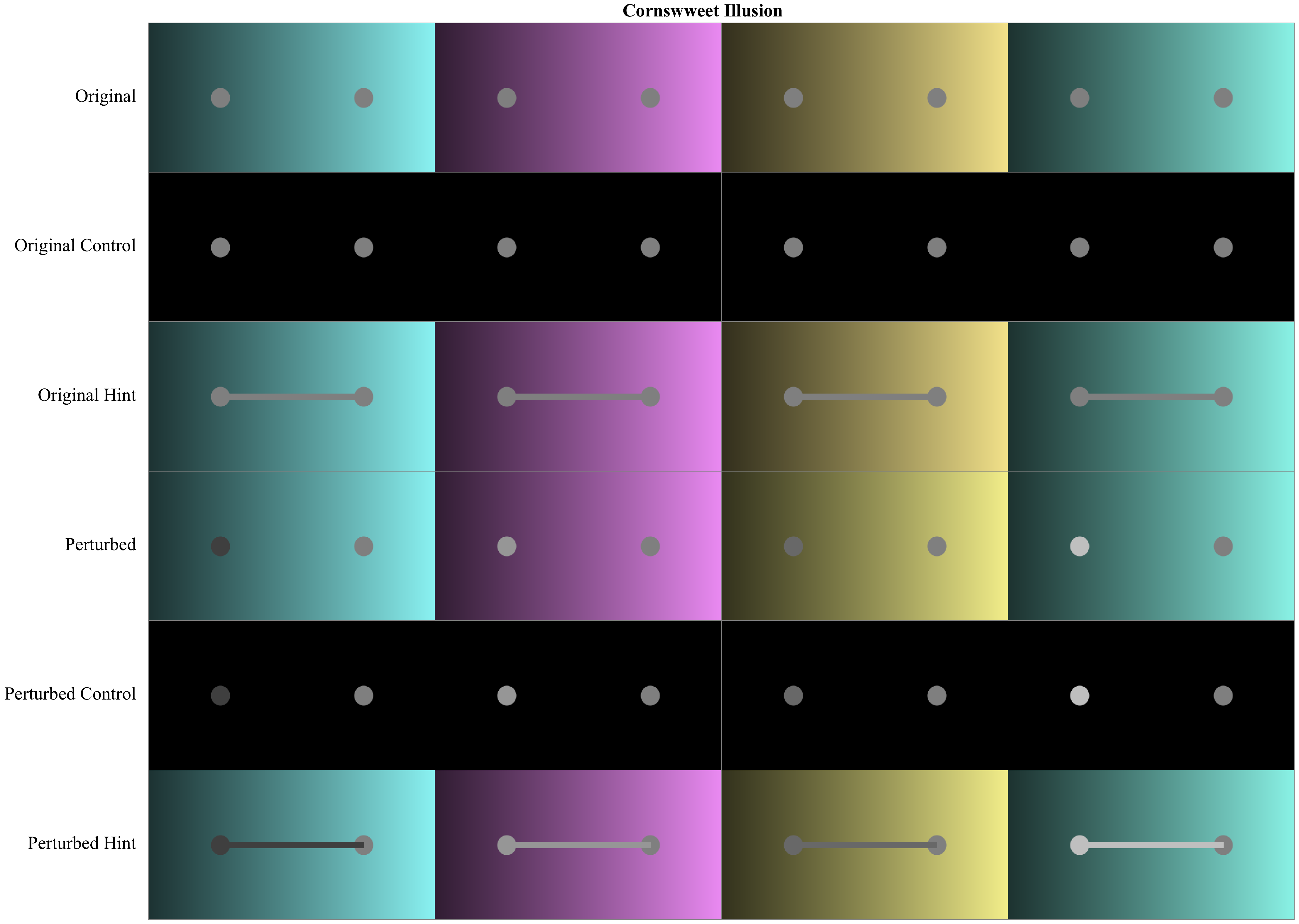}
\label{fig:p_strenth-c}
\end{figure}

\begin{figure}[H]
\centering
\includegraphics[width=0.81\linewidth]{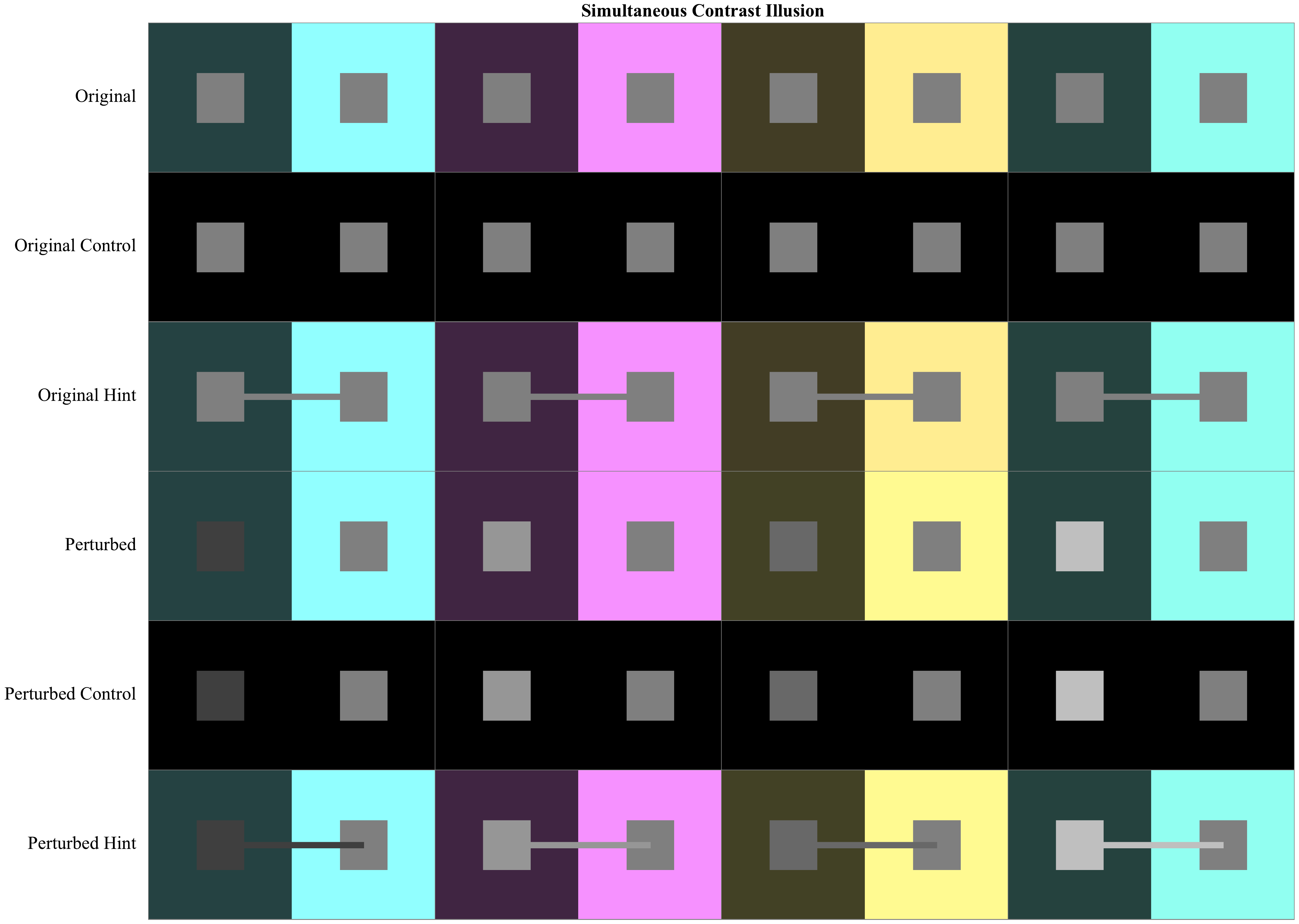}
\label{fig:p_strenth-c}
\end{figure}

\begin{figure}[H]
\centering
\includegraphics[width=0.81\linewidth]{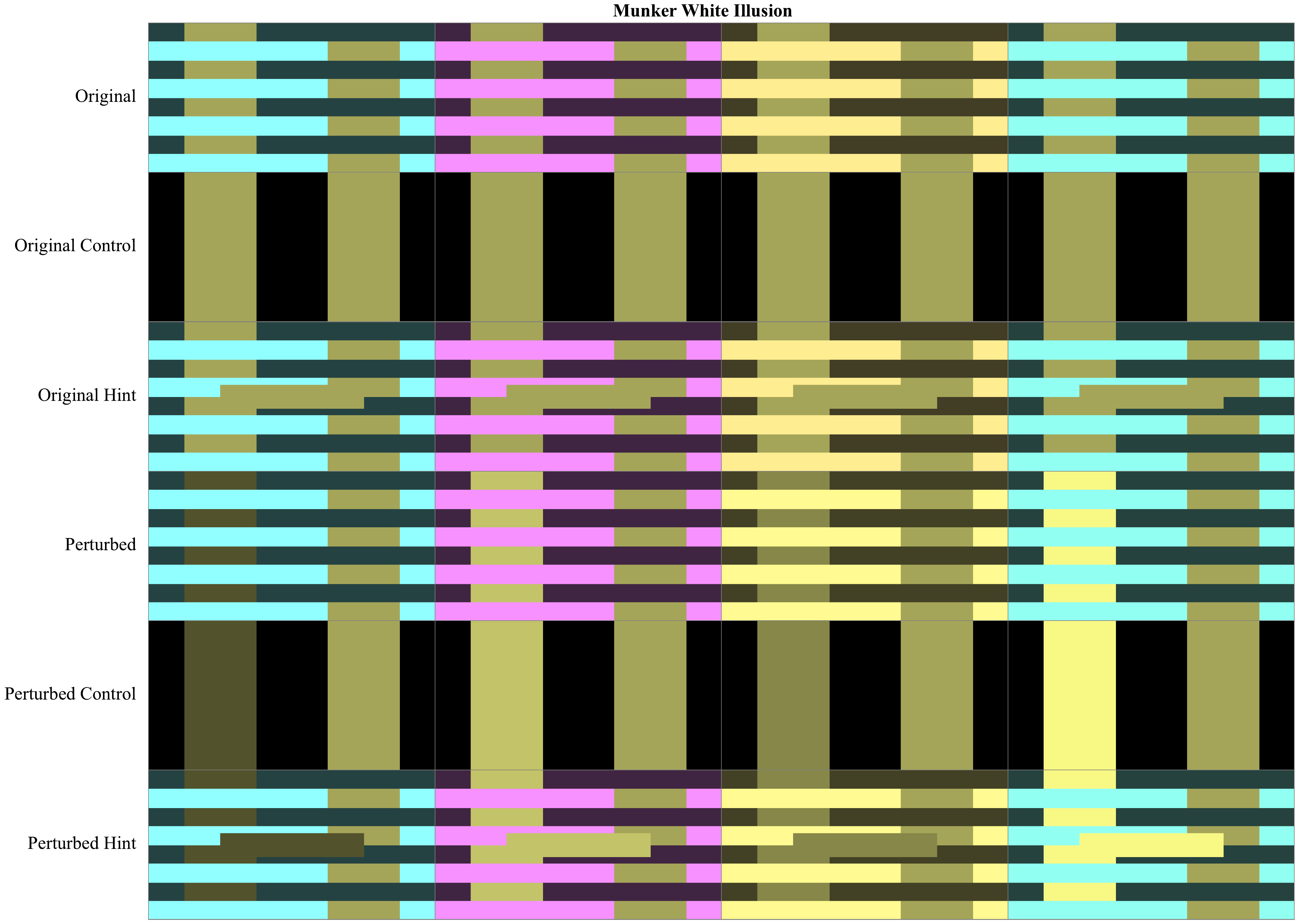}
\label{fig:p_strenth-c}
\end{figure}

\begin{figure}[H]
\centering
\includegraphics[width=0.81\linewidth]{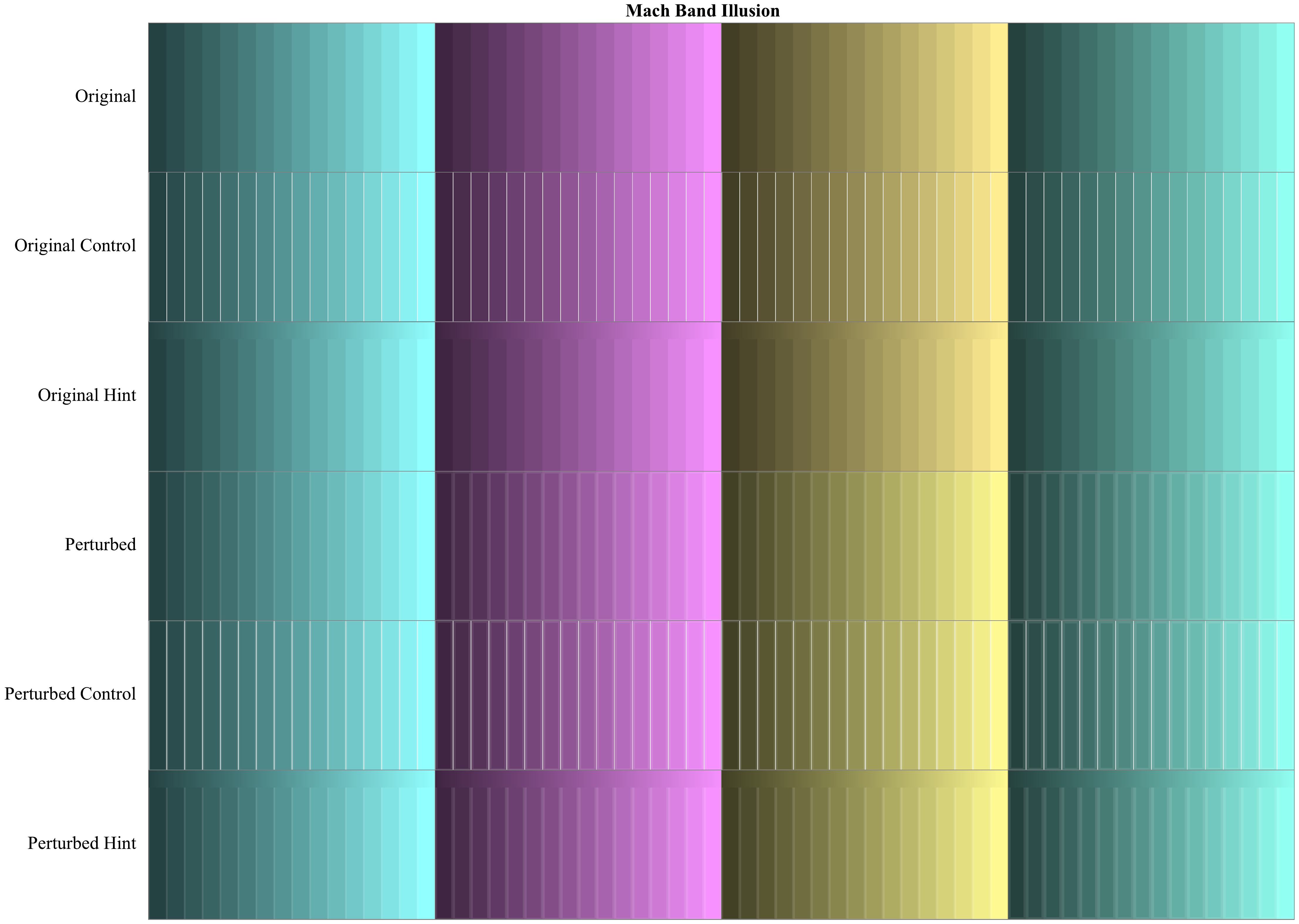}
\label{fig:p_strenth-c}
\end{figure}

\begin{figure}[H]
\centering
\includegraphics[width=0.81\linewidth]{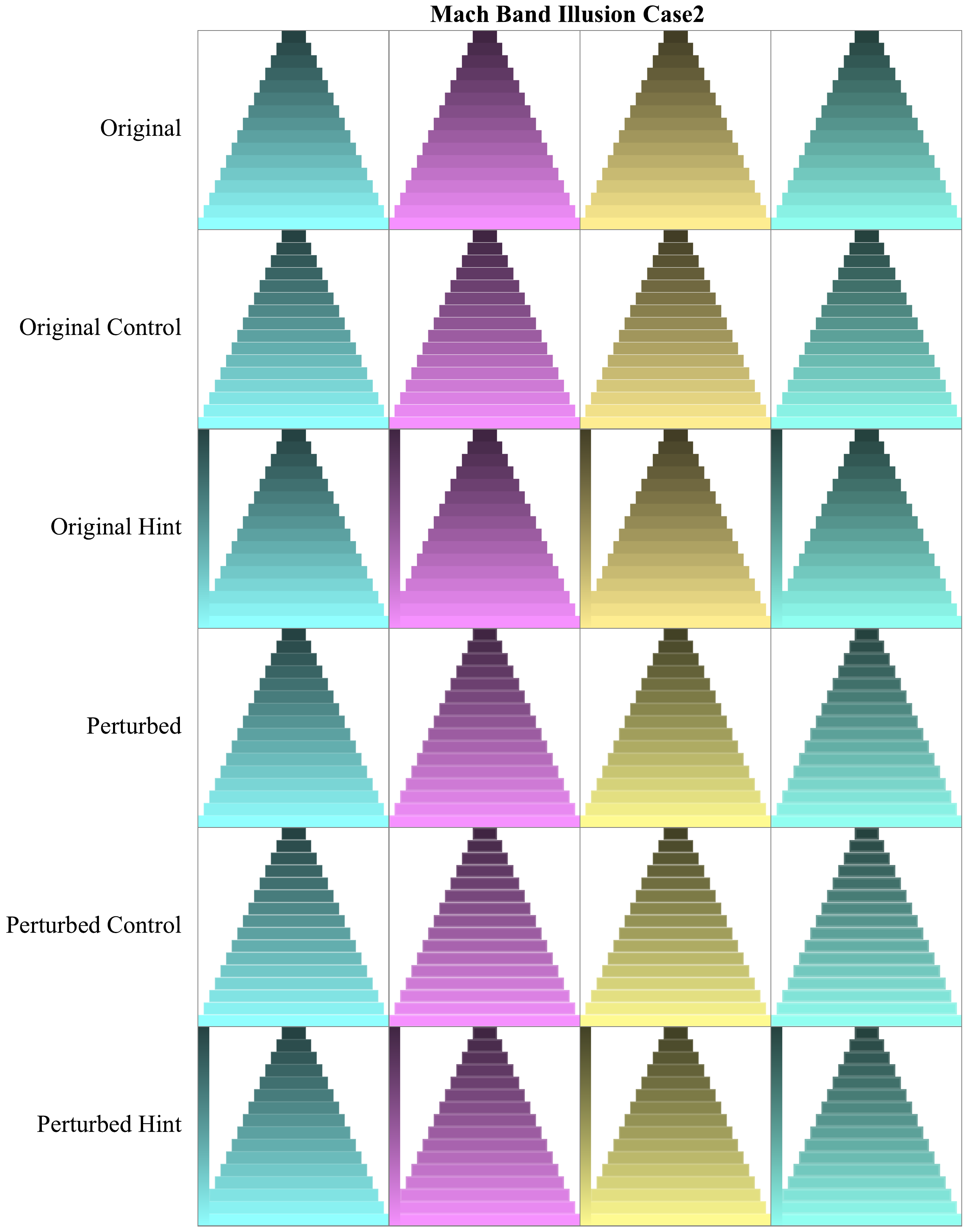}
\label{fig:p_strenth-c}
\end{figure}

\begin{figure}[H]
\centering
\includegraphics[width=0.81\linewidth]{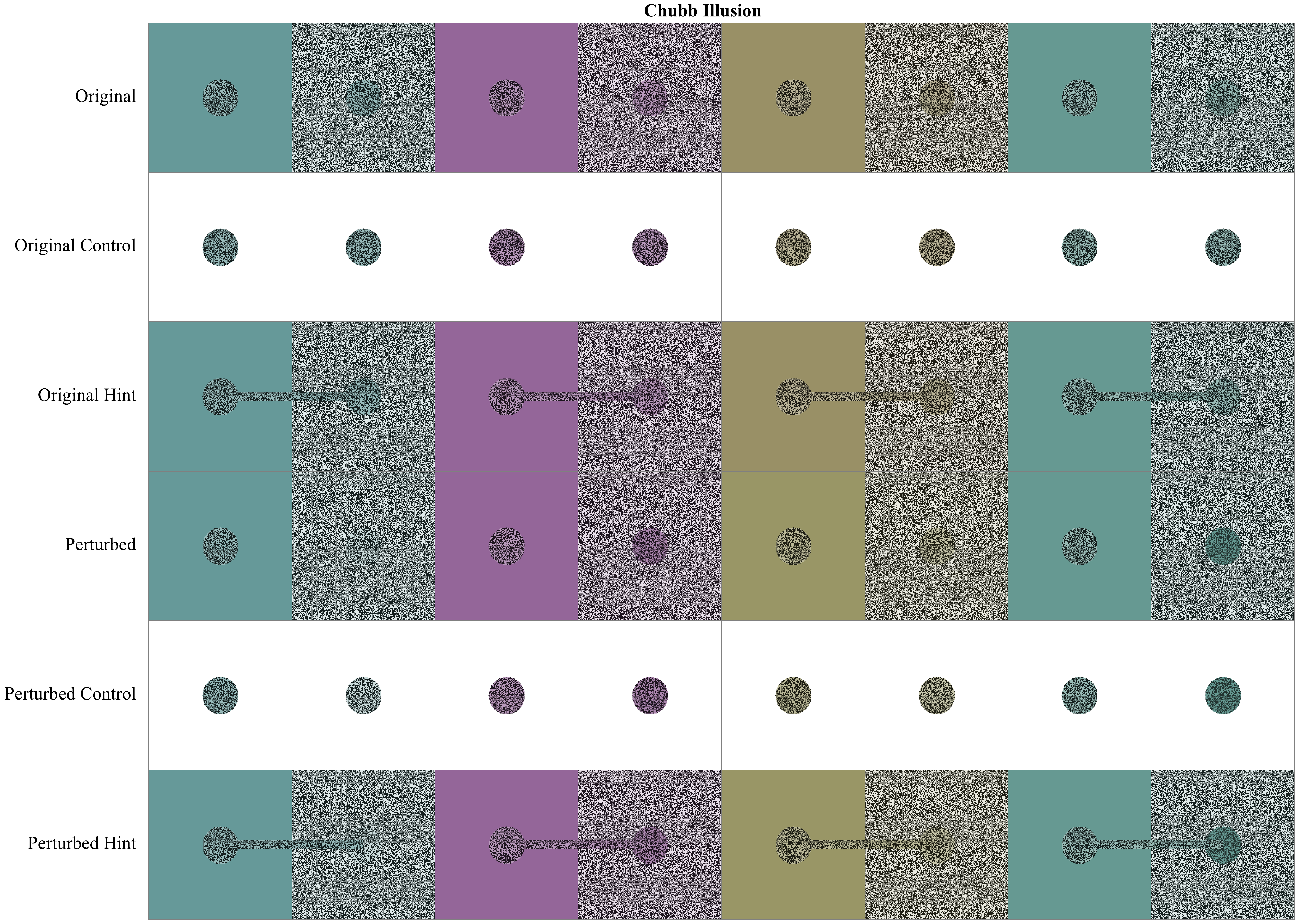}
\label{fig:p_strenth-c}
\end{figure}

\begin{figure}[H]
\centering
\includegraphics[width=0.81\linewidth]{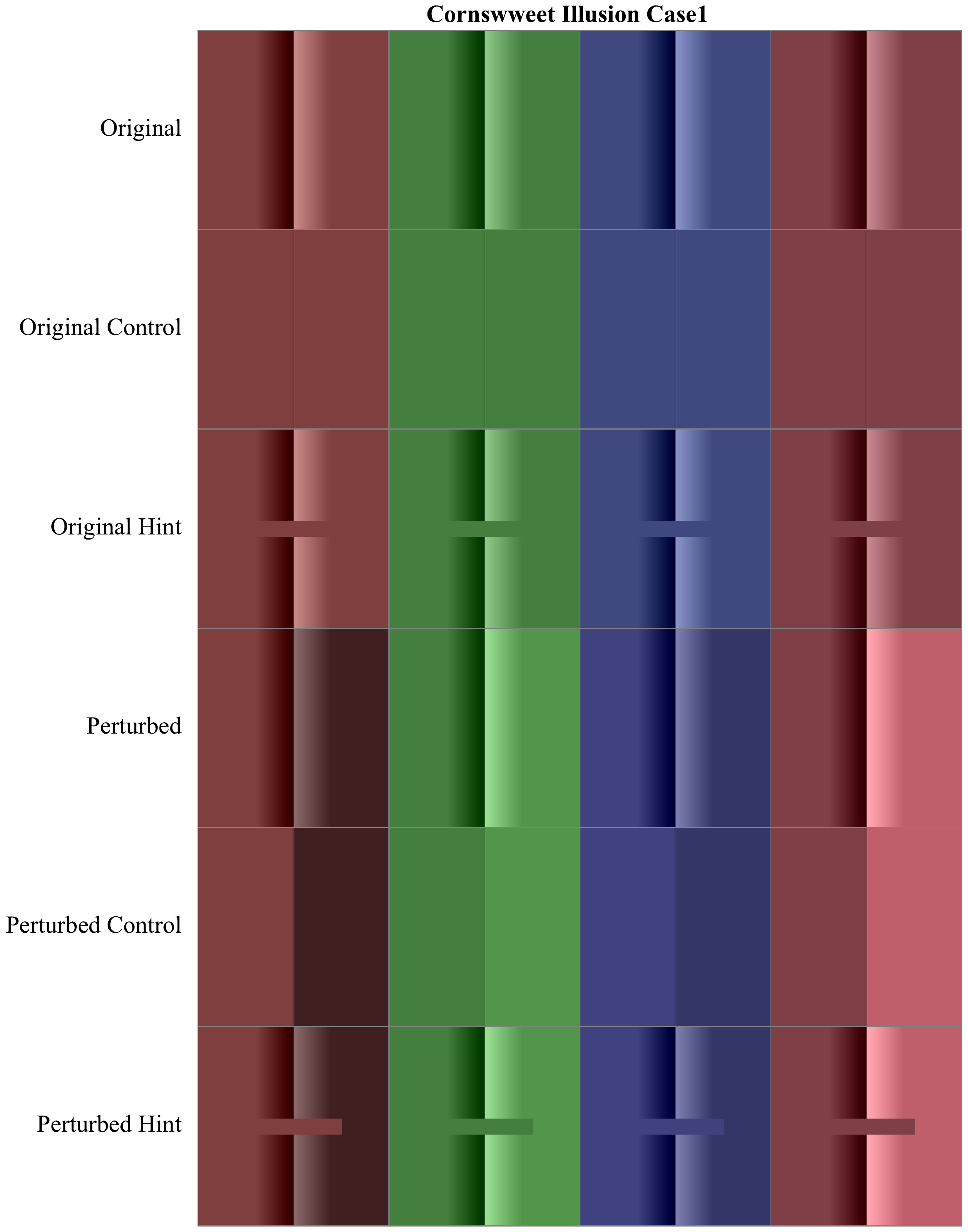}
\label{fig:p_strenth-c}
\end{figure}

\begin{figure}[H]
\centering
\includegraphics[width=0.81\linewidth]{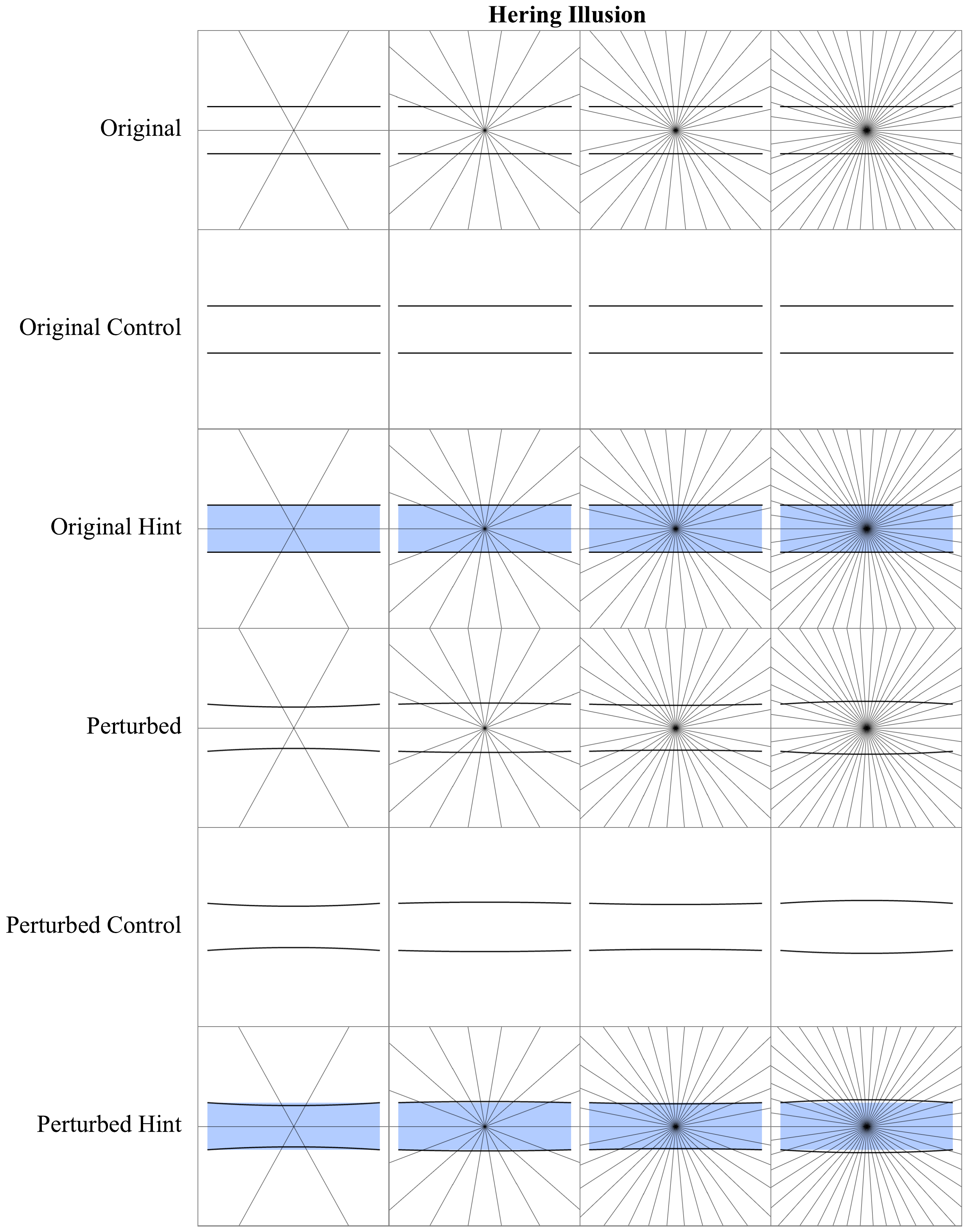}
\label{fig:p_strenth-c}
\end{figure}

\begin{figure}[H]
\centering
\includegraphics[width=0.81\linewidth]{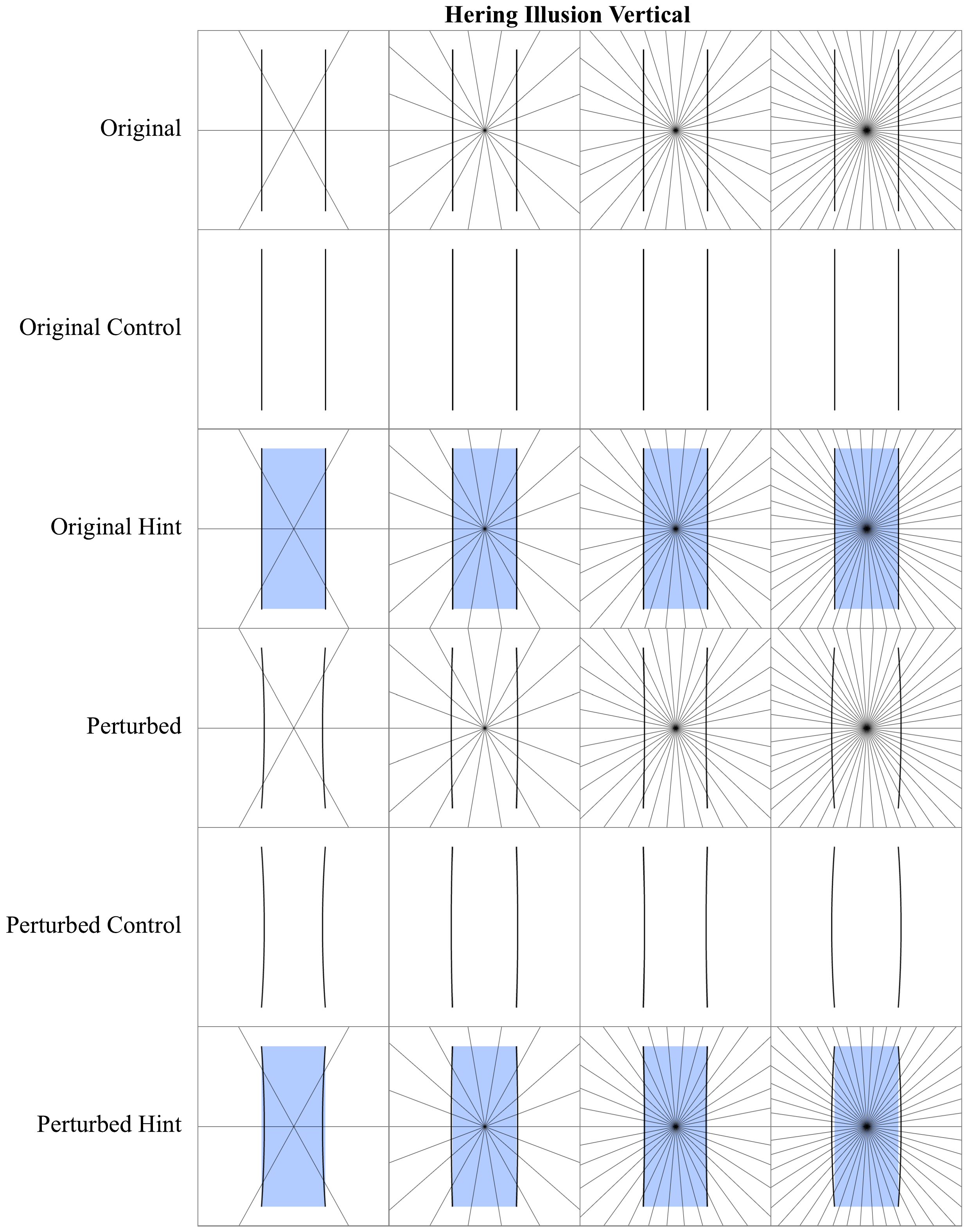}
\label{fig:p_strenth-c}
\end{figure}

\begin{figure}[H]
\centering
\includegraphics[width=0.81\linewidth]{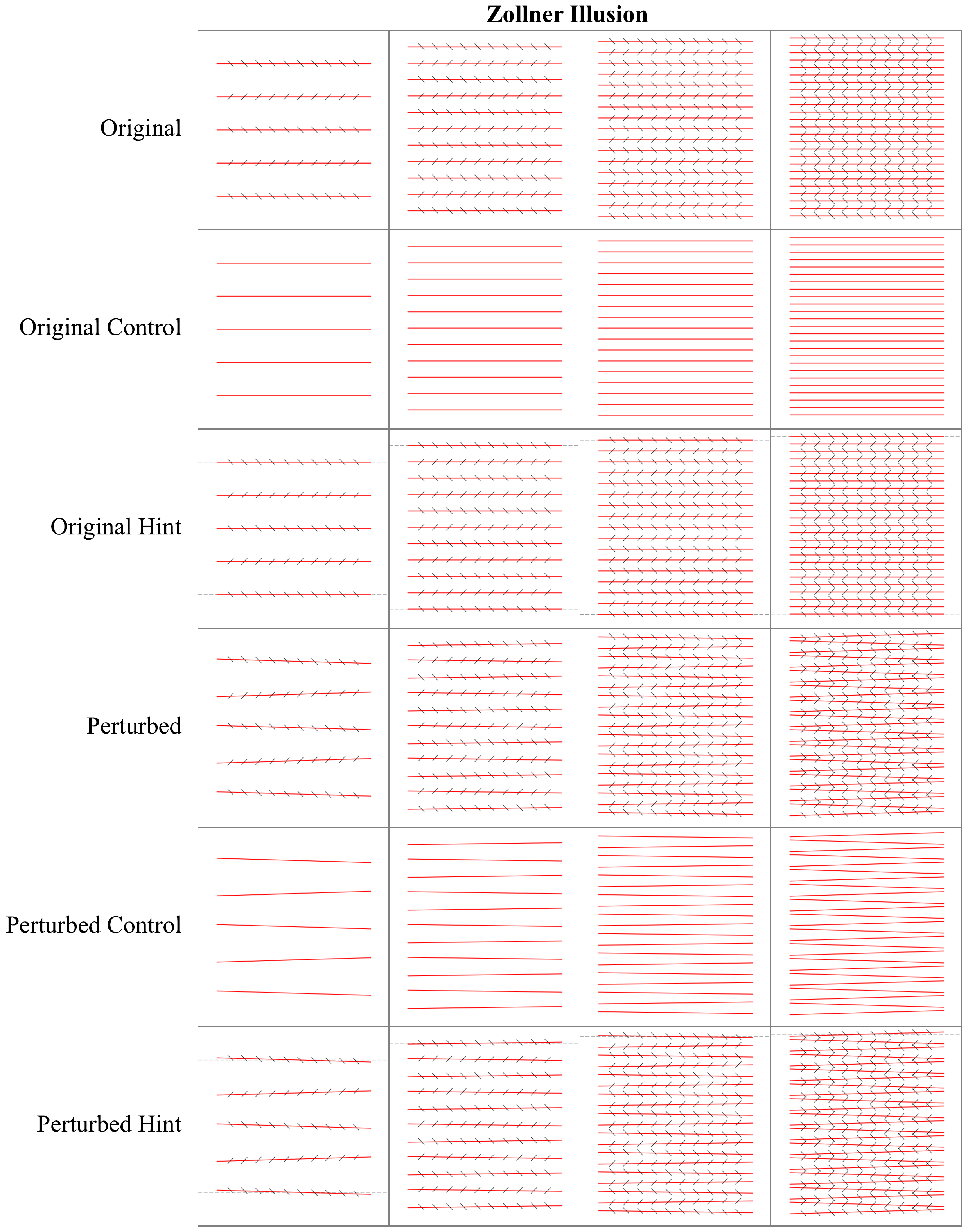}
\label{fig:p_strenth-c}
\end{figure}

\begin{figure}[H]
\centering
\includegraphics[width=0.81\linewidth]{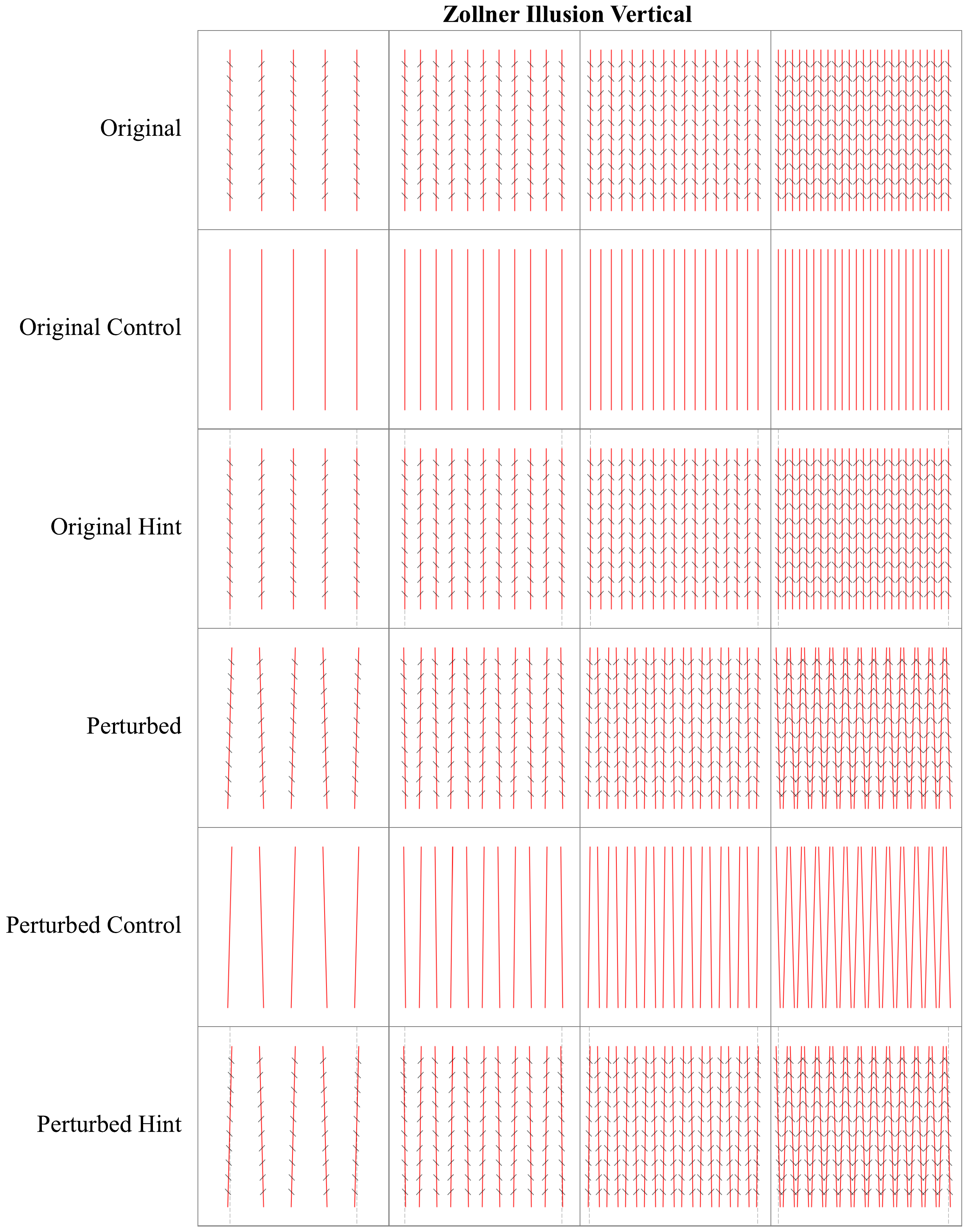}
\label{fig:p_strenth-c}
\end{figure}

\begin{figure}[H]
\centering
\includegraphics[width=0.81\linewidth]{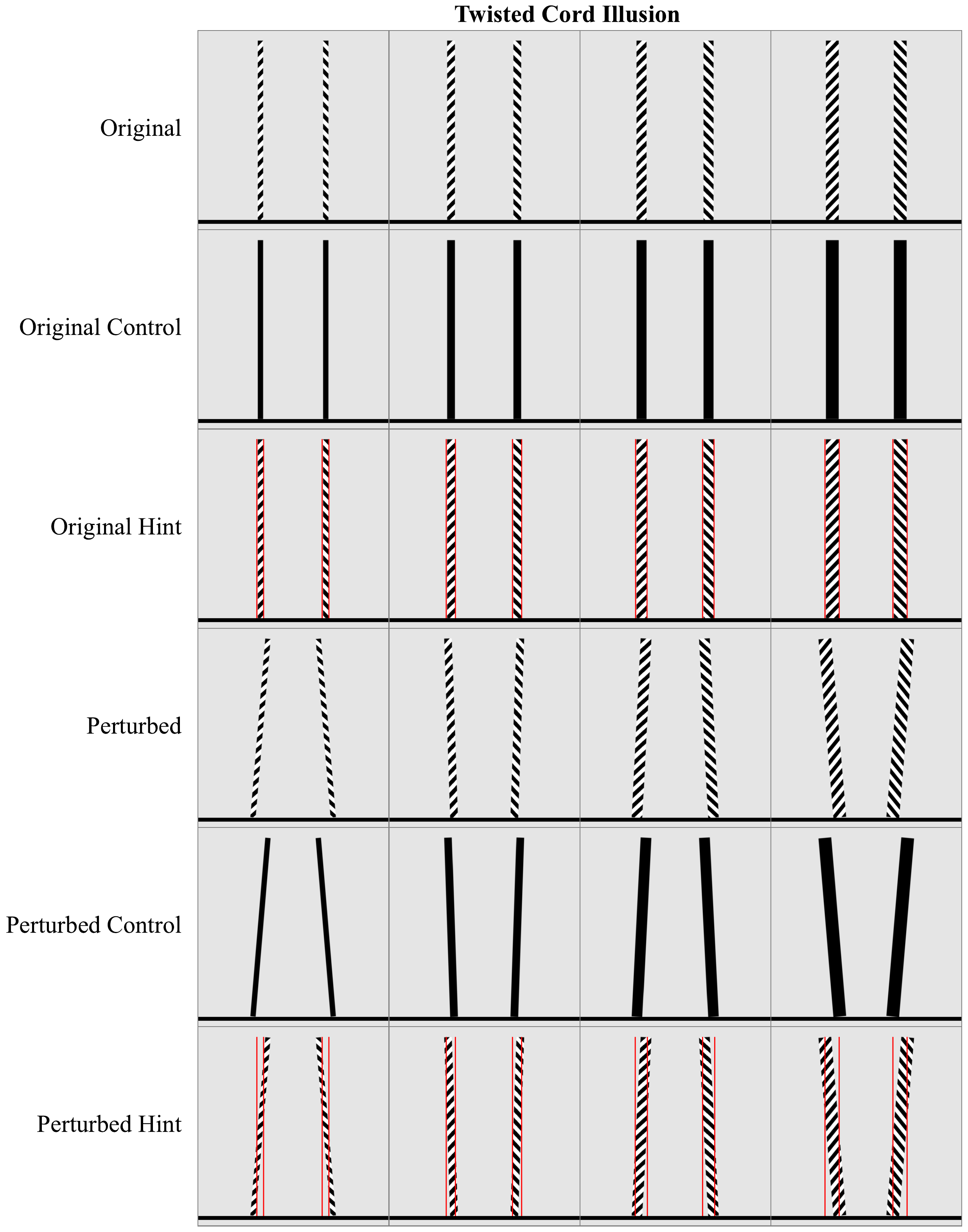}
\label{fig:p_strenth-c}
\end{figure}

\begin{figure}[H]
\centering
\includegraphics[width=0.81\linewidth]{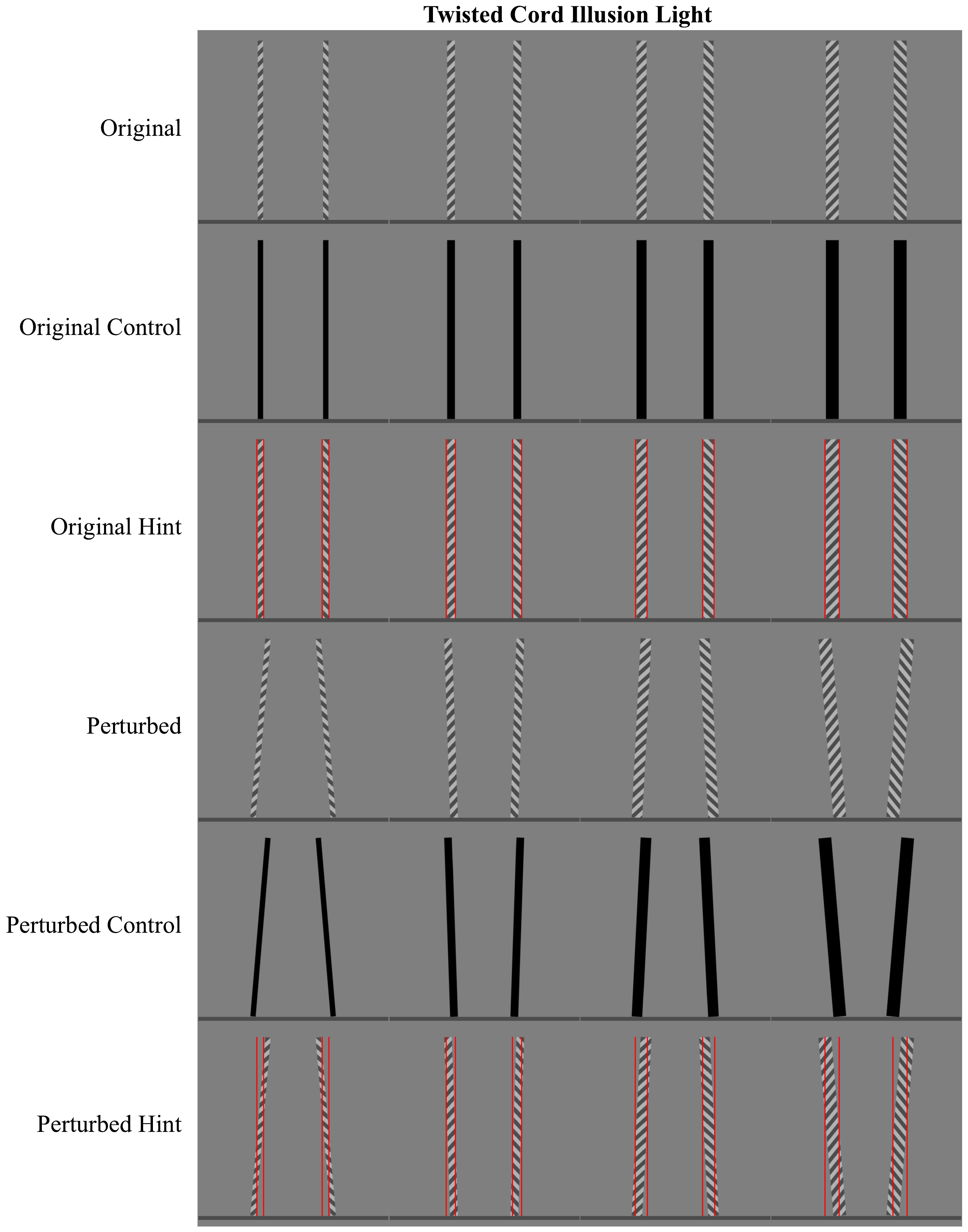}
\label{fig:p_strenth-c}
\end{figure}

\begin{figure}[H]
\centering
\includegraphics[width=0.81\linewidth]{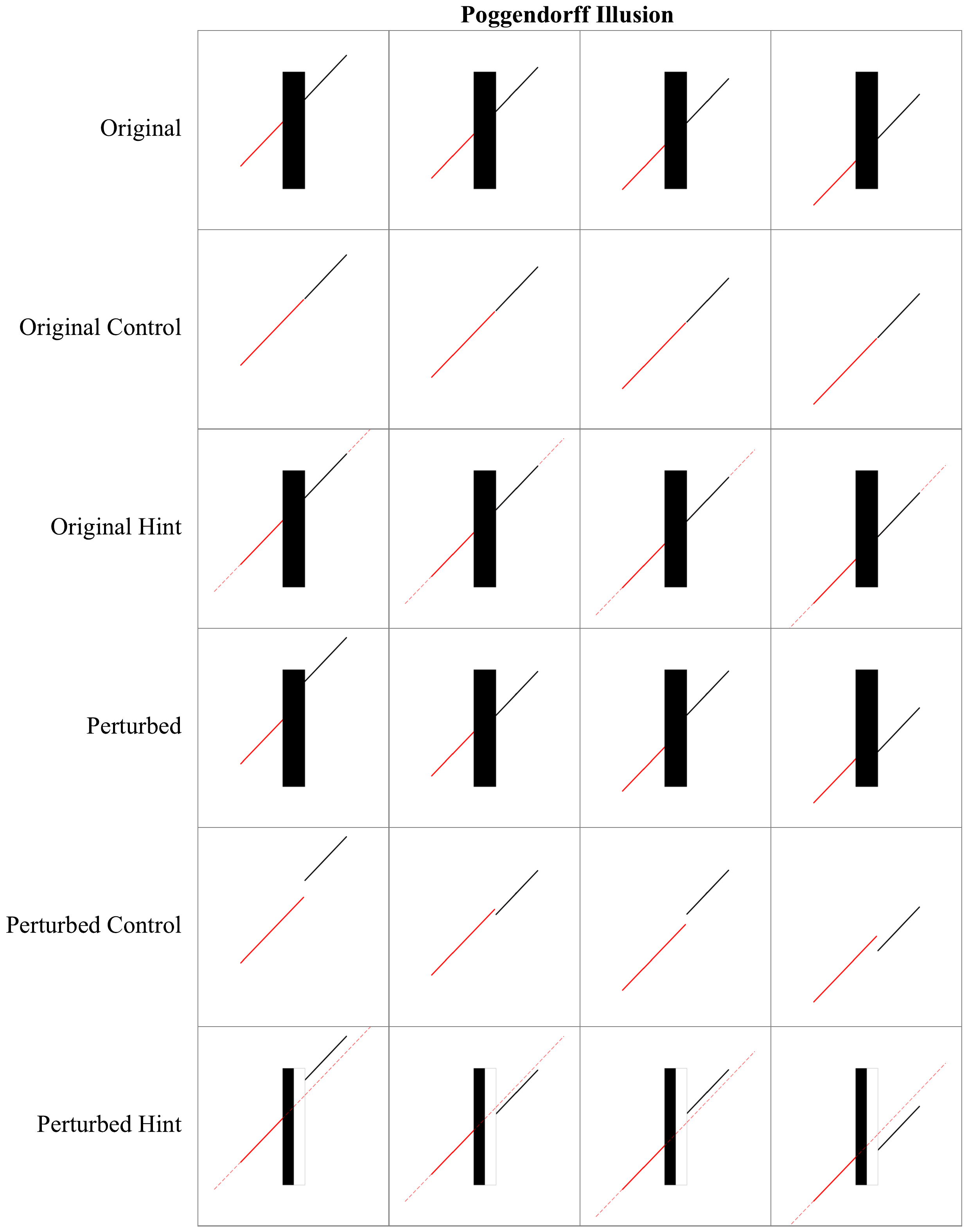}
\label{fig:p_strenth-c}
\end{figure}

\begin{figure}[H]
\centering
\includegraphics[width=0.81\linewidth]{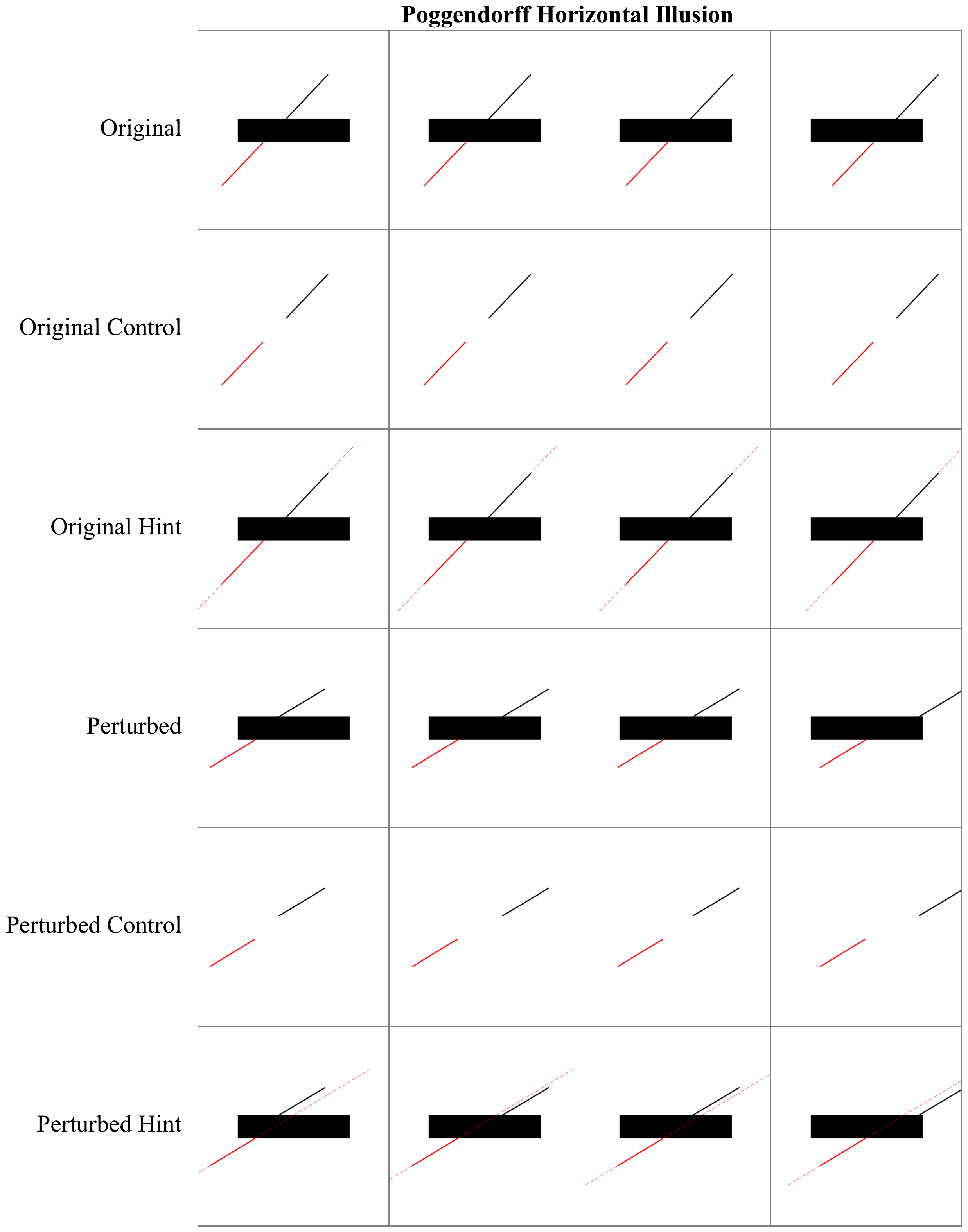}
\label{fig:p_strenth-c}
\end{figure}

\begin{figure}[H]
\centering
\includegraphics[width=0.81\linewidth]{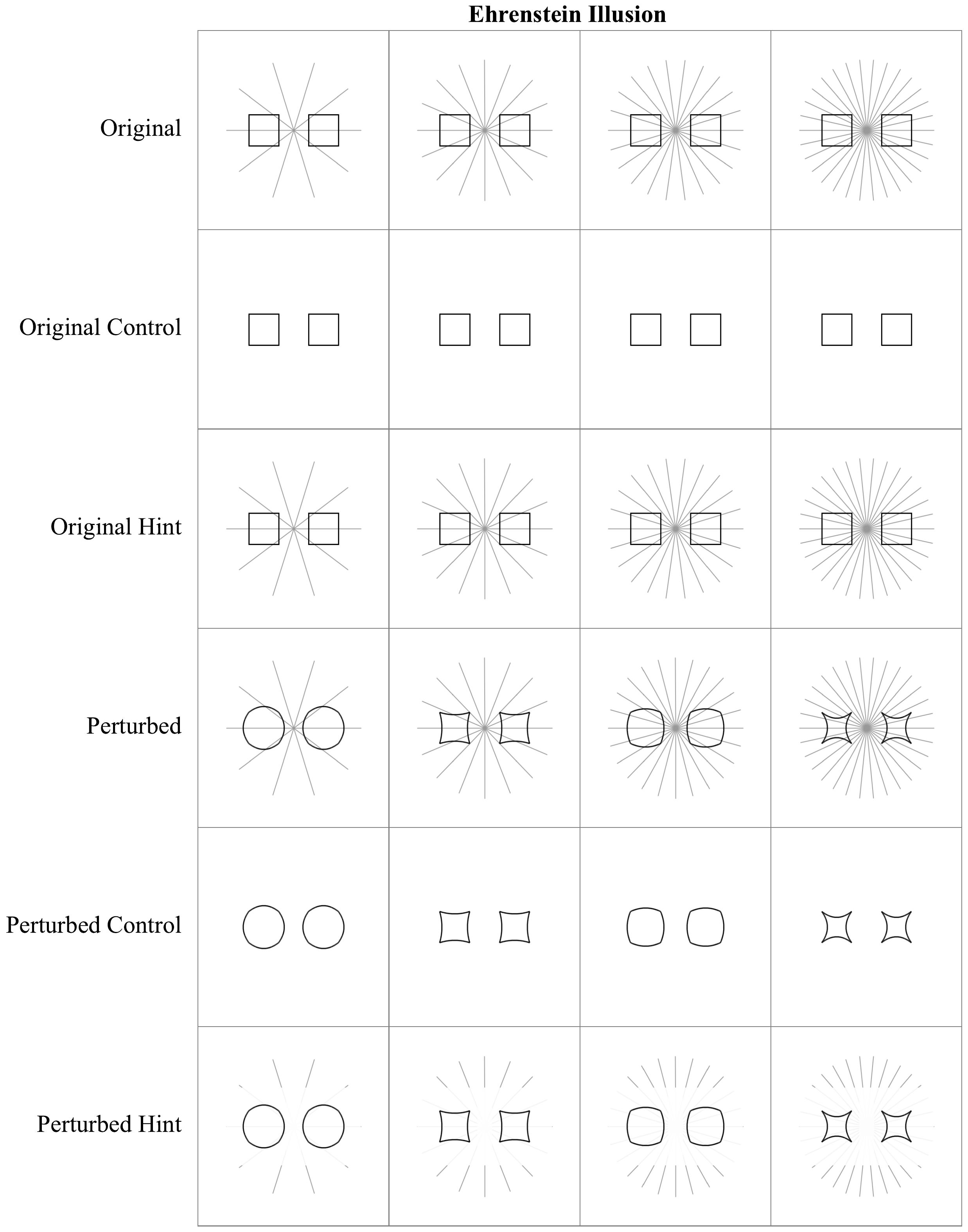}
\label{fig:p_strenth-c}
\end{figure}

\subsection{More examples of visual embeddings}

\begin{figure}[H]
\centering
\includegraphics[width=\linewidth]{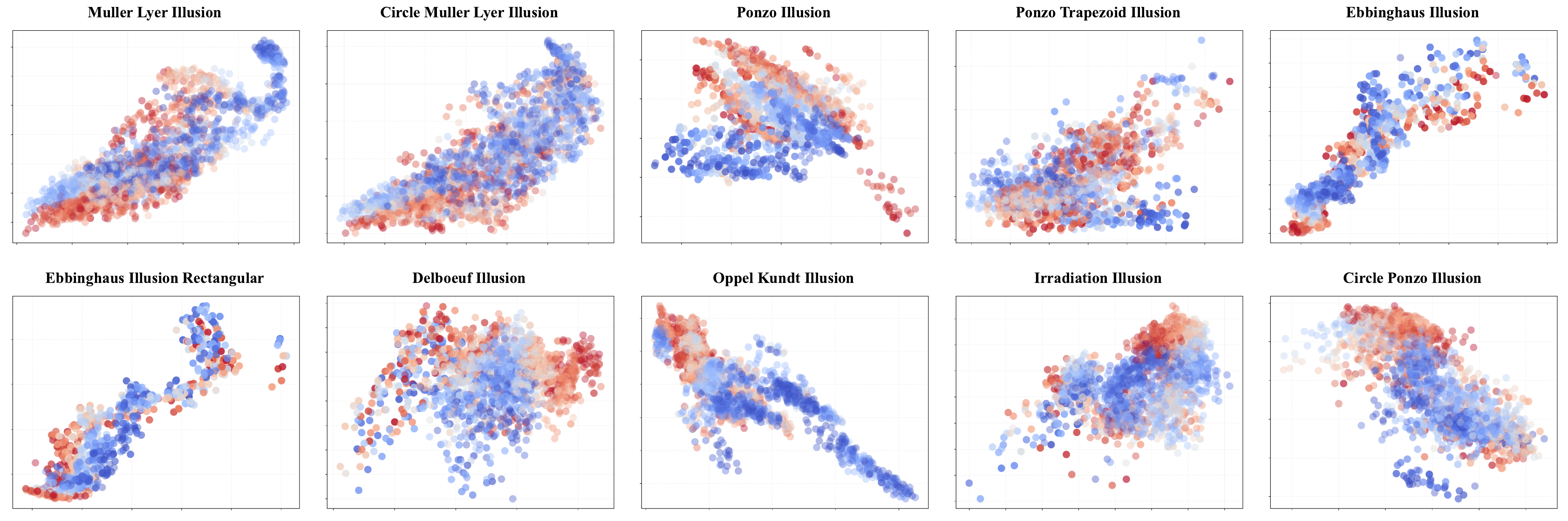}
\caption{Visualization of Qwen2.5-VL-72B embeddings on size cases color-coded by perturbation strength (excluding the case shown in the main paper). The clear separation between perturbation levels validates the data generation pipeline.}
\label{fig:pca_72b}
\end{figure}

\begin{figure}[H]
\centering
\includegraphics[width=\linewidth]{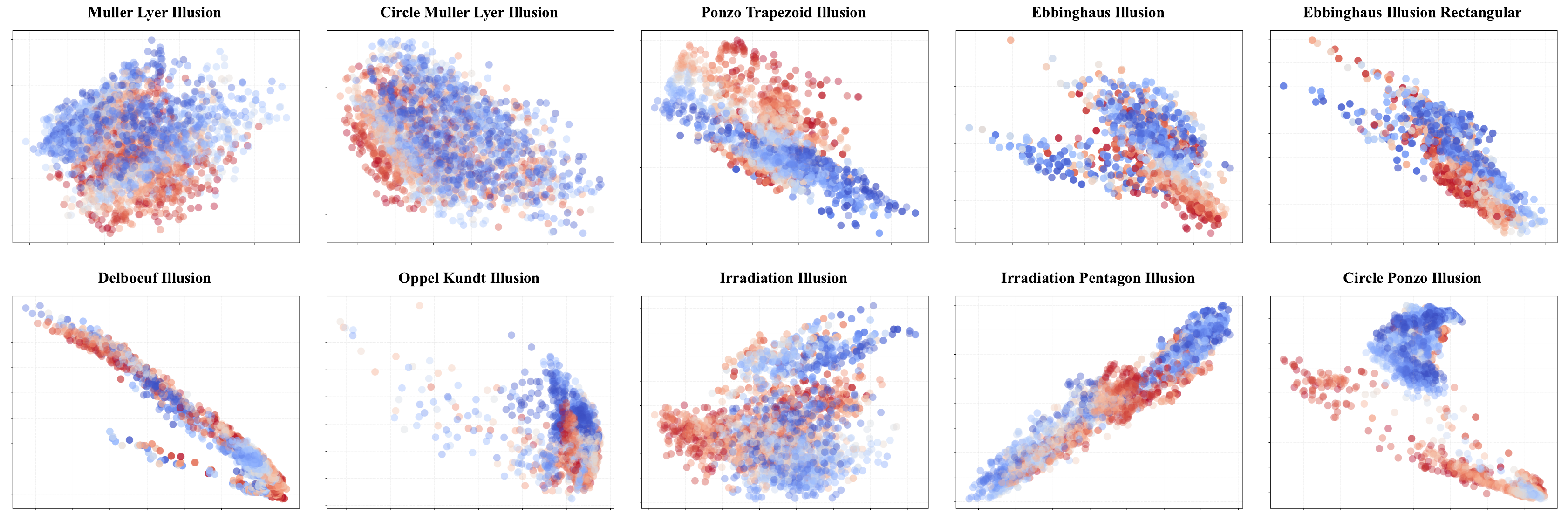}
\caption{Visualization of Qwen2.5-VL-3B embeddings on size cases color-coded by perturbation strength (excluding the case shown in the main paper). The clear separation between perturbation levels validates the data generation pipeline.}
\label{fig:pca_3b}
\end{figure}

\twocolumn



\end{document}